\def\BuildArxivFull{1}
\documentclass[journal]{IEEEtran}

\usepackage{cite}
\usepackage{amsmath,amssymb,amsfonts,amsthm}
\usepackage{booktabs}
\usepackage{multirow}
\usepackage{array}
\usepackage{graphicx}
\usepackage{xcolor}
\usepackage{colortbl}
\definecolor{plugbg}{RGB}{245,248,252}
\definecolor{oursbg}{RGB}{232,242,251}
\usepackage{enumitem}
\setlist[itemize,enumerate]{noitemsep,topsep=2pt,parsep=1pt,partopsep=0pt,leftmargin=1.35em}
\PassOptionsToPackage{hyphens}{url}
\usepackage[hidelinks]{hyperref}

\newcommand{\secbest}[1]{\underline{#1}}

\ifdefined\BuildArxivFull
  \newcommand{\suppsec}{Appendix~\ref{app:tables}}
  \newcommand{\supptabI}{Appendix Table~\ref{tab:app_main_direct}}
  \newcommand{\supptabII}{Appendix Table~\ref{tab:ablation_p12}}
  \newcommand{\supptabIII}{Appendix Table~\ref{tab:ablation_p34}}
  \newcommand{\supptabIV}{Appendix Table~\ref{tab:paradigm_fairness}}
  \newcommand{\supptabV}{Appendix Table~\ref{tab:wisig_mob_deepdive}}
  \newcommand{\supptabVI}{Appendix Table~\ref{tab:timing_burst}}
  \newcommand{\supptabVII}{Appendix Table~\ref{tab:sym_scale_multirx}}
  \newcommand{\supptabsIItoIII}{Appendix Tables~\ref{tab:ablation_p12} and \ref{tab:ablation_p34}}
  \newcommand{\supptabsVtoVII}{Appendix Tables~\ref{tab:wisig_mob_deepdive}, \ref{tab:timing_burst}, and \ref{tab:sym_scale_multirx}}
\else
  \newcommand{\suppsec}{the Supplementary Material}
  \newcommand{\supptabI}{Supplementary Table~S-I}
  \newcommand{\supptabII}{Supplementary Table~S-II}
  \newcommand{\supptabIII}{Supplementary Table~S-III}
  \newcommand{\supptabIV}{Supplementary Table~S-IV}
  \newcommand{\supptabV}{Supplementary Table~S-V}
  \newcommand{\supptabVI}{Supplementary Table~S-VI}
  \newcommand{\supptabVII}{Supplementary Table~S-VII}
  \newcommand{\supptabsIItoIII}{Supplementary Tables~S-II and S-III}
  \newcommand{\supptabsVtoVII}{Supplementary Tables~S-V, S-VI, and S-VII}
\fi

\newtheorem{proposition}{Proposition}
\newcounter{alg}

\AtBeginDocument{%
  \setlength{\abovedisplayskip}{3.8pt plus 1.0pt minus 1.0pt}%
  \setlength{\belowdisplayskip}{3.8pt plus 1.0pt minus 1.0pt}%
  \setlength{\abovedisplayshortskip}{2.0pt plus 0.8pt}%
  \setlength{\belowdisplayshortskip}{2.5pt plus 0.8pt}%
}

\begin{document}

\title{Pluggable Subband-Decoupled Fusion and Unlabeled Capture Prototype Calibration: Bridging High-SNR Null-Subcarrier Regrowth and Low-SNR Coherent Preambles for Cross-Domain IoT RF Fingerprinting}

\author{Fengchong~Yao, Jianbing~Li, Qing~Liu, Kefeng~Song, Haitao~Li, Song~Wang, and Kunpeng~Wang%
\thanks{Fengchong Yao, Jianbing Li, Qing Liu, Kefeng Song, Haitao Li, Song Wang, and Kunpeng Wang are with the School of Information Systems Engineering, Information Engineering University, Zhengzhou 450001, China (e-mail: phoenixly@126.com; li\_jb@126.com; liuqing8123@163.com; annx1990@163.com; lihaitao\_01@163.com; wangsong8190@163.com; wkp0110@163.com).}%
\thanks{Corresponding author: Jianbing Li (e-mail: li\_jb@126.com).}%
\thanks{All three datasets evaluated in this study are publicly available open-access benchmarks: IEEE DataPort-WiFi, UCLA WiSig, and ORBIT MobRFFI.}}

\markboth{IEEE INTERNET OF THINGS JOURNAL}%
{YAO \MakeLowercase{\textit{et al.}}: PLUGGABLE SUBBAND-DECOUPLED FUSION AND UNLABELED CAPTURE PROTOTYPE CALIBRATION FOR IOT RFFI}

\maketitle

\begin{abstract}
Zero-overhead physical-layer authentication at edge wireless access points (APs) is vital for securing resource-constrained Internet of Things (IoT) transceivers against credential cloning, yet cross-domain IEEE 802.11a/g radio frequency fingerprint identification (RFFI) faces an acute signal-to-noise ratio (SNR) crossover. Power-amplifier spectral regrowth on unpopulated null subcarriers resists indoor multipath at high SNR but decays quadratically below the thermal noise floor, whereas populated active in-band preambles survive low SNR yet degrade under severe multipath, causing static fusion to lose up to 44.7 percentage points (pp). To our knowledge, we propose the first RFFI framework that dynamically modulates active in-band waveforms and unpopulated null-subcarrier regrowth via a blind physical SNR gate: Pluggable Gated Subband-Decoupled Fusion with Unlabeled Capture-Aware Prototype Calibration (GSDF-CAPC). Avoiding the 20--35~dB subband gradient imbalance of joint training, GSDF pairs any frozen pretrained host backbone with a 46.3K null-subcarrier plug-in encoder (NS-Branch), fusing scale-standardized logits and gate-warped spherical embeddings under a source-calibrated lag-64 thermal-SNR gate with exact low-SNR fallback. Over unlabeled target AP captures, forward-only unbalanced optimal-transport calibration (UOT-CAPC) aligns shifted prototypes under packet-arrival skew and recovers parameter-free balanced Sinkhorn (U-CAPC) at equipartition. Across DataPort-WiFi, WiSig, and ORBIT MobRFFI, zero-retrain GSDF wins all 72/72 paired runs across eight backbones with 0.00~pp low-SNR loss, while edge IB-Host + GSDF (92.6K--93.3K active parameters, 1.82~ms single-core CPU latency) and timing-synchronized GSDF-STO / GSDF-Tri reach 57.50\%--65.05\% three-benchmark Macro-F1 (89.04\%--100.00\% over 5--10-packet IoT bursts).
\end{abstract}

\begin{IEEEkeywords}
Blind signal-to-noise ratio (SNR) gating, edge device authentication, IEEE 802.11a/g preamble, Internet of Things (IoT) security, null-subcarrier spectral regrowth, pluggable subband-decoupled fusion, radio frequency fingerprint identification (RFFI), unbalanced optimal transport.
\end{IEEEkeywords}

\section{Introduction}
\label{sec:intro}

\IEEEPARstart{P}{hysical-layer} radio frequency fingerprint identification (RFFI) authenticates wireless Internet of Things (IoT) transceivers directly from analog front-end hardware impairments---including power-amplifier (PA) amplitude-to-amplitude (AM/AM) and amplitude-to-phase (AM/PM) nonlinearities, quadrature I/Q mixer imbalance, and baseband filter tolerances---imprinted onto standard synchronization preambles~\cite{brik2008paradis,soltanieh2020review,jagannath2022survey,rffi_iot_auth}. Because MAC identifiers and pre-shared cryptographic keys can be cloned or extracted across low-cost smart-building sensors, industrial actuators, and consumer Wi-Fi IoT nodes, passive waveform-level verification at edge access points (APs) provides an essential zero-overhead defense~\cite{zhu2022sigpro,better_data_not_bigger,tifs_channel_receiver_independent}. Practical indoor Wi-Fi IoT deployments, however, violate the stationary-channel premise of single-room laboratory evaluations: an edge gateway classifier enrolled on one source AP must survive unseen room impulse responses, multi-week hardware drift, uncalibrated replacement APs, co-channel interference, single-core gateway CPU budgets without discrete GPUs, and wide thermal signal-to-noise ratio (SNR) variations spanning $>30$~dB near the AP down to 0--5~dB for cell-edge or through-wall IoT devices~\cite{hanna2022wisig,shen2024mobrffi,xie2024robustrffingerprintidentification}.

\begin{figure*}[!t]
\centering
\includegraphics[width=0.78\textwidth]{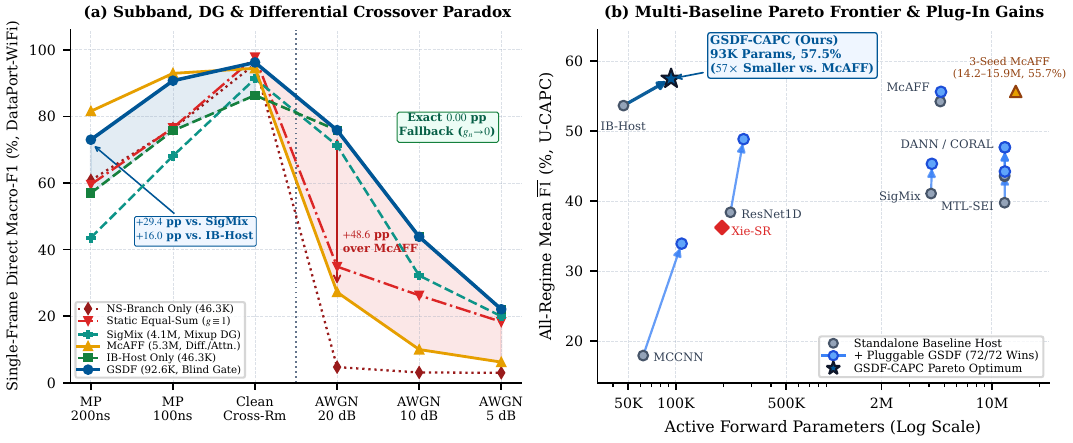}
\caption{Physical crossover paradox and multi-baseline Pareto frontier: (a) on \textbf{DataPort-WiFi} ($E_2,E_3\!\to\!E_1$, single-frame \textbf{Direct}), \textbf{NS-Branch} ($95.7\%$ clean), $5.3\text{M}$ \textbf{McAFF} ($81.6\%$ at $200$~ns), and $4.1\text{M}$ \textbf{SigMix} ($91.3\%$ clean) illustrate the three-way regime trade-off: \textbf{SigMix} drops to $43.5\%$ at $200$~ns multipath, \textbf{NS-Branch} and \textbf{McAFF} collapse at $\le 20$~dB AWGN ($4.7\%$ and $27.4\%$ at 20~dB), and unweighted \textbf{Static-Sum} ($g \equiv 1$) suffers $-44.7$~pp negative transfer, whereas pluggable \textbf{GSDF} gains $+29.4$~pp over \textbf{SigMix} and $+16.0$~pp over \textbf{IB-Host} at $200$~ns ($+48.6$~pp over \textbf{McAFF} at 20~dB) with exact $0.00$~pp low-SNR fallback; (b) on three-benchmark $\overline{\mathrm{F1}}$ (\textbf{U-CAPC}), zero-retrain \textbf{GSDF} improves all $8/8$ backbones ($72/72$ wins), with \textbf{GSDF-CAPC} ($93\text{K}$, $57.50\%$, $1.82$~ms CPU ONNX) surpassing $5.3\text{M}$ \textbf{McAFF} and $14.2\text{M}\text{--}15.9\text{M}$ \textbf{3-Seed McAFF}.}
\label{fig:paradox_pareto}
\end{figure*}

As shown in Fig.~\ref{fig:paradox_pareto}(a), mainstream channel-resilient RFFI paradigms face an acute physical trade-off across SNR and multipath regimes in indoor IoT WLANs. Classifiers operating on populated in-band waveforms---spanning multi-task adversarial/statistical alignment (\textbf{MTL-DANN}, \textbf{MTL-CORAL}, \textbf{MTL-SEI})~\cite{ganin2016domain,sun2016deep,mtl_adversarial_sei}, domain-generalization mixing (\textbf{SigMix})~\cite{wan2025sigmix}, and spatiotemporal CNN/attention backbones~\cite{channel_temporal_attention_rff,cross_attention_transformer,graph_conv_rff}---exploit coherent pilot references on active subcarriers to retain linear SNR scaling down to 5--10~dB additive white Gaussian noise (AWGN), yet degrade under severe room multipath ($\tau_{\mathrm{rms}} = 100\text{--}200$~ns, where \textbf{SigMix} drops to $43.55\%$ Direct F1) because weak emitter distortions are convolved with high-power nominal symbols~\cite{equalization_da_rff}. Multi-representation \textbf{McAFF}~\cite{multichannel_attentive_fusion} cancels quasi-static channels via differential and group-delay inputs, achieving strong high-SNR multipath accuracy, but symbol-pair multiplication squares thermal noise below 20~dB AWGN ($9.68\%$ Direct F1 at $10/5$~dB on \textbf{DataPort-WiFi}). Conversely, Xie \emph{et al.}~\cite{xie2024robustrffingerprintidentification} extracted odd-order PA spectral regrowth on unpopulated IEEE 802.11 guard/null/DC subcarriers ($S_{\mathrm{nom}}[m]=0$). Although unpopulated bins attain $95.70\%$ single-frame cross-room F1 at high SNR, regrowth power lies 20--35~dB below the active carrier without pilot phase references, collapsing to near-chance ($4.74\%$ at 20~dB; $3.03\%$ at 5~dB) once thermal noise overtakes intermodulation leakage, while explicit crystal carrier frequency offset (CFO) priors~\cite{xie2024robustrffingerprintidentification} drift with oscillator aging across unattended IoT operation~\cite{cfo_experimental_review}.

Unifying active in-band and unpopulated null-subcarrier cues at IoT edge gateways without target labels faces three coupled challenges:
\begin{enumerate}
\item \emph{Subband optimization interference and legacy-host incompatibility:} Because active subcarriers carry 20--35~dB higher power than null-bin regrowth, joint training lets active-band gradients dominate while leaking low-SNR null-bin noise into shared hidden layers, and learned fusion projections prevent zero-retrain field attachment to heterogeneous pretrained AP backbones.
\item \emph{Static and confidence-gating negative transfer:} Unweighted logit addition (\textbf{Static-Sum}) injects isotropic noise below 25~dB, losing $-44.67$~pp at 20~dB on \textbf{DataPort-WiFi} (Fig.~\ref{fig:paradox_pareto}(a)) and up to $-30.23$~pp at 10~dB across backbones; entropy weighting (\textbf{Entropy-Gate}) loses up to $-11.09$~pp from overconfident noise posteriors, while binary hard switching discards transition-SNR complementarity.
\item \emph{Edge test-time calibration under domain shift and traffic skew:} Backpropagation-based adaptation (SHOT~\cite{liang2020shot}, Tent~\cite{wang2021tent}) inflates gateway CPU latency and risks class collapse, whereas balanced Sinkhorn over-partitions dominant IoT emitters under sporadic packet arrivals or short capture windows~\cite{prototype_uda_sei,psfan}.
\end{enumerate}

To resolve these challenges on legacy IEEE 802.11a/g IoT preambles, we propose \textbf{GSDF-CAPC}, pairing a \emph{Zero-Retrain Pluggable Dual-Space Gated Subband-Decoupled Fusion} (\textbf{GSDF}) architecture with general forward-only \emph{KL-Relaxed Unbalanced Optimal-Transport Capture-Aware Prototype Calibration} (\textbf{UOT-CAPC}) and its closed-set balanced specialization (\textbf{U-CAPC}). To the best of our knowledge, this work is the first in RFFI to theoretically characterize the SNR Fisher-information crossover between populated active in-band waveforms and unpopulated null/guard-subcarrier spectral regrowth, and to bridge them via subband-decoupled, blind thermal-SNR-gated dual-space modulation for practical edge IoT authentication. Our main contributions are:
\begin{itemize}
\item \textbf{Subband SNR-crossover theory and first cross-subband SNR-gated plug-in (GSDF):} Proving quadratic $\mathcal{O}((\eta_{\mathrm{null}}\gamma)^2)$ low-SNR Fisher decay on phase-unreferenced null bins versus linear $\mathcal{O}(\gamma)$ coherent active scaling (Proposition~\ref{prop:snr_collapse}), \textbf{GSDF} departs from both active-only/coupled-attention RFFI~\cite{multichannel_attentive_fusion,wan2025sigmix} and static null-only RFFI~\cite{xie2024robustrffingerprintidentification} by decoupling parameter training ($\theta_{\mathrm{host}} \perp \theta_{\mathrm{null}}$) between any frozen pretrained host $\mathcal{H}_{\mathrm{host}}$ (instantiated natively as 46.3K \textbf{IB-Host} or as any deployed external RFFI backbone) and a 46.3K 128-bin null-subcarrier plug-in encoder (\textbf{NS-Branch}). Fusing per-frame $z$-standardized logits and gate-warped embeddings $\mathbb{S}^{d_{\mathrm{host}}+d_{\mathrm{null}}-1}$ guarantees projection-free compatibility for arbitrary $d_{\mathrm{host}}$, mixed-SNR noise isolation, and exact host fallback as $g_n \to 0$ (Proposition~\ref{prop:plugin_fallback}).
\item \textbf{Source-calibrated thermal-SNR gating and general UOT-CAPC $\to$ U-CAPC edge calibration:} A closed-form lag-64 median thermal-SNR gate $(\alpha^*,\beta^*)$ calibrated solely on source validation data drives fusion without target labels, paired with a three-step forward-only KL-regularized entropic unbalanced optimal-transport calibrator (\textbf{UOT-CAPC}, $\rho = \gamma_{\mathrm{KL}}/(\gamma_{\mathrm{KL}}+1) \in (0,1)$) whose $\gamma_{\mathrm{KL}} \to \infty$ ($\rho=1$) limit yields parameter-free balanced Sinkhorn calibration (\textbf{U-CAPC}) for roughly balanced closed-set AP sessions.
\item \textbf{72/72 plug-in generality, multi-regime WiSig/MobRFFI scaling, and edge gateway Pareto frontier:} Across \textbf{DataPort-WiFi}~\cite{xie2024robustrffingerprintidentification}, \textbf{WiSig}~\cite{hanna2022wisig}, and \textbf{ORBIT MobRFFI}~\cite{shen2024mobrffi} ($3\times 3$ split-by-seed grid, matched channel augmentations), zero-retrain \textbf{GSDF} wins all $72/72$ paired runs ($p_{\mathrm{Holm}} < 10^{-4}$) across eight backbones with $0.00$~pp low-SNR penalty---enabling \textbf{McAFF + GSDF}, \textbf{ResNet1D + GSDF}, and zero-retrain 3-way fusion ($59.14\%$ $\overline{\mathrm{F1}}$, $66.58\%$ on \textbf{MobRFFI} \texttt{19-19}) to set specialized highs ($94.54\%\text{--}97.94\%$ on two-week \textbf{WiSig}, $85.62\%\text{--}93.21\%$ across $K=15\text{--}27$ IoT emitters) while native \textbf{IB-Host + GSDF} leads three-benchmark $\overline{\mathrm{F1}}$ at 92.6K--93.3K parameters ($1.82$~ms single-core CPU ONNX), beats 5-epoch backpropagation SHOT on \textbf{MobRFFI} ($52.00\%$ vs.\ $45.33\%\text{--}47.29\%$), and climbs to $62.98\%$ (\textbf{GSDF-STO}) and $65.05\%$ (\textbf{GSDF-Tri}, $+10.81$ to $+17.76$~pp over symmetrically timing-tuned $185.6\text{K}\text{--}11.90\text{M}$ baselines, $73.05\%$ at $K_{\max}=47$, and $93.72\%$ under 2-RX source pooling), reaching $88.08\%\text{--}89.04\%$ (\textbf{MobRFFI}) and $99.20\%\text{--}100.00\%$ (\textbf{WiSig}) over zero-label $B=5\text{--}10$ packet bursts.
\end{itemize}

\section{Related Work}
\label{sec:related}

\subsection{Hardware-Impairment, Differential, and Null-Subcarrier IoT RFFI}
\label{sec:rel_phy}

Early radiometric device authentication extracted hand-crafted modulation errors, clock skew, and transient envelopes from IEEE 802.11 frames~\cite{brik2008paradis,soltanieh2020review}, whereas modern deep IoT RFFI learns directly from raw I/Q waveforms, short-time time-frequency maps, or explainable spectral probe features~\cite{jagannath2022survey,zhu2022sigpro,learnable_signal_representation,mrfe_multidimensional_rff,explainable_cnn_rff}. From a physical-layer standpoint, commodity IoT transmitter signatures arise primarily from odd-order memory-polynomial PA nonlinearities~\cite{morgan2006gmp,hardware_impairments_rffi}, quadrature modulator imbalance, and local-oscillator (LO) offsets~\cite{shen_lora_jsac_2021,cfo_experimental_review}. However, steady-state active subcarriers experience multiplicative indoor channel fading $H_a[m]$ that dominates subtle emitter distortions across rooms and AP placements~\cite{hanna2022wisig,better_data_not_bigger}.

To suppress quasi-static channel transfer functions without explicit channel estimation, differential schemes compute lag-$L$ symbol products, Differential Constellation Trace Figures (DCTFs)~\cite{mccnn_lte_open_set}, linear-to-logarithmic differential spectrum ratios~\cite{differential_spectrum_rff}, spatial-stream ratios~\cite{multiantenna_5g_rff}, or multi-representation combinations of I/Q, differential phase, FFT, and STFT maps (\textbf{McAFF})~\cite{multichannel_attentive_fusion,rcqcc_wifi}. While highly effective under clean indoor multipath, multiplying two noisy observations produces a noise-cross-noise term whose variance scales quadratically at low SNR. An orthogonal physical direction by Xie \emph{et al.}~\cite{xie2024robustrffingerprintidentification} isolates PA spectral regrowth on unpopulated IEEE 802.11a/g guard/null/DC subcarriers alongside crystal CFO priors. Yet unpopulated regrowth lies 20--35~dB below the active carrier without pilot phase references, and CFO drifts under ambient temperature variations and multi-week IoT device aging~\cite{cfo_experimental_review}. Critically, existing RFFI studies treat populated active subcarriers and unpopulated guard/null-subcarrier regrowth in isolation: active-band models discard channel-decoupled null-bin intermodulation, whereas null-bin methods~\cite{xie2024robustrffingerprintidentification} discard coherent active waveforms and suffer low-SNR collapse. Bridging both subbands for the first time links repeated-preamble blind SNR estimation~\cite{moose1994cfo,boumard2003snr,bouzegzi2010sigpro} to the exact active-versus-null Fisher-information crossover.

\subsection{Cross-Domain Alignment and Domain Generalization in IoT WLANs}
\label{sec:rel_dg}

A second major research line learns channel- and receiver-invariant representations on populated waveforms across distributed IoT receivers. Unsupervised domain adaptation (UDA) methods align source and target latent distributions via adversarial gradient reversal (\textbf{DANN})~\cite{ganin2016domain,receiver_impact_da,amtda_cross_receiver}, second-order covariance matching (\textbf{Deep CORAL})~\cite{sun2016deep}, dynamic or transformer-based feature alignment (\textbf{FATransformer})~\cite{fatransformer,dynamic_distribution_alignment,cross_day_da_rffi}, multi-task equalization/domain adversarial learning (\textbf{MTL-SEI})~\cite{mtl_adversarial_sei,equalization_da_rff}, or partial-set alignment~\cite{partial_da_cross_domain}. Because UDA requires concurrent access to source and target captures during central training, single- and multi-source domain generalization (DG) frameworks seek out-of-distribution robustness solely from source enrollment domains~\cite{gulrajani2021search,tifs_channel_receiver_independent}. Recent IoT RFFI DG methods introduce cross-time and cross-receiver waveform linear interpolation (\textbf{SigMix})~\cite{wan2025sigmix}, time-frequency contrastive CutMix~\cite{tfcl_cutmix_rff,supervised_contrastive_rff}, causal channel intervention~\cite{causal_channel_sei}, disentangled cross-combination~\cite{dg_cross_receiver_rffi,domain_invariant_rff,disentangled_cross_combination,generalizable_channel_resilient}, knowledge distillation~\cite{bert_kd_rff}, and attention-enhanced graph/temporal backbones~\cite{channel_temporal_attention_rff,cross_attention_transformer,graph_conv_rff}. Despite improving low-to-moderate SNR generalization, shared-band encoders remain physically limited at large indoor delay spreads ($\tau_{\mathrm{rms}} \ge 100$~ns), where multipath notches alias with in-band distortion.

\subsection{Multi-Branch Fusion and Lightweight Edge Test-Time Calibration}
\label{sec:rel_tta}

Combining heterogeneous RFFI views typically relies on end-to-end attentive concatenation or learned gating networks~\cite{multichannel_attentive_fusion,multisource_rff_fusion,jvmd_multi_sensor}. However, joint end-to-end training across subbands with a 20--35~dB power disparity causes high-energy in-band gradients to dominate optimization while entangling collapsed null-bin noise into shared batch/layer normalizations, and learned fusion layers cannot be attached zero-shot to third-party pretrained AP classifiers already deployed in the field. Moreover, prediction-confidence or entropy gates become unreliable when out-of-distribution noise triggers overconfident softmax posteriors.

At edge deployment, source-free domain adaptation and test-time adaptation (TTA) adapt pretrained models using only unlabeled target AP packets. Gradient-based TTA such as SHOT~\cite{liang2020shot} and Tent~\cite{wang2021tent} updates batch-normalization or encoder weights via entropy minimization, incurring backward-pass memory/latency overhead unsuitable for CPU-only IoT gateways and risking degenerate single-class collapse under small batches or skewed arrivals~\cite{boudiaf2022lame}. Forward-only classifier adjustment (\textbf{T3A}~\cite{iwasawa2021t3a}, \textbf{LAME}~\cite{boudiaf2022lame}) and prototype-based RFFI alignment (\textbf{PSFAN}~\cite{psfan}, prototype UDA/calibration~\cite{prototype_uda_sei,open_set_prototype_calibration,pseudo_label_semantic_drift}) avoid backpropagation, while Sinkhorn-Knopp optimal transport~\cite{cuturi2013sinkhorn,caron2020unsupervised} enforces cluster equipartition. Yet strict balanced Sinkhorn degrades when natural IoT packet arrivals are skewed or capture windows are short, motivating unbalanced optimal transport (\textbf{UOT}) with Kullback-Leibler marginal relaxation~\cite{chizat2018scaling,fatras2021unbalanced}. Unlike prior end-to-end multi-feature attention networks~\cite{multichannel_attentive_fusion,multisource_rff_fusion}, static null-subcarrier classifiers~\cite{xie2024robustrffingerprintidentification}, or standalone TTA methods, \textbf{GSDF-CAPC} introduces the first subband-decoupled, physical thermal-SNR-gated dual-space modulation between populated in-band waveforms and unpopulated null-subcarrier regrowth---guaranteeing zero-retrain pluggability and zero low-SNR negative transfer---coupled to a unified $\rho$-parameterized Sinkhorn/UOT edge calibrator.

\section{Physical Signal Model and SNR Crossover Analysis}
\label{sec:model}

\subsection{In-Band Active vs.\ Null-Subcarrier Preamble Observations}
\label{sec:subband_model}

\begin{figure*}[!t]
\centering
\includegraphics[width=0.92\textwidth]{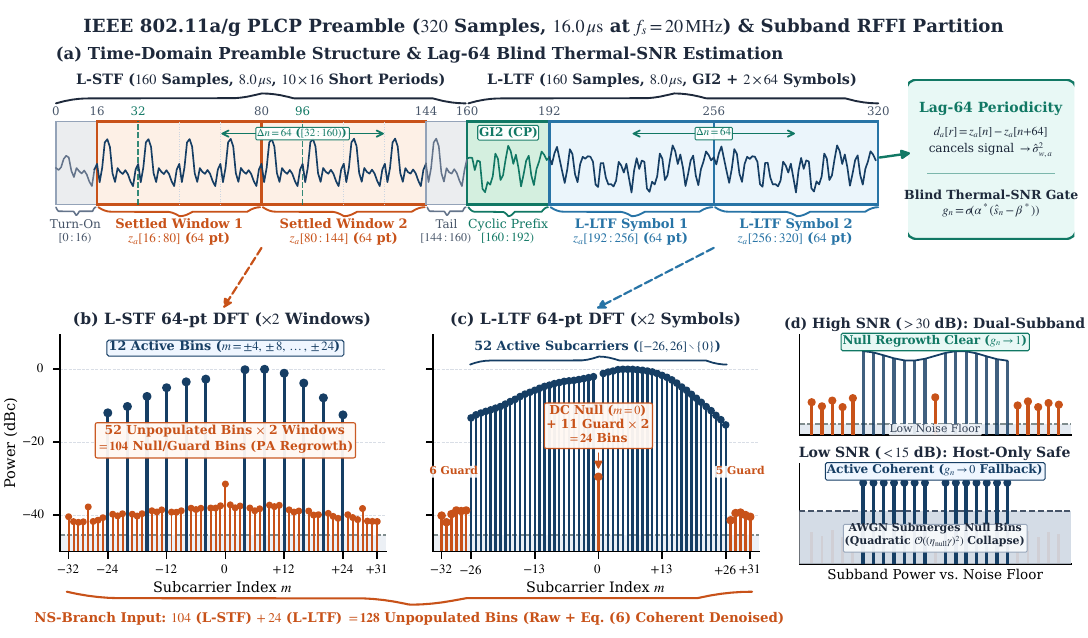}
\caption{IEEE 802.11a/g PLCP preamble time-frequency structure and physical subband trade-off (illustrated on an over-the-air \textbf{DataPort-WiFi} USRP X310 recording): (a)~320-sample time-domain frame showing the two settled 64-sample \textbf{L-STF} DFT windows ($z_a[16{:}80], z_a[80{:}144]$), cyclic prefix \textbf{GI2}, two 64-sample \textbf{L-LTF} symbols ($z_a[192{:}256], z_a[256{:}320]$), and lag-64 periodic cancellation over $z_a[32{:}96]\!\leftrightarrow\!z_a[96{:}160]$ and \textbf{L-LTF} driving the blind thermal-SNR gate $g_n$; (b)--(c)~centered 64-point DFT spectra isolating $2\times 52 = 104$ \textbf{L-STF} null/guard bins and $2\times 12 = 24$ \textbf{L-LTF} guard/DC bins ($128$ unpopulated PA-regrowth bins fed to \textbf{NS-Branch}); (d)~high-SNR multipath immunity ($g_n \to 1$) versus low-SNR quadratic $\mathcal{O}((\eta_{\mathrm{null}}\gamma)^2)$ collapse ($g_n \to 0$ host fallback, Proposition~\ref{prop:snr_collapse}).}
\label{fig:preamble}
\end{figure*}

Consider a closed set of $K$ enrolled IEEE 802.11a/g IoT transmitters indexed by $y \in \{1,\dots,K\}$, observed by an edge access point with $A \ge 1$ synchronized RF chains sampled at $f_s = 20$~MHz. As illustrated in Fig.~\ref{fig:preamble}(a), each physical-layer convergence protocol (PLCP) preamble $\mathbf{Z} \in \mathbb{C}^{A \times 320}$ spans $N_{\mathrm{pre}} = 320$ complex baseband samples ($16~\mu\text{s}$): the 160-sample legacy short training field (\textbf{L-STF}, $n \in [0,159]$, ten 16-sample repetitions), the 32-sample guard interval (\textbf{GI2}, $n \in [160,191]$), and the 128-sample legacy long training field (\textbf{L-LTF}, $n \in [192,319]$, two identical 64-sample OFDM symbols). Prior to feature extraction, every frame is normalized to unit root-mean-square (RMS) power across all $320A$ samples ($\varepsilon = 10^{-12}$) to remove bulk path-loss/distance shortcuts:
\begin{equation}
\tilde{z}_a[n] = \frac{z_a[n]}{\sqrt{\frac{1}{320A}\sum_{a'=1}^{A}\sum_{n'=0}^{319}|z_{a'}[n']|^2 + \varepsilon}},
\label{eq:rms_norm}
\end{equation}
writing $z_a[n]$ hereafter for the unit-RMS complex preamble.

Let $s_{\mathrm{nom}}[n]$ denote the ideal standard preamble. Passing $s_{\mathrm{nom}}[n]$ through emitter $y$'s baseband filter, quadrature upconverter, and nonlinear PA~\cite{morgan2006gmp} produces
\begin{equation}
x_y[n] = s_{\mathrm{nom}}[n] + \delta_{\mathrm{in},y}[n] + \delta_{\mathrm{null},y}[n],
\label{eq:tx_decomp}
\end{equation}
where $\delta_{\mathrm{in},y}[n]$ and $\delta_{\mathrm{null},y}[n]$ represent emitter $y$'s hardware distortion---dominated by odd-order memory-polynomial PA intermodulation---projected onto populated active subcarriers $\mathcal{K}_{\mathrm{act}}$ and unpopulated guard/null/DC subcarriers $\mathcal{K}_{\mathrm{null}}$, respectively. Indexing a 64-point centered DFT by $m \in \{-32,\dots,31\}$ (Fig.~\ref{fig:preamble}(b)--(c)), each 64-sample L-STF window excites 12 subcarriers $\mathcal{K}_{\mathrm{act}}^{\mathrm{STF}} = \{\pm 4, \pm 8, \pm 12, \pm 16, \pm 20, \pm 24\}$ (leaving 52 unpopulated bins), whereas each 64-sample L-LTF symbol excites 52 subcarriers $\mathcal{K}_{\mathrm{act}}^{\mathrm{LTF}} = [-26,26]\setminus\{0\}$ (leaving 12 guard and DC bins $\{-32,\dots,-27,0,27,\dots,31\}$)~\cite{xie2024robustrffingerprintidentification}.

Propagating through a quasi-static multipath channel $H_a[m]$ with circularly symmetric complex AWGN $W_a[m] \sim \mathcal{CN}(0, \sigma_w^2)$, the received 64-point DFT bin at antenna $a$ is
{\small
\begin{equation}
Y_a[m] = \left\{
\begin{array}{@{}l@{\;\;}l@{}}
H_a[m](S_{\mathrm{nom}}[m] + \Delta_{\mathrm{in},y}[m]) + W_a[m], & m \in \mathcal{K}_{\mathrm{act}}, \\[2pt]
H_a[m]\,\Delta_{\mathrm{null},y}[m] + W_a[m], & m \in \mathcal{K}_{\mathrm{null}}.
\end{array}\right.
\label{eq:subband_obs}
\end{equation}}
On active bins $m \in \mathcal{K}_{\mathrm{act}}$, subtle hardware distortion $\Delta_{\mathrm{in},y}[m]$ ($\|\Delta_{\mathrm{in},y}\| \ll \|S_{\mathrm{nom}}\|$) is superposed onto $H_a[m]S_{\mathrm{nom}}[m]$, leaving active waveforms sensitive to multipath dispersion while $S_{\mathrm{nom}}[m]$ provides a coherent phase reference down to low SNR. On unpopulated bins $m \in \mathcal{K}_{\mathrm{null}}$, $S_{\mathrm{nom}}[m] = 0$ eliminates nominal-symbol multipath interference and isolates PA regrowth $\Delta_{\mathrm{null},y}[m]$~\cite{xie2024robustrffingerprintidentification}; however, fractional regrowth power $\eta_{\mathrm{null}} \triangleq \mathbb{E}[|H_a[m]\Delta_{\mathrm{null},y}[m]|^2] / P_{\mathrm{sig}} \ll 1$ is weak ($-35$ to $-20$~dB) and carries no pilot symbol to align random channel phases across captures.

\subsection{Quadratic Low-SNR Collapse of Null-Subcarrier Energy}
\label{sec:prop1}

\begin{figure*}[!t]
\centering
\includegraphics[width=0.95\textwidth]{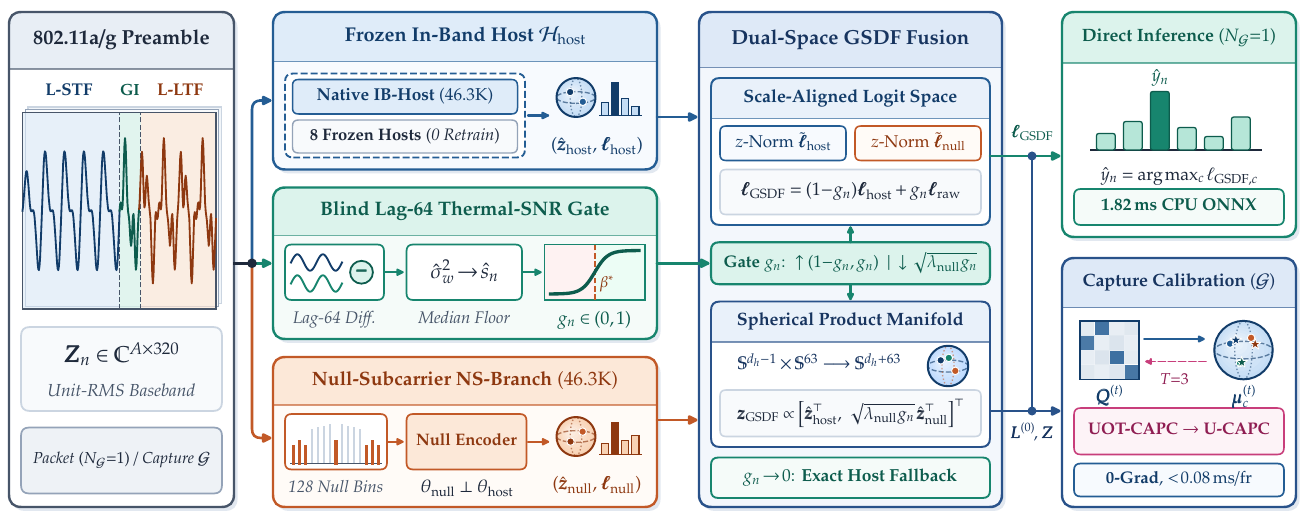}
\caption{Overview of pluggable \textbf{GSDF-CAPC} for IoT edge access points. Each unit-RMS IEEE 802.11a/g preamble $\mathbf{Z} \in \mathbb{C}^{A\times 320}$ feeds a frozen in-band host $\mathcal{H}_{\mathrm{host}}$---natively 46.3K \textbf{IB-Host} or any pretrained external RFFI backbone without retraining---an independently trained 128-bin null-subcarrier encoder (\textbf{NS-Branch}, 46.3K), and a source-calibrated lag-64 thermal-SNR gate $(\alpha^*,\beta^*)$. Dual-space \textbf{GSDF} fuses scale-standardized logits and spherical product embeddings with exact host fallback as $g_n \to 0$, supporting single-frame \textbf{Direct} inference (1.82~ms CPU ONNX) and forward-only optimal-transport calibration (\textbf{UOT-CAPC} at $\rho \in (0,1)$, specializing to balanced \textbf{U-CAPC} as $\rho \to 1$).}
\label{fig:overview}
\end{figure*}

Let $\gamma \triangleq P_{\mathrm{sig}} / \sigma_w^2$ denote per-subcarrier in-band SNR, and let $\theta_y \in \mathbb{R}$ parameterize emitter $y$'s normalized hardware perturbation.

\begin{proposition}[Quadratic Low-SNR Collapse of Null-Subcarrier Energy]
\label{prop:snr_collapse}
Across captures with random subband channel phases $\phi_a[m] \sim \mathcal{U}[0,2\pi)$, suppose an unpopulated bin $m \in \mathcal{K}_{\mathrm{null}}$ is observed via its phase-invariant normalized energy $U_m = 2|Y_a[m]|^2 / \sigma_w^2$ under $W_a[m] \sim \mathcal{CN}(0, \sigma_w^2)$, with non-centrality $\lambda_m(\theta_y) = 2\eta_{\mathrm{null}}\gamma\, g_m(\theta_y)$, where $g_m(\theta_y) > 0$ and $|g'_m(\theta_y)| = \Theta(1)$. On an active bin $m \in \mathcal{K}_{\mathrm{act}}$ where $H_a[m]S_{\mathrm{nom}}[m]$ provides a coherent phase reference, let the phase-aligned residual have mean $\sqrt{\gamma}\,\mu_m(\theta_y)$ with $|\mu'_m(\theta_y)| = \Theta(1)$ in $\mathcal{CN}(0,1)$ noise. As $\eta_{\mathrm{null}}\gamma \to 0$:
\begin{equation}
\begin{aligned}
\mathcal{I}_{\mathrm{null}}(\theta_y) &= (\eta_{\mathrm{null}}\gamma)^2\bigl(g'_m(\theta_y)\bigr)^2 + \mathcal{O}\bigl((\eta_{\mathrm{null}}\gamma)^3\bigr), \\
\mathcal{I}_{\mathrm{in}}(\theta_y) &= 2\gamma\bigl|\mu'_m(\theta_y)\bigr|^2.
\end{aligned}
\label{eq:fisher_scaling}
\end{equation}
Thus, null-subcarrier Fisher information decays quadratically as $\mathcal{O}((\eta_{\mathrm{null}}\gamma)^2)$ at low SNR, whereas coherent active-subcarrier Fisher information scales linearly as $\mathcal{O}(\gamma)$.
\end{proposition}

\begin{IEEEproof}
Conditioned on $|H_a[m]\Delta_{\mathrm{null},y}[m]|^2 = \eta_{\mathrm{null}} P_{\mathrm{sig}} g_m(\theta_y)$ and marginalizing over $\phi_a[m] \sim \mathcal{U}[0,2\pi)$, $U_m = 2|Y_a[m]|^2/\sigma_w^2$ is a phase-invariant sufficient statistic following a non-central $\chi_2^2$ density with non-centrality $\lambda = \lambda_m(\theta_y)$:
\begin{equation}
p(u;\lambda) = \frac{1}{2}\exp\!\left(-\frac{u+\lambda}{2}\right) I_0\!\bigl(\sqrt{\lambda u}\bigr), \quad u \ge 0,
\label{eq:ncx2_pdf}
\end{equation}
where $I_0(z) = \sum_{r=0}^{\infty}\frac{(z/2)^{2r}}{(r!)^2}$. Since $0 \le I'_0(z)/I_0(z) = I_1(z)/I_0(z) \le \min(1, z/2)$ for all $z \ge 0$, $|\partial_\lambda \log p(u;\lambda)| = \bigl|-\frac{1}{2} + \frac{\sqrt{u}}{2\sqrt{\lambda}}\frac{I_1(\sqrt{\lambda u})}{I_0(\sqrt{\lambda u})}\bigr| \le \frac{1}{2} + \frac{u}{4}$, providing a $\lambda$-uniform $p(u;\lambda)$-integrable quadratic envelope on $[0,\lambda_0]$ that justifies differentiation and series expansion inside the Fisher expectation as $\lambda \to 0^+$. Expanding $\log I_0(\sqrt{\lambda u}) = \frac{\lambda u}{4} - \frac{\lambda^2 u^2}{64} + \mathcal{O}(\lambda^3 u^3)$ yields $\partial_\lambda \log p(u;\lambda) = \frac{u - 2}{4} - \frac{\lambda u^2}{32} + \mathcal{O}(\lambda^2 u^3)$. Under $p(u;0) = \frac{1}{2}e^{-u/2}$, $\mathbb{E}_0[(U_m - 2)^2] = 4$, giving $\mathcal{I}_\lambda(\lambda) = \frac{1}{4} + \mathcal{O}(\lambda)$. By $\partial_{\theta_y}\lambda_m(\theta_y) = 2\eta_{\mathrm{null}}\gamma\, g'_m(\theta_y)$, the Fisher chain rule yields $\mathcal{I}_{\mathrm{null}}(\theta_y) = (\eta_{\mathrm{null}}\gamma)^2(g'_m(\theta_y))^2 + \mathcal{O}((\eta_{\mathrm{null}}\gamma)^3)$, while differentiating $-\bigl|z - \sqrt{\gamma}\,\mu_m(\theta_y)\bigr|^2 - \log\pi$ on $m \in \mathcal{K}_{\mathrm{act}}$ gives $\mathcal{I}_{\mathrm{in}}(\theta_y) = 2\gamma|\mu'_m(\theta_y)|^2$.
\end{IEEEproof}

Proposition~\ref{prop:snr_collapse} establishes a quantitative physical crossover at $\eta_{\mathrm{null}}\gamma_{\times} \approx 1$, corresponding to $\gamma_{\times} \approx -10\log_{10}\eta_{\mathrm{null}} \in [25, 35]$~dB for commodity Wi-Fi IoT PAs ($\eta_{\mathrm{null}} \in [-35, -25]$~dB). Above $\gamma_{\times}$, unpopulated bins avoid active-symbol multipath entanglement (whereas on active bins, $\mathcal{O}(\gamma)$ scaling retains emitter components orthogonal to smooth $H_a[m]$ until large delay spreads shrink channel coherence bandwidth and attenuate $|\mu'_m(\theta_y)|$); below $\gamma_{\times}$, null bins suffer a 20~dB Fisher-information cliff per 10~dB SNR drop---matching Fig.~\ref{fig:paradox_pareto}(a) where \textbf{NS-Branch} plunges from $95.70\%$ clean to $4.74\%$ at 20~dB AWGN---and motivating decoupled training with SNR-gated dual-space fusion (Fig.~\ref{fig:overview}).

\section{Proposed Pluggable GSDF-CAPC Framework}
\label{sec:method}

\subsection{Host-Agnostic Subband Interface and Decoupled Time-Frequency Encoders}
\label{sec:encoders}

\subsubsection{Why Subband Parameter Decoupling (\texorpdfstring{$\theta_{\mathrm{host}} \perp \theta_{\mathrm{null}}$}{theta\_host perp theta\_null}) Is Essential}
Jointly training active and unpopulated subbands in a shared-trunk network suffers from two pathologies rooted in~\eqref{eq:subband_obs}: (i)~\emph{gradient starvation under a 20--35~dB power gap}, where high-SNR active waveforms provide an easy source-domain shortcut that starves weak odd-order PA intermodulation gradients; and (ii)~\emph{irreversible hidden-layer noise entanglement}, where thermal noise on collapsed null bins below $\gamma_{\times}$ contaminates shared normalization statistics and activations, precluding exact test-time isolation. For IoT edge gateways, \textbf{GSDF} enforces strict parameter orthogonality ($\theta_{\mathrm{host}} \perp \theta_{\mathrm{null}}$) and a host-agnostic plug-in interface: any pretrained AP host $\mathcal{H}_{\mathrm{host}}$ mapping $\mathbf{Z}_n$ to logits $\boldsymbol{\ell}_{\mathrm{host}}(n) \in \mathbb{R}^K$ and unit embedding $\hat{\mathbf{z}}_{\mathrm{host}}(n) \in \mathbb{S}^{d_{\mathrm{host}}-1}$ ($d_{\mathrm{host}} \ge 1$) pairs at inference with frozen $\mathcal{E}_{\mathrm{null}}$ without retraining deployed host weights or fitting cross-model projections.

\subsubsection{Native Lightweight Time-Domain Host (\textbf{IB-Host})}
Our default compact edge host ($\boldsymbol{\ell}_{\mathrm{host}} = \boldsymbol{\ell}_{\mathrm{in}}$, $\hat{\mathbf{z}}_{\mathrm{host}} = \hat{\mathbf{z}}_{\mathrm{in}} \in \mathbb{S}^{63}$, $d_{\mathrm{host}}=64$) formats $\mathbf{Z}_n \in \mathbb{C}^{A\times 320}$ as two real I/Q channels per antenna and applies a shared patch stem $\mathrm{Conv1d}(2 \to 32,\text{kernel }8,\text{stride }8)$ ($40$ temporal tokens, $d = 32$), sliced learnable positional embeddings, and a two-layer Transformer encoder ($H = 4$, $d_{\mathrm{ff}} = 128$, GELU, dropout $0.1$)~\cite{vaswani2017attention}. Adaptive pooling condenses each antenna to $8$ tokens; all $8A$ tokens interact via a single-layer $4$-head graph self-attention block~\cite{velickovic2018graph} with active-antenna mean pooling (smoothly reducing to $8$-token temporal self-attention at $A=1$), followed by a depthwise-separable $\mathrm{Conv1d}(\text{kernel }3)$ block, scalar attention pooling, a $32 \to 64 \to 64$ LayerNorm/$\ell_2$ projection head, and a linear classifier ($4.36\text{M}$ MACs / $8.71$~MFLOPs at $A=4$).

\subsubsection{Null-Subcarrier Spectral Plug-In Encoder (\textbf{NS-Branch})}
Following Xie \emph{et al.}'s 128-bin unpopulated partition~\cite{xie2024robustrffingerprintidentification} (Fig.~\ref{fig:preamble}(b)--(c)), $\mathcal{E}_{\mathrm{null}}$ computes four 64-point centered DFTs per antenna across two settled L-STF windows ($z_a[16:80]$, $z_a[80:144]$, skipping the 16-sample turn-on transient) and two L-LTF symbols ($z_a[192:256]$, $z_a[256:320]$). Collecting 52 unpopulated bins ($\mathcal{K}_{\mathrm{null}}^{\mathrm{STF}} = [-32,31]\setminus\{\pm 4,\pm 8,\pm 12,\pm 16,\pm 20,\pm 24\}$) per L-STF window ($104$ bins) and 12 guard/DC bins ($\mathcal{K}_{\mathrm{null}}^{\mathrm{LTF}} = \{-32,\dots,-27,0,27,\dots,31\}$) per L-LTF symbol ($24$ bins) forms 128 bins per antenna. Stacking raw magnitudes $|Y_a[m]|$ with symbol-pair noise-subtracted coherent magnitudes forms $\mathbf{V}_a \in \mathbb{R}^{2 \times 128}$; for symmetric pair $(Y_{a,1}[m], Y_{a,2}[m])$ on $m \in \mathcal{K}_{\mathrm{null}}$ with $\bar{Y}_a[m] = \frac{1}{2}(Y_{a,1}[m] + Y_{a,2}[m])$ and $\hat{\nu}_a^2 = \frac{1}{4|\mathcal{K}_{\mathrm{null}}|}\sum_{m' \in \mathcal{K}_{\mathrm{null}}}|Y_{a,1}[m'] - Y_{a,2}[m']|^2$:
\begin{equation}
V_{a,\mathrm{coh}}[m] = \sqrt{\max\!\bigl(|\bar{Y}_a[m]|^2 - \hat{\nu}_a^2,\, 10^{-12}\bigr)}.
\label{eq:null_coh}
\end{equation}
Duplicating the $52$ L-STF and $12$ L-LTF pair-denoised values $V_{a,\mathrm{coh}}[m]$ across both windows of each symmetric pair yields $128$ denoised entries aligned index-by-index with the $128$ raw magnitudes in $\mathbf{V}_a \in \mathbb{R}^{2\times 128}$. An isomorphic neural architecture ($\mathrm{Conv1d}(2 \to 32,\text{kernel }8,\text{stride }8)$ over the 128 spectral bins yielding $16$ spectral tokens, two-layer Transformer, $8A$-token graph attention, and $64$-D head) outputs $\hat{\mathbf{z}}_{\mathrm{null}}(n) \in \mathbb{S}^{63}$ ($d_{\mathrm{null}}=64$) and $\boldsymbol{\ell}_{\mathrm{null}}(n) \in \mathbb{R}^K$. Because token length drops from $40$ to $16$, \textbf{NS-Branch} has the exact same active parameter count as \textbf{IB-Host} (46,315--46,640; Section~\ref{sec:setup}) while requiring $2.2\times$ fewer encoder operations ($1.94\text{M}$ encoder MACs / $3.89$~MFLOPs, plus $0.13\text{M}$ MACs / $0.26$~MFLOPs for the four fixed $64$-point DFT real matrix multiplications, totaling $2.07\text{M}$ MACs / $4.14$~MFLOPs at $A=4$). Beyond isolating PA regrowth, time-frequency decoupling adds timing complementarity: cyclic-prefix shifts $\Delta n < 0$ alias \textbf{IB-Host}'s $\text{stride}=8$ stem while leaving 64-point DFT magnitudes invariant ($|e^{-j2\pi m\Delta n/64}|=1$). Standalone \textbf{NS-Branch} matches Xie \emph{et al.}'s spectral regrowth CNN (\textbf{Xie-SR}, 195.4K--196.0K params; $36.52\%$ vs.\ $36.22\%$ $\overline{\mathrm{F1}}$, Section~\ref{sec:module_ablations}) at $4.2\times$ smaller size.

\subsubsection{Independent Source Training and Dual Parameter Audit}
Each encoder is trained independently on labeled source preambles for 80 epochs (AdamW, learning rate $10^{-3}$) under $\mathcal{L}_{\mathrm{CE}} + 0.35\,\mathcal{L}_{\mathrm{SupCon}}$~\cite{khosla2020supervised} and identical \emph{matched channel augmentations} (applied across all baselines, Section~\ref{sec:setup}): Rician/Rayleigh multipath ($\tau_{\mathrm{rms}} \in [10, 200]$~ns), AWGN ($\mathrm{SNR} \in [5, 35]$~dB), cyclic-prefix jitter, and passband ripple. Pruning unexecuted optional style/CFO ablation heads from raw checkpoints (86,364--86,689 params/branch; 172,730--173,380 total) leaves \textbf{46,315--46,640 active forward parameters per branch}---plus $+2$ blind SNR-gate scalars $(\alpha^*,\beta^*)$ totaling \textbf{92,632} ($K=10$) and \textbf{93,282} ($K=15$)---with exact $0.0$ output difference across all 224,640 test frames (detailed in Section~\ref{sec:setup}).

\subsection{Pluggable Dual-Space Gated Subband-Decoupled Fusion (GSDF)}
\label{sec:gsdf_fusion}

Given frame $n$'s reliability gate $g_n \in (0,1)$ (Section~\ref{sec:snr_gate}), \textbf{GSDF} dynamically modulates any frozen active-subband host $\mathcal{H}_{\mathrm{host}}$ and frozen null-subband \textbf{NS-Branch} simultaneously in logit space and on a gate-warped hypersphere. Unlike conventional neural channel/feature attention~\cite{multichannel_attentive_fusion}---which couples branch gradients during training, locks fusion layers to one architecture, and can assign nonzero weights to OOD noise---\textbf{GSDF} uses zero cross-branch parameters while preserving single-frame margins and multi-frame geodesic geometry.

\subsubsection{Scale-Aligned Logit-Space Fusion}
Pretrained backbones across CNNs, ResNets, and adversarial networks exhibit disparate logit dynamic ranges and class-mean offsets; naive addition lets high-temperature models dominate or be dominated by \textbf{NS-Branch}. For $\mathrm{b} \in \{\mathrm{host}, \mathrm{null}\}$, \textbf{GSDF} standardizes logits per frame across $K$ classes via $\mu_{\mathrm{b},n} = \frac{1}{K}\sum_{c=1}^K \ell_{\mathrm{b},c}(n)$ and $\sigma_{\mathrm{b},n} = \bigl[\frac{1}{K-1}\sum_{c=1}^K (\ell_{\mathrm{b},c}(n) - \mu_{\mathrm{b},n})^2\bigr]^{1/2}$:
\begin{equation}
\tilde{\boldsymbol{\ell}}_{\mathrm{b}}(n) = \frac{\boldsymbol{\ell}_{\mathrm{b}}(n) - \mu_{\mathrm{b},n}\mathbf{1}_K}{\max(\sigma_{\mathrm{b},n},\, 10^{-4})}, \quad \mathrm{b} \in \{\mathrm{host}, \mathrm{null}\}.
\label{eq:logit_znorm}
\end{equation}
Forming $\boldsymbol{\ell}_{\mathrm{raw}}(n) = \kappa_{\ell}\bigl(\tilde{\boldsymbol{\ell}}_{\mathrm{host}}(n) + \lambda_{\mathrm{null}} g_n \tilde{\boldsymbol{\ell}}_{\mathrm{null}}(n)\bigr)$, \textbf{GSDF} interpolates back to the host's native unstandardized logits:
\begin{equation}
\boldsymbol{\ell}_{\mathrm{GSDF}}(n) = (1 - g_n)\,\boldsymbol{\ell}_{\mathrm{host}}(n) + g_n\,\boldsymbol{\ell}_{\mathrm{raw}}(n),
\label{eq:gsdf_logit}
\end{equation}
with $(\lambda_{\mathrm{null}}, \kappa_{\ell}) = (1.15, 2.5)$ frozen across all datasets, seeds, and hosts (eliminated in our temperature-calibrated log-probability verification arm, Section~\ref{sec:wisig_mob_deepdive}). Equation~\eqref{eq:logit_znorm} aligns arbitrary host logit scales without per-host tuning, while~\eqref{eq:gsdf_logit} preserves $\mathcal{H}_{\mathrm{host}}$'s exact native boundaries as $g_n \to 0$.

\subsubsection{Gate-Warped Spherical Product-Manifold Fusion}
Because $\hat{\mathbf{z}}_{\mathrm{host}}(n) \in \mathbb{S}^{d_{\mathrm{host}}-1}$ ($d_{\mathrm{host}} \in \{64, 84, 256, 512\}$ across our eight hosts) and $\hat{\mathbf{z}}_{\mathrm{null}}(n) \in \mathbb{S}^{d_{\mathrm{null}}-1}$ ($d_{\mathrm{null}} = 64$) inhabit unaligned spaces of distinct dimensions, \textbf{GSDF} avoids trainable cross-model projections by mapping $\mathbb{S}^{d_{\mathrm{host}}-1} \times \mathbb{S}^{d_{\mathrm{null}}-1}$ onto $\mathbb{S}^{d_{\mathrm{host}} + d_{\mathrm{null}} - 1}$ via gate-modulated orthogonal concatenation:
\begin{equation}
\begin{aligned}
\mathbf{z}_{\mathrm{GSDF}}(n) &= \frac{\bigl[\hat{\mathbf{z}}_{\mathrm{host}}(n)^{\top},\; \sqrt{\lambda_{\mathrm{null}} g_n}\,\hat{\mathbf{z}}_{\mathrm{null}}(n)^{\top}\bigr]^{\top}}{\sqrt{1 + \lambda_{\mathrm{null}} g_n}} \\
&\in \mathbb{S}^{d_{\mathrm{host}} + d_{\mathrm{null}} - 1}.
\end{aligned}
\label{eq:gsdf_emb}
\end{equation}

\begin{proposition}[Dimension-Free Plug-In Compatibility, Cross-Frame Noise Isolation, and Exact Host Fallback]
\label{prop:plugin_fallback}
For any pretrained host $\mathcal{H}_{\mathrm{host}}$ of arbitrary dimension $d_{\mathrm{host}} \ge 1$ with $\|\hat{\mathbf{z}}_{\mathrm{host}}(n)\|_2 = 1$, any $\hat{\mathbf{z}}_{\mathrm{null}}(n) \in \mathbb{S}^{d_{\mathrm{null}}-1}$, and any frame pair $(n, n')$ in capture $\mathcal{G}$:
\begin{enumerate}[leftmargin=*]
\item \emph{Projection-free spherical metric decomposition:} $\|\mathbf{z}_{\mathrm{GSDF}}(n)\|_2 = 1$, and pairwise cosine similarity decomposes without cross-subspace interference as
{\small
\begin{equation}
\begin{aligned}
&\bigl\langle \mathbf{z}_{\mathrm{GSDF}}(n), \mathbf{z}_{\mathrm{GSDF}}(n')\bigr\rangle \\
&= \frac{\langle \hat{\mathbf{z}}_{\mathrm{host}}(n), \hat{\mathbf{z}}_{\mathrm{host}}(n')\rangle + \lambda_{\mathrm{null}}\sqrt{g_n g_{n'}}\langle \hat{\mathbf{z}}_{\mathrm{null}}(n), \hat{\mathbf{z}}_{\mathrm{null}}(n')\rangle}{\sqrt{(1 + \lambda_{\mathrm{null}} g_n)(1 + \lambda_{\mathrm{null}} g_{n'})}}.
\end{aligned}
\label{eq:gram_decomp}
\end{equation}}
\item \emph{Cross-frame noise isolation in mixed-SNR captures:} If frame $n \in \mathcal{G}$ suffers a fade or noise burst ($g_n \to 0$) while $n' \in \mathcal{G}$ has high SNR ($g_{n'} \approx 1$), the null cross-term $\lambda_{\mathrm{null}}\sqrt{g_n g_{n'}}\langle\hat{\mathbf{z}}_{\mathrm{null}}(n), \hat{\mathbf{z}}_{\mathrm{null}}(n')\rangle$ vanishes at rate $\mathcal{O}(\sqrt{g_n})$, preventing noisy null-bin embeddings from polluting high-SNR prototypes.
\item \emph{Exact low-SNR host fallback:} As $g_{\max}(\mathcal{G}) \triangleq \max_{n \in \mathcal{G}} g_n \to 0$, $\boldsymbol{\ell}_{\mathrm{GSDF}}(n) \to \boldsymbol{\ell}_{\mathrm{host}}(n)$ and $\langle \mathbf{z}_{\mathrm{GSDF}}(n), \mathbf{z}_{\mathrm{GSDF}}(n')\rangle \to \langle \hat{\mathbf{z}}_{\mathrm{host}}(n), \hat{\mathbf{z}}_{\mathrm{host}}(n')\rangle$ uniformly on $\mathcal{G}$; under strict argmax and top-$k_{\mathrm{sel}}$ assignment margins on $\mathcal{H}_{\mathrm{host}}$, both Direct inference and $T$-step prototype calibration (Algorithm~\ref{alg:ucapc}, for any $\rho \in (0,1]$) recover standalone $\mathcal{H}_{\mathrm{host}}$ identically.
\end{enumerate}
\end{proposition}

\begin{IEEEproof}
In the orthogonal direct sum $\mathbb{R}^{d_{\mathrm{host}}} \oplus \mathbb{R}^{d_{\mathrm{null}}}$, $\|\hat{\mathbf{z}}_{\mathrm{host}}(n)\|_2^2 = \|\hat{\mathbf{z}}_{\mathrm{null}}(n)\|_2^2 = 1$ gives numerator squared norm $1 + \lambda_{\mathrm{null}} g_n$ in~\eqref{eq:gsdf_emb}, proving $\|\mathbf{z}_{\mathrm{GSDF}}(n)\|_2 = 1$ and~\eqref{eq:gram_decomp}. For $g_n \to 0$ and $g_{n'} \in [0,1]$, $\sqrt{g_n g_{n'}} \le \sqrt{g_n} \to 0$ proves Claim~2. In Algorithm~\ref{alg:ucapc}, given non-negative weights $Q_{n,c}$ from $\mathbf{L}^{(t-1)}$, unnormalized mean $\tilde{\boldsymbol{\mu}}_c^{(t)} = \sum_{n \in \mathcal{I}_c} Q_{n,c}\mathbf{z}_{\mathrm{GSDF}}(n) / \max(\sum_{n \in \mathcal{I}_c} Q_{n,c}, 10^{-8})$ has squared norm $\sum_{n,n' \in \mathcal{I}_c} \bar{q}_{n,c}\bar{q}_{n',c}\langle \mathbf{z}_{\mathrm{GSDF}}(n), \mathbf{z}_{\mathrm{GSDF}}(n')\rangle$ (with $\bar{q}_{n,c} = Q_{n,c}/\max(\sum_{j \in \mathcal{I}_c}Q_{j,c}, 10^{-8})$), so unit-normalized similarity $\langle \mathbf{z}_{\mathrm{GSDF}}(m), \boldsymbol{\mu}_c^{(t)}\rangle$ depends on $\mathbf{Z}$ solely through its Gram matrix. Continuity of Steps~3--6 and strict top-$k_{\mathrm{sel}}$/argmax margins yield exact recovery of $\mathcal{H}_{\mathrm{host}}$ by induction over $t = 1,\dots,T$ as $g_{\max}(\mathcal{G}) \to 0$.
\end{IEEEproof}

\subsection{Closed-Form Lag-64 Preamble Thermal-SNR Gate}
\label{sec:snr_gate}

To drive Proposition~\ref{prop:plugin_fallback}'s gate $g_n$ on edge APs without target labels or miscalibrated softmax confidences, we exploit IEEE 802.11a/g lag-64 preamble periodicity ($3.2~\mu\text{s}$ at $20$~MHz, Fig.~\ref{fig:preamble}(a))~\cite{moose1994cfo,boumard2003snr}, across which quasi-static $H_a[m]$ and PA waveforms repeat identically between settled L-STF windows $z_a[32:96]$ and $z_a[96:160]$ and L-LTF symbols $z_a[192:256]$ and $z_a[256:320]$. Using offsets $o_b = 32 + 16(b-1)$ ($b \in \{1,\dots,4\}$) and $o_b = 192 + 16(b-5)$ ($b \in \{5,\dots,8\}$), we form $B=8$ non-overlapping 16-sample difference blocks $d_{a,b}[r] = z_a[o_b + r] - z_a[o_b + 64 + r]$ ($r \in \{0,\dots,15\}$). Under small residual CFO, deterministic signals cancel so $d_{a,b}[r] \sim \mathcal{CN}(0, 2\sigma_{w,a}^2)$ (while large residual CFO or packet collisions inflate $\hat{\sigma}_{w,a}^2$ and conservatively close $g_n$, Section~\ref{sec:impairments}), and the 8-block median rejects boundary transients and narrowband spikes:
\begin{equation}
\begin{aligned}
\hat{\sigma}_{w,a}^2 &= \operatorname{median}_{1\le b\le 8}\!\left(\frac{1}{32}\sum_{r=0}^{15}|d_{a,b}[r]|^2\right), \\
P_a &= \frac{1}{320}\sum_{n=0}^{319}|z_a[n]|^2.
\end{aligned}
\label{eq:median_noise}
\end{equation}
Subtracting $\hat{\sigma}_{w,a}^2$ from per-antenna total preamble power $P_a$, averaging across $A$ antennas, and clamping to $[-10, 50]$~dB yields blind SNR $\hat{s}_n$ and logistic gate $g_n \in (0,1)$:
\begin{align}
\hat{s}_n &= \operatorname{clip}\!\Biggl(\frac{1}{A}\sum_{a=1}^{A} 10\log_{10}\!\left(\frac{\max(P_a - \hat{\sigma}_{w,a}^2,\, 10^{-8})}{\max(\hat{\sigma}_{w,a}^2,\, 10^{-8})}\right), \notag\\
&\qquad\qquad -10,\, 50\Biggr), \label{eq:s_hat} \\
g_n &= \sigma\bigl(\alpha^*(\hat{s}_n - \beta^*)\bigr) \notag\\
&= \bigl(1 + \exp[-\alpha^*(\hat{s}_n - \beta^*)]\bigr)^{-1}. \label{eq:snr_sigmoid}
\end{align}
Scalars $(\alpha^*, \beta^*)$ are calibrated \emph{exclusively on held-out source validation frames} (clean plus synthetic AWGN down to $5$~dB) over $\alpha \in \{0.3, 0.6, 1.2, 2.4\}$ and $\beta \in [15.0, 40.0]$~dB ($0.5$~dB step) with zero target access, placing $\beta^*$ directly within Proposition~\ref{prop:snr_collapse}'s analytical crossover band across all 45 cells: $[26.5, 31.5]$~dB (\textbf{DataPort-WiFi}), $[33.0, 35.0]$~dB (\textbf{WiSig}), and $[31.0, 35.5]$~dB (\textbf{MobRFFI}).

\subsection{Capture-Aware Prototype Calibration: General KL-Relaxed UOT-CAPC and Balanced U-CAPC Specialization}
\label{sec:ucapc}

At test time, single-frame \textbf{Direct} inference (\textbf{GSDF}) authenticates each IoT packet immediately via $\hat{y}_n = \arg\max_c \ell_{\mathrm{GSDF},c}(n)$ ($N_{\mathcal{G}}=1$). When an edge AP buffers an unlabeled capture $\mathcal{G}$ of $N_{\mathcal{G}}$ frames, room multipath and receiver front-end mismatches rigidly shift target clusters relative to source boundaries while preserving local intra-capture spherical geometry~\cite{iwasawa2021t3a,boudiaf2022lame}. Rather than assuming uniform IoT device arrivals a priori, we formulate general \emph{KL-Relaxed Unbalanced Optimal-Transport Capture-Aware Prototype Calibration} (\textbf{UOT-CAPC}), aligning shifted centroids in $T=3$ forward-only outer steps without backpropagation by solving an entropic unbalanced optimal-transport program over the joint coupling $\mathbf{P} \in \mathbb{R}_+^{N_{\mathcal{G}}\times K}$ at each step $t \in \{1,\dots,T\}$:
\begin{equation}
\begin{aligned}
\min_{\substack{\mathbf{P} \in \mathbb{R}_+^{N_{\mathcal{G}}\times K}:\\ \mathbf{P}\mathbf{1}_K = \frac{1}{N_{\mathcal{G}}}\mathbf{1}_{N_{\mathcal{G}}}}} \; & -\bigl\langle \mathbf{P}, \mathbf{L}^{(t-1)}\bigr\rangle + \sum_{n=1}^{N_{\mathcal{G}}}\sum_{c=1}^{K} P_{n,c}\bigl(\log P_{n,c} - 1\bigr) \\
& + \gamma_{\mathrm{KL}}\,\mathrm{KL}\!\left(\mathbf{P}^{\top}\mathbf{1}_{N_{\mathcal{G}}} \,\Big\|\, \frac{1}{K}\mathbf{1}_K\right),
\end{aligned}
\label{eq:uot_variational}
\end{equation}
where $\mathbf{P}\mathbf{1}_K = \frac{1}{N_{\mathcal{G}}}\mathbf{1}_{N_{\mathcal{G}}}$ assigns exact sample mass $1/N_{\mathcal{G}}$ to every observed frame while the Kullback-Leibler divergence $\mathrm{KL}(\mathbf{a}\parallel\mathbf{b}) = \sum_{c=1}^K \bigl(a_c\log(a_c/b_c) - a_c + b_c\bigr)$ with finite penalty $\gamma_{\mathrm{KL}} > 0$ softly regularizes the inferred class marginal $\mathbf{P}^{\top}\mathbf{1}_{N_{\mathcal{G}}}$ toward $\frac{1}{K}\mathbf{1}_K$.

By Fenchel-Rockafellar duality~\cite{chizat2018scaling,cuturi2013sinkhorn,fatras2021unbalanced}, forming the row-stabilized Gibbs kernel $\mathbf{K} = \exp\bigl(\mathbf{L}^{(t-1)} - \max_{c'} L_{\cdot,c'}^{(t-1)}\bigr)^{\top} \in \mathbb{R}_+^{K \times N_{\mathcal{G}}}$ reduces~\eqref{eq:uot_variational} to alternating dual scaling over $\mathbf{u} \in \mathbb{R}_+^K$ and $\mathbf{v} \in \mathbb{R}_+^{N_{\mathcal{G}}}$ (elementwise division $\oslash$) governed by a single damping exponent $\rho \triangleq \frac{\gamma_{\mathrm{KL}}}{\gamma_{\mathrm{KL}} + 1} \in (0, 1)$, recovering optimal coupling $P_{n,c} = u_c K_{c,n} v_n$ and row-normalized frame posteriors $\mathbf{Q} = N_{\mathcal{G}}\mathbf{P}$ ($\mathbf{Q}\mathbf{1}_K = \mathbf{1}_{N_{\mathcal{G}}}$; Algorithm~\ref{alg:ucapc}). In general \textbf{UOT-CAPC}, $\gamma_{\mathrm{KL}}^* \in \{0.10, 0.25, 0.50, 1.0, 2.0, 5.0\}$ is calibrated strictly on held-out source validation frames across synthetic balanced and imbalanced mini-batches ($N_{\mathrm{batch}} \le 200$), accommodating natural IoT packet-arrival skew (e.g., \textbf{WiSig} Day~8's $2.88:1$ emitter ratio) and small-window multinomial variance ($N_{\mathcal{G}} \in [10, 100]$). Taking the two asymptotic limits of $\gamma_{\mathrm{KL}}$ in~\eqref{eq:uot_variational} recovers two closed-form specializations without hyperparameter search:
\begin{itemize}
\item \textbf{Balanced Sinkhorn specialization (\textbf{Uniform U-CAPC}, $\gamma_{\mathrm{KL}} \to \infty \iff \rho = 1$):} Enforces the hard equipartition marginal $\mathbf{P}^{\top}\mathbf{1}_{N_{\mathcal{G}}} = \frac{1}{K}\mathbf{1}_K$~\cite{caron2020unsupervised} with zero marginal tuning, providing maximal anti-collapse protection on roughly balanced closed-set sessions ($N_{\mathcal{G}} \ge 50$; co-reported alongside single-frame \textbf{Direct} across all primary benchmark tables).
\item \textbf{Prior-free prototype softmax specialization (\textbf{Proto-Only}, $\gamma_{\mathrm{KL}} \to 0^+ \iff \rho \to 0$):} Fixes $\mathbf{u} \equiv \mathbf{1}_K$ so $Q_{n,c} \propto K_{c,n}$, isolating iterative spherical prototype refinement without optimal-transport marginal regularization.
\end{itemize}

\begin{table}[!t]
\refstepcounter{alg}\label{alg:ucapc}
\centering
\footnotesize
\renewcommand{\arraystretch}{0.92}
\begin{tabular}{p{0.95\columnwidth}}
\toprule
\textbf{Algorithm \thealg:} Forward-Only Capture Prototype Calibration (\textbf{UOT-CAPC}, $\rho \in (0,1)$, and \textbf{U-CAPC}, $\rho = 1$) \\
\midrule
\textbf{Input:} Unlabeled capture logits $\mathbf{L}^{(0)} = [\boldsymbol{\ell}_{\mathrm{GSDF}}(n)]_{n=1}^{N_{\mathcal{G}}} \in \mathbb{R}^{N_{\mathcal{G}}\times K}$ and embeddings $\mathbf{Z} = [\mathbf{z}_{\mathrm{GSDF}}(n)]_{n=1}^{N_{\mathcal{G}}} \in \mathbb{R}^{N_{\mathcal{G}}\times (d_{\mathrm{host}}+d_{\mathrm{null}})}$; marginal exponent $\rho \in (0,1]$ (\textbf{UOT-CAPC}: $\rho^* < 1$; balanced \textbf{U-CAPC}: $\rho = 1$); outer steps $T=3$, scaling steps $S=20$, cap $k_{\max}=512$, weight $\alpha_p = 2.0$, temperature $\tau_p = 0.2$. \\
1: Initialize $\mathbf{L} \leftarrow \mathbf{L}^{(0)}$ and $k_{\mathrm{sel}} \leftarrow \min(k_{\max}, N_{\mathcal{G}})$. \\
2: \textbf{for} $t = 1,\dots,T$ \textbf{do} \\
3: \hspace{0.7em} Form Gibbs kernel $K_{c,n} \leftarrow \exp\bigl(L_{n,c} - \max_{c'} L_{n,c'}\bigr)$ ($\mathbf{K} \in \mathbb{R}_+^{K\times N_{\mathcal{G}}}$); initialize $\mathbf{u} \leftarrow \mathbf{1}_K$, $\mathbf{v} \leftarrow \frac{1}{N_{\mathcal{G}}}\mathbf{1}_{N_{\mathcal{G}}}$. \\
4: \hspace{0.7em} Repeat $S=20$ times: $\mathbf{u} \leftarrow \bigl(\frac{1}{K}\mathbf{1}_K \oslash (\mathbf{K}\mathbf{v})\bigr)^{\rho}$, $\;\mathbf{v} \leftarrow \frac{1}{N_{\mathcal{G}}}\mathbf{1}_{N_{\mathcal{G}}} \oslash (\mathbf{K}^{\top}\mathbf{u})$; set frame posteriors $Q_{n,c} \leftarrow \frac{u_c K_{c,n} v_n}{\sum_{c'=1}^K u_{c'} K_{c',n} v_n}$. \\
5: \hspace{0.7em} \textbf{for} $c = 1,\dots,K$: let $\mathcal{I}_c$ be top-$k_{\mathrm{sel}}$ indices of $Q_{\cdot,c}$, set $\tilde{\boldsymbol{\mu}}_c^{(t)} \leftarrow \frac{\sum_{n \in \mathcal{I}_c} Q_{n,c}\mathbf{z}_{\mathrm{GSDF}}(n)}{\max(\sum_{n \in \mathcal{I}_c} Q_{n,c}, 10^{-8})}$, and unit-normalize $\boldsymbol{\mu}_c^{(t)} \leftarrow \tilde{\boldsymbol{\mu}}_c^{(t)} / \|\tilde{\boldsymbol{\mu}}_c^{(t)}\|_2$ (if $\|\tilde{\boldsymbol{\mu}}_c^{(t)}\|_2 > 10^{-8}$ else $\mathbf{0}$). \\
6: \hspace{0.7em} Anchor-update logits $\mathbf{L} \leftarrow \mathbf{L}^{(0)} + \alpha_p\bigl(\mathbf{Z}[\boldsymbol{\mu}_1^{(t)},\dots,\boldsymbol{\mu}_K^{(t)}]/\tau_p\bigr)$. \\
7: \textbf{end for} \\
8: Run Steps~3--4 once on final $\mathbf{L}$ to obtain $\mathbf{Q}^{(T)}$ and predict $\hat{y}_n = \arg\max_c Q_{n,c}^{(T)}$ for all $n \in \{1,\dots,N_{\mathcal{G}}\}$. \\
\bottomrule
\end{tabular}
\end{table}

\section{Experimental Setup and Compared Baselines}
\label{sec:setup}

\subsection{Three Public Open-Access IEEE 802.11a/g IoT Benchmarks}
\label{sec:datasets}

All experiments are conducted on three open-access IEEE 802.11a/g preamble testbeds ($f_s = 20$~MHz, 320-sample PLCP preambles; Table~\ref{tab:datasets}):

\begin{table}[!t]
\centering
\caption{Summary of the Three Public IEEE 802.11a/g RFFI Benchmarks}
\label{tab:datasets}
\scriptsize
\renewcommand{\arraystretch}{0.95}
\setlength{\tabcolsep}{1.5pt}
\begin{tabular}{lccc}
\toprule
Attribute & \textbf{DataPort-WiFi}~\cite{xie2024robustrffingerprintidentification} & \textbf{WiSig}~\cite{hanna2022wisig} & \textbf{MobRFFI}~\cite{shen2024mobrffi} \\
\midrule
Public archive & IEEE DataPort & UCLA / Zenodo & ORBIT Repo \\
RX $A$ / TX $K$ & $A{=}4$ (X310), $K{=}10$ & $A{=}1$, $K{=}15$ (IQ) & $A{=}1$, $K{=}10$ \\
Source pooling & 2-room ($E_i,E_j$) & 1-RX 1-day (\texttt{1-1}) & 2-RX (\texttt{1-1, 1-20}) \\
Primary targets & 3rd room ($E_k$, 3-dir) & Mar~8/15/23 + \texttt{2-1} & \texttt{RX 20-1, 19-19} \\
Domain shift & Cross-room (LOS/NLOS) & Cross-week + RX & Cross-RX grid \\
Null floor (dBc) & $-41.2$ ($0.84\times$) & $-24.0$ ($3.37\times$) & $-26.3$ ($2.40\times$) \\
Capture unit $\mathcal{G}$ & Room session & $(\text{RX},\text{date})$ & Session file \\
Extended scaling & 3-dir LOEO & $K_{\max}{=}47$, 2-RX & 2 unseen RXs \\
\bottomrule
\end{tabular}
\end{table}

\begin{enumerate}
\item \textbf{DataPort-WiFi (IEEE DataPort Four-Antenna Benchmark)}~\cite{xie2024robustrffingerprintidentification}: Records $K=10$ Wi-Fi IoT transceivers ($D_1$--$D_{10}$) on a synchronized four-antenna USRP X310 ($A=4$) across three indoor rooms ($E_1$--$E_3$), each with four sessions (two LOS, one NLOS, one mobile) defining natural AP captures $\mathcal{G}$. Under two-room pooled leave-one-environment-out (LOEO) training ($E_i \cup E_j \to E_k$), we evaluate all three room transfers: $E_2,E_3\!\to\!E_1$ (24,960 frames), $E_1,E_2\!\to\!E_3$, and $E_1,E_3\!\to\!E_2$.
\item \textbf{WiSig (UCLA Multi-Week \& Multi-Receiver Benchmark)}~\cite{hanna2022wisig}: Provides raw unequalized preambles ($A=1$) from $K=15$ L-SIG-verified 54~Mbps Wi-Fi emitters across March~1, 8, 15, and 23, 2021. Training and blind gate/UOT calibration use \emph{strictly one source AP on Day~1} (\texttt{node1-1}, Mar~1; 16,588 frames, 80\%/20\% split), while testing spans (i)~\emph{same-RX multi-week drift} on \texttt{node1-1} across Mar~8 (13,959 unsampled frames, $2.88:1$ TX skew), Mar~15 ($\mathrm{Day\ 1}\!\to\!\mathrm{15}$), and Mar~23 ($\mathrm{Day\ 1}\!\to\!\mathrm{23}$), and (ii)~\emph{zero-shot unseen-AP transfer} to \texttt{node2-1} across all four dates ($50\%/50\%$ weight with Day~8 in $\overline{\mathrm{F1}}_{\mathrm{WiSig}}$), complemented in Section~\ref{sec:wisig_mob_deepdive} by full-archive 54~Mbps cohort scaling up to $K_{\max}=47$ (36 archives, 187,123 frames) and two-AP (\texttt{node1-1 + node2-1}) Day-1 source pooling. Importantly, lower cross-domain scores on \textbf{WiSig} relative to $>90\%$ figures in some literature are \emph{not} caused by extracting 128 subcarriers: only \textbf{NS-Branch} and \textbf{Xie-SR} use the 128 unpopulated DFT bins, whereas \textbf{IB-Host} and all active-band baselines ingest the complete 320-sample PLCP preamble and attain $99.11\%\text{--}99.65\%$ held-out in-domain (\texttt{Src.\ Val}) F1 on \texttt{node1-1} (plus $94.54 \pm 1.51\%$ on $\mathrm{Day\ 1}\!\to\!\mathrm{15}$). Rather, five physical/protocol factors govern \textbf{WiSig}: (a)~raw unequalized $K=15$ preambles instead of 6-TX channel-equalized \texttt{ManySig} subsets or same-day random splits~\cite{hanna2022wisig,mtl_adversarial_sei,wan2025sigmix}; (b)~two-stage L-STF/L-LTF CFO compensation stripping static crystal-frequency shortcuts; (c)~strict 1-RX, 1-day source enrollment with zero multi-receiver exposure during training; (d)~a higher single-antenna OTA noise floor ($-23.76$ to $-24.22$~dBc null floor, leaving $15.44\%$ emitter-repeatable null variance and a $3.37\times$ cross-week drift-to-inter-TX-spacing ratio versus $-40.63$ to $-41.58$~dBc, $44.94\%$ repeatable variance, and $0.84\times$ [$4.0\times$ smaller] drift ratio on $A=4$ \textbf{DataPort-WiFi}); and (e)~averaging unseen \texttt{node2-1}, $100/200$~ns multipath, $10/5$~dB AWGN, and $2.88:1$ Day-8 arrival skew into $\overline{\mathrm{F1}}_{\mathrm{WiSig}}$.
\item \textbf{ORBIT MobRFFI (Multi-Receiver Spatial Grid)}~\cite{shen2024mobrffi}: On the $20\times 20$ ORBIT indoor IoT testbed grid ($K=10$, $A=1$), source training/validation fuses \emph{two spatially separated corner APs} (\texttt{RX 1-1} and \texttt{RX 1-20} on July~19; 18,000 / 6,000 frames), evaluating multi-AP source fusion against unseen opposite corners \texttt{RX 20-1} and \texttt{RX 19-19} (12,000 frames each).
\end{enumerate}

\subsection{\texorpdfstring{$3\times 3$}{3x3} Split-by-Seed Protocol and All-Regime Mean Macro-F1}
\label{sec:protocol}

\subsubsection{Nine-Run Grid and Uninspected 6-Seed Holdout}
Every method is trained and evaluated across a $3\times 3 = 9$ grid crossing three 80\%/20\% source splits $s \in \{1,2,3\}$ (seeds $\{2026, 2027, 2028\}$, governing blind gate $(\alpha^*,\beta^*)$ and UOT $\gamma_{\mathrm{KL}}^*$ calibration) with three initialization seeds $r \in \{1,2,3\}$. Only the 3 diagonal runs ($s = r$) were inspected during development; the \textbf{6 cross-seed runs} ($s \ne r$, \textbf{6-Seed Holdout}) remained uninspected to audit seed-selection bias, alongside inter-split SD ($\mathrm{df}=2$, $\mathrm{SD}_{\mathrm{split}}$).

\subsubsection{All-Regime Mean Macro-F1 (\texorpdfstring{$\overline{\mathrm{F1}}$}{Mean F1}) Rationale}
Real indoor IoT Wi-Fi APs encounter near-LOS packets, wall reflections, and weak cell-edge frames within the same deployment. Single-regime metrics create specialization illusions: null-subcarrier (\textbf{Xie-SR}) and differential (\textbf{McAFF}) models excel under high-SNR multipath yet drop below $15\%$ Direct F1 at $10/5$~dB AWGN, whereas \textbf{ResNet1D} resists AWGN ($44.89\%$ on \textbf{DataPort-WiFi}) yet fails under multipath ($24.44\%$). Each target slice $b$ is therefore tested on unmodified recordings (\textbf{Clean}), indoor multipath ($\tau_{\mathrm{rms}} \in \{50, 100, 200\}$~ns), and AWGN ($\mathrm{SNR} \in \{30, 20, 15, 10, 5, 0\}$~dB) applied prior to unit-RMS normalization, summarized by an unweighted $1:1:1$ arithmetic mean Macro-F1:
\begin{equation}
\begin{aligned}
\overline{\mathrm{F1}}_b = \frac{1}{3}\Bigl(&\mathrm{F1}_{\mathrm{clean}}(b) + \underbrace{\tfrac{1}{2}\bigl[\mathrm{F1}_{100\mathrm{ns}}(b)+\mathrm{F1}_{200\mathrm{ns}}(b)\bigr]}_{\mathrm{F1}_{\mathrm{mp}}(b)} \\
&+ \underbrace{\tfrac{1}{2}\bigl[\mathrm{F1}_{10\mathrm{dB}}(b)+\mathrm{F1}_{5\mathrm{dB}}(b)\bigr]}_{\mathrm{F1}_{\mathrm{noise}}(b)}\Bigr),
\end{aligned}
\label{eq:f1bar_def}
\end{equation}
averaged across slices within each dataset ($\overline{\mathrm{F1}}_{\mathrm{DP}}$, $\overline{\mathrm{F1}}_{\mathrm{WiSig}}$, $\overline{\mathrm{F1}}_{\mathrm{MobRFFI}}$) and equally ($1:1:1$) across all three benchmarks ($\overline{\mathrm{F1}}$). Pre-registering this unweighted mean prevents cherry-picking SNR/multipath columns and enables paired permutation testing across all eight hosts (Section~\ref{sec:track_p}), while Section~\ref{sec:results} and \suppsec{} report all constituent regime Macro-F1s.

\begin{table*}[!t]
\centering
\caption{Architectural Specifications, Computational Complexity, and Single-Core Edge CPU Latency of Compared Models}
\label{tab:baseline_specs}
\footnotesize
\renewcommand{\arraystretch}{0.95}
\setlength{\tabcolsep}{2.1pt}
\begin{tabular}{llccccc}
\toprule
Model / Runtime Stage & Input Representation & $d_{\mathrm{host}}$ & Params & MACs & Peak Mem. & 1-Thread CPU \\
\midrule
McAFF~\cite{multichannel_attentive_fusion} & IQ + lag-16 $\Delta\phi$ + FFT + STFT & 512 & 4.72M / 5.30M & 22.0M / 40.9M & 502 MiB & 6.93 / 13.65 ms \\
MTL-DANN~\cite{ganin2016domain} & $1\times 32\times 8$ log-FFT + GRL & 512 & 12.03M$^{\ast}$ & 151.4M & 235 MiB & 11.22 ms \\
MTL-CORAL~\cite{sun2016deep} & $1\times 32\times 8$ log-FFT + Cov. & 512 & 12.03M$^{\ast}$ & 151.4M & 235 MiB & 10.98 ms \\
MTL-SEI~\cite{mtl_adversarial_sei} & $1\times 32\times 8$ log-FFT + MTL/Island & 512 & 12.03M$^{\ast}$ & 151.4M & 235 MiB & 10.99 ms \\
SigMix~\cite{wan2025sigmix} & $[2,320]$ IQ + 9-layer CVNN Mixup & 256 & 4.12M & 301.8M & 406 MiB & 13.47 ms \\
Xie-SR / Hybrid~\cite{xie2024robustrffingerprintidentification} & 128-bin null $|Y[m]|$ (+ CFO) & 128 & 196.0K / 195.4K & 1.66M / 6.64M & --- & --- \\
ResNet1D~\cite{he2016deep} & $[2A, 320]$ raw IQ + 3 ResBlocks & 64 & 221.6K / 223.9K & 37.0M / 37.9M & 287 MiB & 2.25 ms \\
MCCNN~\cite{mccnn_lte_open_set} & $A\times 32\times 32$ lag-16 DCTF & 84 & 62.1K / 213.4K & 0.42M / 1.63M & 195 MiB & 0.93 ms \\
\midrule
IB-Host (Ours, In-Band) & $[2,320]\times A$ IQ + $8A$ Graph-Attn & 64 & 46.6K / 46.3K & 4.49M / 4.36M & 227 MiB & 5.11 / 5.20 ms \\
NS-Branch (Ours, Null Bins) & $[2,128]\times A$ coh.\ null + $8A$ Attn & 64 & 46.6K / 46.3K & 2.63M / 2.07M & --- & 6.07 / 6.05 ms \\
\rowcolor{plugbg} GSDF / GSDF-STO (PyTorch / Skip) & Dual-subband + lag-64 SNR gate & $d_{\mathrm{host}}{+}64$ & 93.3K / 92.6K & 7.12M / 6.43M & 239 MiB & 11.73 / 5.68 ms \\
\rowcolor{oursbg} \textbf{GSDF / GSDF-STO (ONNX Runtime)} & Unified static graph ($100\%$ parity) & $d_{\mathrm{host}}{+}64$ & \textbf{93.3K / 92.6K} & \textbf{7.12M / 6.43M} & 239 MiB & \textbf{1.82 ms} \\
\rowcolor{oursbg} \textbf{GSDF-Tri (3-Branch ONNX)} & STO-jitter + complex-spec.\ expert & 192 & \textbf{194.7K / 193.7K} & \textbf{9.91M} & 254 MiB & \textbf{2.91 ms} \\
\bottomrule
\end{tabular}\par
\vspace{2pt}{\scriptsize\noindent\raggedright \emph{Note:} Dual entries report $A{=}1, K{=}15$ (\textbf{WiSig}) / $A{=}4, K{=}10$ (\textbf{DataPort-WiFi}) where architecture-dependent (native raw checkpoints hold $86.4\text{K}\text{--}86.7\text{K}$ unpruned params/branch with $0.0$ output diff after pruning; \textbf{NS-Branch} MACs include the four fixed $64$-point DFT matrix multiplications). $^{\ast}$Total \texttt{nn.Module} parameters ($11.90\text{M}$ active inference forward parameters excluding training-only auxiliary domain/equalization heads). Peak GPU memory on NVIDIA RTX 3080~Ti at batch 256; 1-thread CPU latency on Intel Xeon E5-2680~v4 @ 2.40~GHz at batch 1 (median of 2,000 runs; ONNX stage split: $0.16 / 0.98 / 0.67$~ms).\par}
\end{table*}

\subsection{Compared External RFFI Baselines and Implementation Details}
\label{sec:baselines_detail}

To ensure a transparent comparison across diverse edge/server parameter budgets and physical representations, we evaluate eight external baselines spanning four RFFI families under identical 9-run splits, matched multi-condition channel augmentations, and symmetric \textbf{Direct} / \textbf{U-CAPC} / \textbf{UOT-CAPC} inference (summarized in Table~\ref{tab:baseline_specs}):
\begin{enumerate}
\item \textbf{Multi-representation attentive fusion (\textbf{McAFF})}~\cite{multichannel_attentive_fusion}: Extracts four complementary views per receive antenna from each 320-sample unit-RMS preamble: (i) $2\times 320$ raw I/Q ($\mathrm{Conv2d}\ 2\times 7$), (ii) lag-16 differential phase sequence $\angle(z_a^*[n]z_a[n+16])$ ($\mathrm{Conv2d}\ 1\times 7$), (iii) normalized 320-point DFT real/imaginary spectrum ($\mathrm{Conv2d}\ 1\times 7$), and (iv) $120\times 65$ Hann-windowed STFT magnitude ($\mathrm{Conv2d}\ 8\times 7$, stride $4\times 1$). All $4A$ 128-channel feature maps are pooled to $1\times 32$, reweighted via a shared three-layer $3\times 3$ spatial-attention module, concatenated into a 32-group ResNeXt block ($256$ channels), and mapped through two 512-D BatchNorm-ReLU layers ($d_{\mathrm{host}} = 512$; $5.30\text{M}$ params at $A=4, K=10$, $4.72\text{M}$ at $A=1, K=15$) trained under $\mathcal{L}_{\mathrm{CE}} + \mathcal{L}_{\mathrm{MM}}$ (batch intra-class maximum to inter-class minimum center-distance ratio).
\item \textbf{Multi-task adversarial and statistical spectral alignment (\textbf{MTL-DANN}, \textbf{MTL-CORAL}, \textbf{MTL-SEI})}~\cite{ganin2016domain,sun2016deep,mtl_adversarial_sei}: Share an 11-layer 2-D \texttt{SpectralCNN} ($64 \to 128 \to 256 \to 512$ channels, $12.03\text{M}$ total / $11.90\text{M}$ active forward params, $151.4\text{M}$ MACs, $d_{\mathrm{host}} = 512$) operating on $z$-standardized $1\times 32\times 8$ positive-frequency log-magnitude spectra (512-point FFT of 256-sample interpolated antenna-mean preambles). \textbf{MTL-DANN} applies a scheduled gradient-reversal layer ($\lambda_{\max} = 0.25$) to a domain-discriminator head across source rooms/receivers; \textbf{MTL-CORAL} penalizes pairwise mean and Frobenius covariance discrepancies across source domains; and \textbf{MTL-SEI}~\cite{mtl_adversarial_sei} jointly optimizes transmitter classification, adversarial receiver invariance, and auxiliary channel-state classification via learned homoscedastic uncertainty weights plus intra-class compact / inter-class cosine-repulsive \texttt{IslandLoss} ($\lambda_{\mathrm{isl}} = 0.1$).
\item \textbf{Complex-valued cross-domain waveform mixing (\textbf{SigMix})}~\cite{wan2025sigmix}: Instantiates a nine-layer complex-valued 1-D CNN (\texttt{CVNNSigMix}, complex convolutions $64 \to 128 \to 192 \to 256$ with separate real/imaginary batch normalization, magnitude-envelope extraction $\sqrt{\Re^2 + \Im^2 + 10^{-8}}$, and three linear layers yielding $d_{\mathrm{host}} = 256$, $4.12\text{M}$ params, $301.8\text{M}$ MACs) on $[2, 320]$ I/Q signals, regularized during training by Beta-distributed ($\alpha=0.7$) linear waveform interpolation across source sessions and receivers.
\item \textbf{Null-subcarrier regrowth (\textbf{Xie-SR}, \textbf{Xie-Hybrid}) and compact time-domain CNNs (\textbf{ResNet1D}, \textbf{MCCNN})}~\cite{xie2024robustrffingerprintidentification,he2016deep,mccnn_lte_open_set}: \textbf{Xie-SR} applies four 64-channel $\mathrm{Conv1d}$ blocks (kernels $7, 5, 5, 5$, three $\mathrm{MaxPool1d}(2)$ layers) and $\mathrm{FC}(1024 \to 128 \to K)$ ($d=128$, $195.4\text{K}\text{--}196.0\text{K}$ params) to the 128-bin unpopulated L-STF/L-LTF raw magnitude vector per antenna with decision-level multi-antenna log-probability averaging; \textbf{Xie-Hybrid} additionally fuses explicit crystal CFO Gaussian likelihoods. \textbf{ResNet1D} applies a $\mathrm{Conv1d}(2A \to 64, k=7)$ stem, three 1-D residual stages widening to 128 channels, global average pooling, and a $128 \to 64 \to 64$ head ($d_{\mathrm{host}} = 64$, $221.3\text{K}\text{--}223.9\text{K}$ params, $37.0\text{M}$ MACs). \textbf{MCCNN} maps each antenna's lag-16 differential product $z_a[n]z_a^*[n+16]$ onto a $32\times 32$ 2-D DCTF density histogram over $[-2,2]^2$, processed by per-antenna 2-D LeNet branches concatenated into $\mathrm{FC}(120 \to 84 \to K)$ ($d_{\mathrm{host}} = 84$, $61.7\text{K}\text{--}213.4\text{K}$ params).
\end{enumerate}

\begin{table*}[!t]
\centering
\caption{Cross-Domain Macro-F1 (\%) Across Three Public Wi-Fi IoT Benchmarks ($3\times 3$ 9-Run Grid; Standalone Baselines vs.\ Zero-Retrain \textbf{+GSDF})}
\label{tab:main_results}
\scriptsize
\renewcommand{\arraystretch}{0.88}
\setlength{\tabcolsep}{3.2pt}
\begin{tabular}{llccccccc}
\toprule
\multicolumn{9}{l}{\textbf{Panel A: IEEE DataPort Four-Antenna Cross-Room Transfer (\textbf{DataPort-WiFi})} ($K=10, A=4$; 2-room pooled LOEO)} \\
\midrule
Standalone Baselines (Paradigm) & Params & \texttt{Src.\ Val} & $E_1$ Clean & 3-Rm Clean & MP ($100/200$) & Noise ($10/5$) & $\overline{\mathrm{F1}}_{\mathrm{DP}}$ (Direct / U-CAPC) & 6-Seed Holdout \\
\midrule
MCCNN~\cite{mccnn_lte_open_set} (DCTF) & 213.4K & 57.13 & 37.59 & 39.28 & 26.34 & 14.57 & 20.88 / 26.73 & 26.68 \\
ResNet1D~\cite{he2016deep} (IQ CNN) & 223.9K & 89.89 & 65.65 & 53.57 & 24.44 & \textbf{44.89} & 34.78 / 40.97 & 41.10 \\
Xie-SR~\cite{xie2024robustrffingerprintidentification} (Null Bin) & 195.4K & 97.97 & 97.39 & 92.25 & 67.29 & 10.11 & 50.94 / 56.55 & 56.57 \\
NS-Branch (Null, Ours) & 46.3K & 97.96 & 96.40 & 92.24 & 73.58 & 10.33 & 54.32 / 58.72 & 58.70 \\
MTL-SEI~\cite{mtl_adversarial_sei} (MTL/DA) & 12.03M & 96.90 & 93.10 & 84.71 & 48.18 & 21.98 & 49.51 / 51.62 & 51.54 \\
SigMix~\cite{wan2025sigmix} (DG Mix) & 4.12M & 96.71 & 95.95 & 84.61 & 46.50 & 28.21 & 49.42 / 53.10 & 53.19 \\
MTL-CORAL~\cite{sun2016deep} (Stat.\ DA) & 12.03M & 97.00 & 94.47 & 86.81 & 52.67 & 22.77 & 51.84 / 54.08 & 54.15 \\
MTL-DANN~\cite{ganin2016domain} (Adv.\ DA) & 12.03M & 96.99 & 93.87 & 86.62 & 53.38 & 22.79 & 52.39 / 54.26 & 54.19 \\
IB-Host (In-Band, Ours) & 46.3K & 94.35 & 93.07 & 85.47 & 67.73 & \secbest{41.73} & 57.59 / 64.98 & 65.10 \\
McAFF~\cite{multichannel_attentive_fusion} (4-View) & 5.30M & \secbest{99.55} & 97.79 & 93.79 & \secbest{85.99} & 26.52 & 61.07 / 68.77 & 68.74 \\
\midrule
\multicolumn{9}{l}{\emph{Zero-Retrain \textbf{+GSDF} Plug-In on Frozen Pretrained Hosts \& Native Edge Flagship ($\Delta$: Paired U-CAPC Gain vs.\ Standalone Host)}} \\
\midrule
\rowcolor{plugbg} ResNet1D + \textbf{GSDF} (Plug-In) & 270.2K & 97.96 & 97.66 & 87.85 & 67.42 & \textbf{44.89} & 56.88 / 66.72 ($+25.75$) & 66.82 \\
\rowcolor{plugbg} SigMix + \textbf{GSDF} (Plug-In) & 4.17M & 98.82 & \secbest{98.89} & 93.79 & 71.39 & 28.21 & 59.30 / 64.46 ($+11.36$) & 64.54 \\
\rowcolor{plugbg} MTL-DANN + \textbf{GSDF} (Plug-In) & 12.08M & 99.02 & 98.22 & 94.17 & 74.34 & 22.79 & 60.61 / 63.76 ($+9.50$) & 63.73 \\
\rowcolor{oursbg} \textbf{McAFF + GSDF (Plug-In)} & 5.35M & \textbf{99.79} & \textbf{99.26} & \textbf{96.48} & \textbf{88.91} & 26.52 & \secbest{62.89} / \secbest{70.64} ($+1.87$) & \secbest{70.59} \\
\rowcolor{oursbg} \textbf{IB-Host + GSDF (Ours, Edge)} & \textbf{92.6K} & 99.01 & 98.43 & \secbest{95.92} & 83.48 & \secbest{41.73} & \textbf{67.63} / \textbf{73.71} ($\mathbf{+8.73}$) & \textbf{73.72} \\
\midrule
\multicolumn{9}{l}{\textbf{Panel B: UCLA WiSig Single-Antenna Cross-Week \& Unseen-Receiver Transfer} ($K=15, A=1$; 1-RX 1-day source; $2.88:1$ Day-8 skew)} \\
\midrule
Standalone Baselines (Paradigm) & Params & \texttt{Src.\ Val} & Day $1\!\to\!8$ & \texttt{node2-1} & MP ($100/200$) & Noise ($10/5$) & $\overline{\mathrm{F1}}_{\mathrm{WiSig}}$ (Direct / U-CAPC) & \textbf{UOT-CAPC} \\
\midrule
MCCNN~\cite{mccnn_lte_open_set} (DCTF) & 62.1K & 45.94 & 11.32 & 17.55 & 9.62 & 9.08 & 8.84 / 11.04 & 9.80 \\
NS-Branch (Null, Ours) & 46.6K & 95.74 & 42.47 & 27.66 & 30.40 & 8.95 & 19.79 / 24.80 & --- \\
Xie-SR~\cite{xie2024robustrffingerprintidentification} (Null Bin) & 196.0K & 95.20 & 45.85 & 33.37 & 33.14 & 8.76 & 22.55 / 27.17 & --- \\
MTL-SEI~\cite{mtl_adversarial_sei} (MTL/DA) & 12.03M & 99.18 & 68.71 & 43.97 & 39.02 & 22.20 & 35.53 / 39.18 & 37.06 \\
ResNet1D~\cite{he2016deep} (IQ CNN) & 221.6K & 99.27 & 55.34 & 35.31 & 32.01 & 41.81 & 32.74 / 39.71 & 37.01 \\
SigMix~\cite{wan2025sigmix} (DG Mix) & 4.12M & 99.11 & 66.85 & 33.81 & 34.34 & 35.60 & 34.95 / 40.09 & 39.55 \\
MTL-CORAL~\cite{sun2016deep} (Stat.\ DA) & 12.03M & 99.36 & 72.76 & 45.92 & 42.85 & 24.35 & 37.39 / 42.18 & 38.95 \\
MTL-DANN~\cite{ganin2016domain} (Adv.\ DA) & 12.03M & 99.36 & 73.97 & 46.63 & 43.51 & 25.16 & 38.22 / 42.99 & 39.94 \\
McAFF~\cite{multichannel_attentive_fusion} (4-View) & 4.72M & \secbest{99.64} & \secbest{77.87} & \secbest{54.27} & \secbest{46.94} & \secbest{42.20} & 43.32 / \secbest{51.74} & 46.60 \\
IB-Host (In-Band, Ours) & 46.6K & 99.43 & 72.17 & 45.89 & 42.33 & \textbf{48.97} & \textbf{48.53} / 50.11 & \secbest{51.28} \\
\midrule
\multicolumn{9}{l}{\emph{Zero-Retrain \textbf{+GSDF} Plug-In on Frozen Pretrained Hosts \& Native Edge Flagship ($\Delta$: Paired Gain vs.\ Standalone Host)}} \\
\midrule
\rowcolor{plugbg} SigMix + \textbf{GSDF} (Plug-In) & 4.17M & 99.19 & 63.82 & 35.06 & 38.03 & 35.60 & 35.64 / 41.02 ($+0.93$) & 40.95 ($+1.40$) \\
\rowcolor{plugbg} ResNet1D + \textbf{GSDF} (Plug-In) & 268.2K & 99.33 & 73.26 & 34.71 & 35.23 & 41.81 & 33.31 / 43.68 ($+3.97$) & 38.40 ($+1.39$) \\
\rowcolor{plugbg} MTL-DANN + \textbf{GSDF} (Plug-In) & 12.08M & 99.37 & 76.83 & 45.73 & 46.15 & 25.16 & 38.54 / 44.20 ($+1.21$) & 41.57 ($+1.63$) \\
\rowcolor{oursbg} \textbf{McAFF + GSDF (Plug-In)}$^{\dagger}$ & 4.77M & \textbf{99.65} & \textbf{80.04} & \textbf{55.79} & \textbf{51.02} & \secbest{42.20} & 43.98 / \textbf{53.71} ($\mathbf{+1.97}$) & 49.28 ($+2.68$) \\
\rowcolor{oursbg} \textbf{IB-Host + GSDF (Ours, Edge)}$^{\dagger}$ & \textbf{93.3K} & 99.43 & 71.38 & 44.71 & 44.90 & \textbf{48.97} & \secbest{48.32} / 50.64 ($+0.53$) & \textbf{52.80} ($\mathbf{+1.52}$) \\
\midrule
\multicolumn{9}{l}{\textbf{Panel C: ORBIT MobRFFI Cross-Receiver Spatial-Grid Transfer} ($K=10, A=1$; 2-RX source \texttt{\{1-1, 1-20\}} $\to$ unseen \texttt{RX 20-1, 19-19})} \\
\midrule
Standalone Baselines (Paradigm) & Params & \texttt{Src.\ Val} & \texttt{20-1} Clean & \texttt{19-19} Clean & MP ($100/200$) & Noise ($10/5$) & $\overline{\mathrm{F1}}_{\mathrm{Mob}}$ (Direct / U-CAPC) & 6-Seed Holdout \\
\midrule
MCCNN~\cite{mccnn_lte_open_set} (DCTF) & 61.7K & 49.69 & 17.97 & 20.62 & 19.41 & 9.10 & 12.28 / 15.94 & 15.84 \\
Xie-SR~\cite{xie2024robustrffingerprintidentification} (Null Bin) & 195.4K & 87.34 & 39.19 & 31.27 & 28.75 & 10.84 & 17.78 / 24.94 & 24.90 \\
NS-Branch (Null, Ours) & 46.3K & 88.53 & 46.48 & 32.12 & 28.63 & 10.16 & 18.00 / 26.03 & 25.95 \\
MTL-SEI~\cite{mtl_adversarial_sei} (MTL/DA) & 12.03M & 97.41 & 31.38 & 47.16 & 27.68 & 18.52 & 18.02 / 28.49 & 29.31 \\
SigMix~\cite{wan2025sigmix} (DG Mix) & 4.12M & 99.22 & 30.94 & 50.81 & 29.36 & 19.87 & 19.29 / 30.04 & 30.53 \\
MTL-DANN~\cite{ganin2016domain} (Adv.\ DA) & 12.03M & 99.39 & 43.74 & 56.11 & 35.03 & 16.13 & 22.43 / 33.70 & 34.52 \\
MTL-CORAL~\cite{sun2016deep} (Stat.\ DA) & 12.03M & 99.40 & 42.26 & 58.65 & 35.84 & 16.84 & 22.70 / 34.38 & 34.14 \\
ResNet1D~\cite{he2016deep} (IQ CNN) & 221.3K & 99.61 & 34.71 & \secbest{63.84} & 26.07 & 28.16 & 18.84 / 34.50 & 34.37 \\
McAFF~\cite{multichannel_attentive_fusion} (4-View) & 4.72M & \textbf{99.99} & 54.83 & 51.31 & \secbest{41.90} & \secbest{31.69} & 25.72 / 42.22 & 42.16 \\
IB-Host (In-Band, Ours) & 46.3K & 99.43 & \secbest{62.41} & 61.23 & 34.91 & \textbf{40.95} & \secbest{37.21} / \secbest{45.89} & \secbest{45.96} \\
\midrule
\multicolumn{9}{l}{\emph{Zero-Retrain \textbf{+GSDF} Plug-In on Frozen Pretrained Hosts \& Native Edge Flagship ($\Delta$: Paired U-CAPC Gain vs.\ Standalone Host)}} \\
\midrule
\rowcolor{plugbg} SigMix + \textbf{GSDF} (Plug-In) & 4.17M & 99.31 & 32.48 & 50.97 & 29.96 & 19.87 & 19.71 / 30.52 ($+0.48$) & 30.83 \\
\rowcolor{plugbg} MTL-DANN + \textbf{GSDF} (Plug-In) & 12.08M & 99.45 & 46.20 & 59.38 & 36.46 & 16.13 & 23.42 / 35.13 ($+1.43$) & 35.95 \\
\rowcolor{plugbg} ResNet1D + \textbf{GSDF} (Plug-In) & 267.6K & \secbest{99.69} & 35.33 & \textbf{68.26} & 28.61 & 28.16 & 19.05 / 36.19 ($+1.69$) & 35.98 \\
\rowcolor{oursbg} \textbf{McAFF + GSDF (Plug-In)} & 4.77M & \textbf{99.99} & 56.53 & 50.51 & \textbf{42.43} & \secbest{31.69} & 25.26 / 42.54 ($+0.32$) & 42.44 \\
\rowcolor{oursbg} \textbf{IB-Host + GSDF (Ours, Edge)} & \textbf{92.6K} & 99.45 & \textbf{67.24} & 62.70 & 38.57 & \textbf{40.95} & \textbf{38.78} / \textbf{48.16} ($\mathbf{+2.27}$) & \textbf{48.17} \\
\bottomrule
\end{tabular}\par
\vspace{2pt}{\scriptsize\noindent\raggedright \emph{Note:} \texttt{Src.\ Val} reports 20\% in-domain Clean Direct F1; target regime columns report calibrated \textbf{U-CAPC} ($\rho{=}1$) Macro-F1, while $\overline{\mathrm{F1}}$ co-reports \textbf{Direct / U-CAPC} with paired gain ($+\Delta$ in pp over the standalone host) in shaded zero-retrain \textbf{+GSDF} rows (per-slice Direct scores, $E_3/E_2$ splits, and $\mathrm{SD}_{\mathrm{split}}$ are in \supptabI; \textbf{bold} / \secbest{underlined}: best / second-best). $^{\dagger}$Under \textbf{UOT-CAPC} ($\rho^*{<}1$, last column of Panel~B), \textbf{IB-Host + GSDF} reaches $\mathbf{75.75\%}$ on skewed Day~8 (\textbf{McAFF + GSDF}: $\mathbf{77.65\%}$) and $\mathbf{49.43\%}$ on \texttt{node2-1}.\par}
\end{table*}

\section{Experimental Results and Discussion}
\label{sec:results}

To make every experimental comparison clear at a glance for edge IoT evaluation, our experiments address five progressive research questions (\textbf{RQ1}--\textbf{RQ5}):
(i)~\textbf{RQ1 (Section~\ref{sec:main_results}, Table~\ref{tab:main_results}, Fig.~\ref{fig:snr_mp}):} How do standalone active-band, differential, and null-subcarrier RFFI paradigms compare against zero-retrain \textbf{+GSDF} across cross-room (\textbf{DataPort-WiFi}), cross-week/AP (\textbf{WiSig}), and spatial-grid (\textbf{ORBIT MobRFFI}) IoT transfers?
(ii)~\textbf{RQ2 (Section~\ref{sec:track_p}, Table~\ref{tab:track_p}, Fig.~\ref{fig:track_p}):} Does attaching frozen \textbf{NS-Branch} and our blind SNR gate upgrade \emph{any} deployed host backbone ($61.7\text{K}\text{--}12.03\text{M}$ params) in-place with zero host retraining and zero low-SNR negative transfer?
(iii)~\textbf{RQ3 (Section~\ref{sec:module_ablations}, Table~\ref{tab:ablation_core}, Fig.~\ref{fig:module_ablations}):} Is each of the four design pillars---subband encoders ($A{=}1$ vs.\ $A{=}4$), dual-space fusion, thermal-SNR gating, and forward-only \textbf{UOT-CAPC}/\textbf{U-CAPC} (plus matched channel-augmentation fairness)---necessary?
(iv)~\textbf{RQ4 (Section~\ref{sec:wisig_mob_deepdive}, Table~\ref{tab:wisig_mob_summary}, Fig.~\ref{fig:wisig_mob_ext}):} How does \textbf{GSDF-CAPC} scale to multi-week IoT drift (\texttt{Day 1$\to$23}), larger emitter cohorts ($K{=}10\text{--}47$) and 2-AP source pooling, AP trigger-timing synchronization (\textbf{GSDF-STO}/\textbf{GSDF-Tri}), and multi-packet IoT bursts ($B{=}1\text{--}10$)?
(v)~\textbf{RQ5 (Section~\ref{sec:gates}, Figs.~\ref{fig:gates}--\ref{fig:impairments}):} How does the framework perform across natural OTA SNR quintiles, capture window sizes $N_{\mathcal{G}}$, receiver impairments, interference, 1-thread edge gateway CPU runtimes, and symmetrically timing-tuned / capacity-matched baselines?

\subsection{RQ1: Cross-Domain Multi-Dataset IoT Benchmark}
\label{sec:main_results}

\begin{figure*}[!t]
\centering
\includegraphics[width=0.71\textwidth]{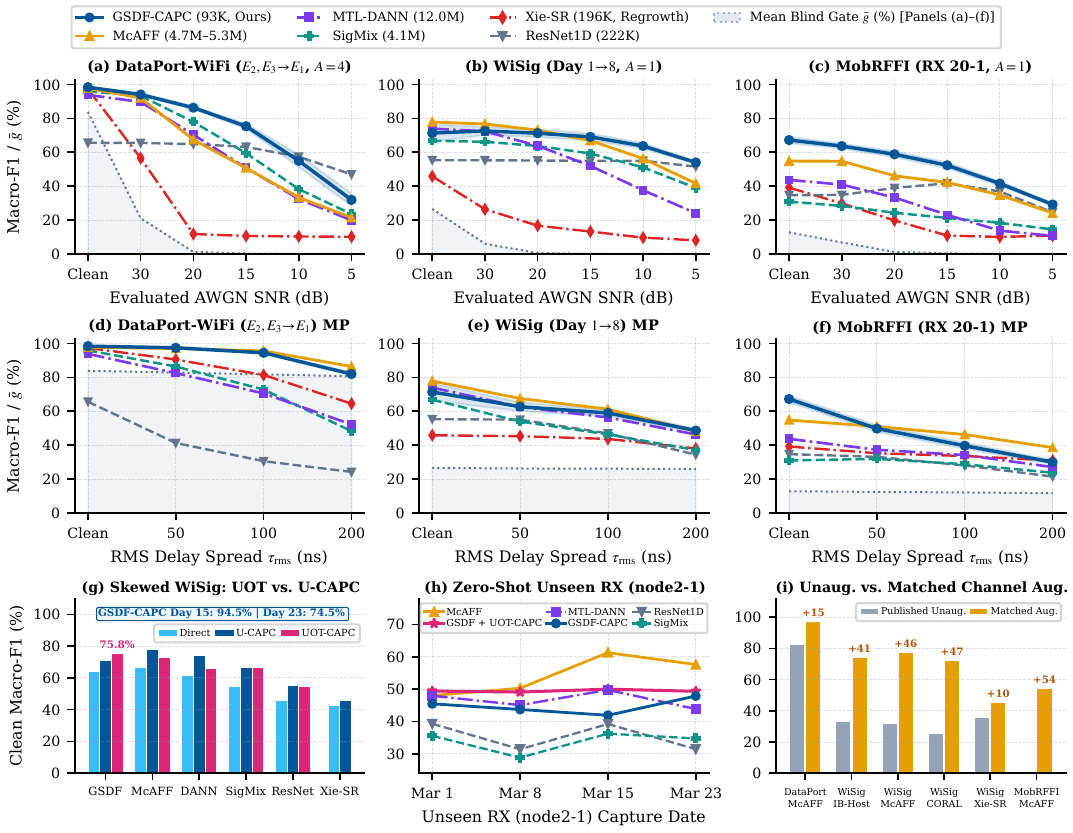}
\caption{Multi-baseline channel sweeps, multi-week / unseen-RX transfers, and one-per-paradigm augmentation audit ($3\times 3$ 9-run mean; blue shading shows $\pm 1$~SD of \textbf{GSDF-CAPC}, gray dotted fill shows mean blind gate $\bar{g}$): (a)--(c) AWGN sweeps (Clean to 5~dB) and (d)--(f) indoor multipath sweeps ($\tau_{\mathrm{rms}} \in \{0, 50, 100, 200\}$~ns) comparing \textbf{GSDF-CAPC}, \textbf{McAFF}, \textbf{MTL-DANN}, \textbf{SigMix}, \textbf{Xie-SR}, and \textbf{ResNet1D}; (g) skewed \textbf{WiSig} Day~8 ($2.88:1$) comparing \textbf{Direct}, balanced \textbf{U-CAPC}, and general \textbf{UOT-CAPC} (with \textbf{Xie-SR} evaluated under \textbf{Direct} and \textbf{U-CAPC}) plus Day~15/23 drift; (h) zero-shot unseen-RX (\texttt{node2-1}) across the 8-host plug-in suite over four dates; (i) clean \textbf{U-CAPC} jump from unaugmented published training to matched channel augmentations across representative backbones (including \textbf{IB-Host}).}
\label{fig:snr_mp}
\end{figure*}

Table~\ref{tab:main_results} and Fig.~\ref{fig:snr_mp} jointly evaluate held-out source separability (\texttt{Src.\ Val}), standalone RFFI baselines across all four paradigm families (upper block of each panel), and zero-retrain pluggable \textbf{+GSDF} (lower shaded block, reporting paired $+\Delta$ gains directly alongside $\overline{\mathrm{F1}}$) across all three Wi-Fi IoT testbeds:

\subsubsection{In-Domain Source Separability vs.\ Physical Cross-Domain Regimes}
Reporting held-out 20\% \texttt{Src.\ Val} alongside cross-domain target slices separates representation capacity from channel/AP shift. On source validation preambles, every populated-band and \textbf{+GSDF} model achieves near-ceiling separability ($94.35\%\text{--}99.79\%$ on \textbf{DataPort-WiFi}, $99.11\%\text{--}99.65\%$ on \textbf{WiSig}, and $97.41\%\text{--}99.99\%$ on two-RX \textbf{MobRFFI}), confirming that cross-domain drops stem purely from propagation and front-end transfer rather than source underfitting. Across target regimes, Fig.~\ref{fig:snr_mp}(a)--(f) mirrors Proposition~\ref{prop:snr_collapse}: (i)~unpopulated-bin encoders (\textbf{Xie-SR} and $4.2\times$ smaller \textbf{NS-Branch}) achieve $96.40\%\text{--}97.39\%$ clean U-CAPC on $E_2,E_3\!\to\!E_1$ ($67.29\%\text{--}73.58\%$ multipath) yet collapse quadratically below $\gamma_{\times} \approx 25\text{--}35$~dB to $1.82\%\text{--}4.51\%$ Direct ($8.76\%\text{--}10.84\%$ U-CAPC) at $10/5$~dB across all three panels, right where our blind gate $\bar{g}$ shuts off; (ii)~active-band $12.03\text{M}$ \textbf{MTL-DANN/CORAL/SEI} and $4.12\text{M}$ \textbf{SigMix} degrade to $46.50\%\text{--}53.38\%$ under \textbf{DataPort-WiFi} $100/200$~ns multipath due to nominal-symbol entanglement~\eqref{eq:subband_obs}; and (iii)~$4.72\text{M}\text{--}5.30\text{M}$ \textbf{McAFF} excels under high-SNR multipath ($85.99\%$ on \textbf{DataPort-WiFi}) yet squares AWGN in its lag-16 differential views ($9.68\%$ and $2.00\%$ Direct at $10/5$~dB on \textbf{DataPort-WiFi} and \textbf{MobRFFI}).

\subsubsection{Foregrounding Plug-In Generality vs.\ Native Edge Gateway Pareto Optimality}
Comparing the upper and lower blocks of Table~\ref{tab:main_results} clarifies the central contribution of \textbf{GSDF} for heterogeneous IoT deployments: no single fixed-band representation dominates every SNR, multipath, and gateway parameter budget, whereas \emph{attaching frozen \textbf{NS-Branch} and our blind SNR gate with zero host retraining improves overall $\overline{\mathrm{F1}}$ on every backbone ($72/72$ runs) and claims first place across every calibrated target-regime and overall $\overline{\mathrm{F1}}$ column of all three benchmarks} (while standalone \textbf{IB-Host} edges out \textbf{IB-Host + GSDF} by $0.21$~pp on uncalibrated single-frame \textbf{WiSig} Direct, $48.53\%$ vs.\ $48.32\%$, and isolated single-antenna clean slices such as \textbf{SigMix} Day~8 dip slightly prior to multipath/all-regime averaging). Specifically, when deployed with our lightweight native edge host, \textbf{IB-Host + GSDF (Ours, Edge)} ($92.6\text{K}\text{--}93.3\text{K}$ active params, $57\times$ smaller than \textbf{McAFF}) leads overall $\overline{\mathrm{F1}}_{\mathrm{DP}}$ ($67.63\% / 73.71\%$, $+8.73$~pp over \textbf{IB-Host} and $+6.56 / +4.94$~pp over \textbf{McAFF}) and $\overline{\mathrm{F1}}_{\mathrm{Mob}}$ ($38.78\% / 48.16\%$, $+2.27$~pp over \textbf{IB-Host} and $+13.06 / +5.94$~pp over \textbf{McAFF}), while leading \textbf{WiSig} under general \textbf{UOT-CAPC} ($\mathbf{52.80\%}$ vs.\ $46.60\%$ standalone \textbf{McAFF} and $49.28\%$ \textbf{McAFF + GSDF}). Simultaneously, in high-SNR regimes where server-grade backbones hold an architectural edge over a $46.3\text{K}$ host, plugging \textbf{+GSDF} into those exact backbones resolves their regime blind spots without losing $0.00$~pp at $10/5$~dB: \textbf{McAFF + GSDF} reaches $\mathbf{99.26\%}$ clean and $\mathbf{88.91\%}$ multipath on \textbf{DataPort-WiFi} ($+1.87$~pp $\overline{\mathrm{F1}}_{\mathrm{DP}}$) plus $\mathbf{80.04\%}$ ($\mathbf{77.65\%}$ UOT) Day~8, $\mathbf{55.79\%}$ unseen \texttt{node2-1}, and $\mathbf{53.71\%}$ $\overline{\mathrm{F1}}_{\mathrm{WiSig}}$ ($+1.97$~pp U-CAPC, $+2.68$~pp UOT); \textbf{ResNet1D + GSDF} retains its $\mathbf{44.89\%}$ low-SNR ceiling while jumping $\mathbf{+25.75}$~pp on $\overline{\mathrm{F1}}_{\mathrm{DP}}$ ($\mathbf{+42.98}$~pp multipath: $24.44\% \to 67.42\%$), $\mathbf{+17.92}$~pp on \textbf{WiSig} Day~8 ($55.34\% \to 73.26\%$), $\mathbf{97.94\%}$ on two-week $\mathrm{Day\ 1}\!\to\!\mathrm{15}$ (Section~\ref{sec:wisig_mob_deepdive}), and $\mathbf{68.26\%}$ ($\#1$) on \textbf{MobRFFI} \texttt{RX 19-19}; and \textbf{MTL-DANN + GSDF} / \textbf{SigMix + GSDF} gain $\mathbf{+9.50}$ and $\mathbf{+11.36}$~pp on $\overline{\mathrm{F1}}_{\mathrm{DP}}$ ($\mathbf{+20.96}$ and $\mathbf{+24.89}$~pp multipath). Across all panels, the \textbf{6-Seed Holdout} ($s \ne r$) matches 9-run $\overline{\mathrm{F1}}$ within $0.01\text{--}0.54$~pp.

\subsubsection{Cross-Dataset Physical Diagnostics, Source AP Fusion, and \texorpdfstring{UOT-CAPC $\to$ U-CAPC}{UOT-CAPC to U-CAPC}}
Table~\ref{tab:main_results} also explains why standalone \textbf{NS-Branch} / \textbf{Xie-SR} drop from $97.96\%$ \texttt{Src.\ Val} ($58.72\%$ cross-room U-CAPC) on $A=4$ \textbf{DataPort-WiFi} to $24.80\% / 27.17\%$ on $A=1$ \textbf{WiSig} and $26.03\% / 24.94\%$ on $A=1$ \textbf{MobRFFI}. On \textbf{DataPort-WiFi}, four synchronized X310 chains yield a $-41.2$~dBc null floor ($44.94\%$ emitter-repeatable null variance) and a $0.84\times$ cross-room drift-to-inter-TX-spacing ratio; on single-antenna \textbf{WiSig} and \textbf{MobRFFI}, higher OTA noise floors ($-24.0$ and $-26.3$~dBc) reduce emitter-repeatable null variance to $15.44\%$ and $26.92\%$ while elevating cross-week/cross-RX drift ratios to $3.37\times$ and $2.40\times$ (further reflected in \textbf{MobRFFI}'s $88.53\%$ null-only \texttt{Src.\ Val} across two pooled source APs versus $99.43\%$ for active-band \textbf{IB-Host}), compounded by $1\text{--}2$~sample cross-session packet-trigger timing offsets analyzed in Section~\ref{sec:wisig_mob_deepdive}. Moreover, comparing source pooling regimes shows that two-AP enrollment on \textbf{MobRFFI} enables $67.24\%$ transfer to unseen \texttt{RX 20-1}, whereas strict 1-AP 1-day enrollment on \textbf{WiSig} leaves unseen \texttt{node2-1} front-end offsets uncalibrated ($44.71\%\text{--}55.79\%$). Finally, validating Section~\ref{sec:ucapc}'s hierarchy on skewed \textbf{WiSig} Day~8 ($2.88:1$ emitter ratio, Fig.~\ref{fig:snr_mp}(g)), general KL-relaxed \textbf{UOT-CAPC} ($\rho^* < 1$) lifts \textbf{IB-Host + GSDF} from $71.38\%$ (balanced \textbf{U-CAPC}, $\rho=1$) to $\mathbf{75.75\%}$ ($\mathbf{52.80\%}$ $\overline{\mathrm{F1}}_{\mathrm{WiSig}}$) and unseen \texttt{node2-1} from $44.71\%$ to $\mathbf{49.43\%}$ (while balanced two-week $\mathrm{Day\ 1}\!\to\!\mathrm{15}$ reaches $\mathbf{94.54\%\text{--}97.94\%}$, Table~\ref{tab:wisig_mob_summary}).

\subsection{RQ2: Zero-Retrain Plug-and-Play Generality Across Eight Host Backbones}
\label{sec:track_p}

\begin{table*}[!t]
\centering
\caption{Zero-Retrain Plug-In Generality Across Eight RFFI Host Backbones ($3\times 3$ 9-Run Three-Benchmark Grid)}
\label{tab:track_p}
\scriptsize
\renewcommand{\arraystretch}{0.95}
\setlength{\tabcolsep}{3.6pt}
\begin{tabular}{llcccccccc}
\toprule
Host Backbone & Params & \begin{tabular}[c]{@{}c@{}}Host Alone\\(Direct / U-CAPC)\end{tabular} & \begin{tabular}[c]{@{}c@{}}+ Hard\\Switch\end{tabular} & \begin{tabular}[c]{@{}c@{}}\textbf{+ Soft GSDF}\\(Direct / U-CAPC)\end{tabular} & $\Delta$ Direct & $\Delta$ U-CAPC [95\% CI] & $\Delta$ UOT & \begin{tabular}[c]{@{}c@{}}9-Run\\W/T/L\end{tabular} & $\Delta_{10/5\mathrm{dB}}$ \\
\midrule
\rowcolor{oursbg} \textbf{IB-Host (Ours, Edge)} & \textbf{46.3K--46.6K} & 47.78 / 53.66 & 52.09 & \textbf{51.58} / \textbf{57.50} & +3.80 & \textbf{+3.84} [+3.60, +4.09] & +5.10 & 9/0/0 & $-0.00$ \\
\rowcolor{oursbg} \textbf{McAFF}~\cite{multichannel_attentive_fusion} & 4.72M--5.30M & 43.37 / 54.24 & 49.76 & \textbf{44.04} / \textbf{55.63} & +0.68 & \textbf{+1.39} [+1.19, +1.58] & +1.52 & 9/0/0 & $+0.00$ \\
MTL-DANN~\cite{ganin2016domain} & 12.03M & 37.68 / 43.65 & 42.68 & \textbf{40.86} / \textbf{47.70} & +3.18 & \textbf{+4.05} [+3.87, +4.22] & +4.22 & 9/0/0 & $+0.00$ \\
MTL-CORAL~\cite{sun2016deep} & 12.03M & 37.31 / 43.55 & 42.62 & \textbf{40.51} / \textbf{47.69} & +3.19 & \textbf{+4.14} [+3.92, +4.39] & +4.42 & 9/0/0 & $+0.00$ \\
SigMix~\cite{wan2025sigmix} & 4.12M & 34.56 / 41.08 & 43.59 & \textbf{38.22} / \textbf{45.34} & +3.66 & \textbf{+4.26} [+3.84, +4.74] & +4.39 & 9/0/0 & $-0.00$ \\
MTL-SEI~\cite{mtl_adversarial_sei} & 12.03M & 34.35 / 39.77 & 40.29 & \textbf{37.59} / \textbf{44.24} & +3.24 & \textbf{+4.47} [+4.14, +4.84] & +4.62 & 9/0/0 & $+0.00$ \\
\rowcolor{plugbg} ResNet1D~\cite{he2016deep} & 221.3K--223.9K & 28.79 / 38.39 & 46.74 & \textbf{36.41} / \textbf{48.86} & +7.63 & \textbf{+10.47} [+10.09, +10.79] & +9.59 & 9/0/0 & $-0.00$ \\
MCCNN~\cite{mccnn_lte_open_set} & 61.7K--213.4K & 14.00 / 17.90 & \textbf{37.57} & 26.14 / 33.92 & +12.14 & \textbf{+16.01} [+15.78, +16.21] & +14.47 & 9/0/0 & $+0.04$ \\
\bottomrule
\end{tabular}\par
\vspace{2pt}{\scriptsize\noindent\raggedright \emph{Note:} Three-benchmark mean $\overline{\mathrm{F1}}$ (\%, $\Delta$ in pp); frozen \textbf{NS-Branch} and thermal-SNR gate are attached with zero host retraining ($72/72$ paired run wins, Holm-Bonferroni exact permutation $p_{\mathrm{Holm}} < 10^{-4}$ on all eight hosts; \textbf{Hard Switch}: U-CAPC under $\mathbb{I}[\hat{s}_n \ge \beta^*]$).\par}
\end{table*}

\begin{figure*}[!t]
\centering
\includegraphics[width=0.73\textwidth]{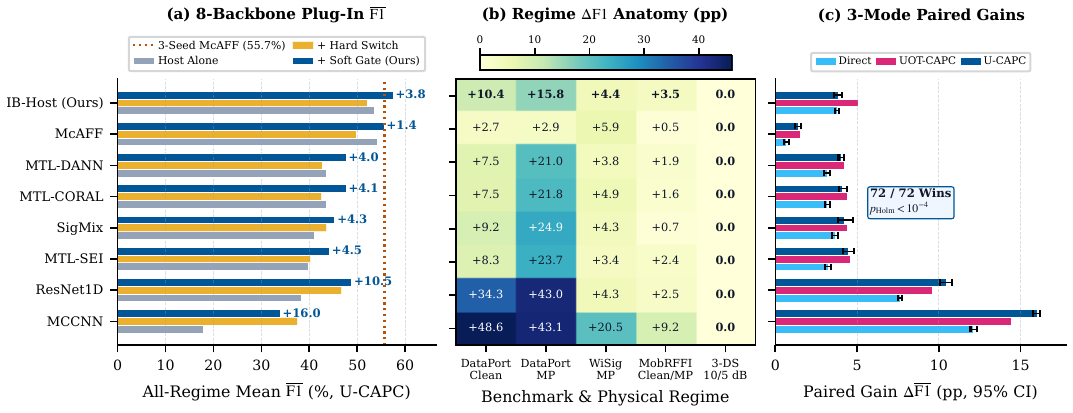}
\caption{Zero-retrain plug-in generality across eight heterogeneous RFFI backbones ($n=9$ runs): (a) three-benchmark $\overline{\mathrm{F1}}$ comparing standalone hosts, binary \textbf{Hard Switch}, and dual-space \textbf{Soft GSDF} against 3-seed \textbf{McAFF}; (b) regime-resolved $\Delta\mathrm{F1}$ heatmap (pp, U-CAPC) showing multipath/clean rescues with $0.0$~pp low-SNR loss; (c) paired $\Delta\overline{\mathrm{F1}}$ gains (95\% bootstrap CIs) across \textbf{Direct}, general \textbf{UOT-CAPC}, and balanced \textbf{U-CAPC}.}
\label{fig:track_p}
\end{figure*}

Directly corroborating Proposition~\ref{prop:plugin_fallback} (Claim~1) and extending Table~\ref{tab:main_results} to all eight backbones in our benchmark suite, Table~\ref{tab:track_p} and Fig.~\ref{fig:track_p} demonstrate seamless in-field upgradability for deployed IoT AP classifiers by attaching frozen \textbf{NS-Branch} ($46.3\text{K}$) and our blind thermal-SNR gate to eight diverse pretrained hosts spanning $d_{\mathrm{host}} \in \{64, 84, 256, 512\}$ with zero weight fine-tuning or projection layers. Adding $<1.2\%$ active parameters to \textbf{SigMix}/\textbf{McAFF}/\textbf{MTL} models yields \textbf{unanimous $72/72$ paired run wins} ($p_{\mathrm{Holm}} < 10^{-4}$) across \textbf{Direct}, \textbf{Balanced U-CAPC}, and \textbf{UOT-CAPC}, confirming that per-frame $z$-score standardization~\eqref{eq:logit_znorm} and orthogonal spherical concatenation~\eqref{eq:gsdf_emb} bridge $194\times$ parameter disparities ($61.7\text{K}\text{--}12.03\text{M}$) without cross-model tuning. As shown in Fig.~\ref{fig:track_p}(b), on \textbf{DataPort-WiFi} $100/200$~ns multipath, \textbf{GSDF} lifts \textbf{MTL-DANN} from $53.38\%$ to $74.34\%$ ($\mathbf{+20.96}$~pp), \textbf{MTL-CORAL} by $\mathbf{+21.85}$~pp, \textbf{SigMix} from $46.50\%$ to $71.39\%$ ($\mathbf{+24.89}$~pp), and \textbf{ResNet1D} from $24.44\%$ to $67.42\%$ ($\mathbf{+42.98}$~pp), while raising \textbf{McAFF} to $\mathbf{96.48\%}$ three-direction clean ($\mathbf{99.26\%}$ on $E_2,E_3\!\to\!E_1$), $\mathbf{88.91\%}$ multipath ($+2.92$~pp), and $\mathbf{80.04\%}$ on \textbf{WiSig} Day~8 ($+5.87$~pp on \textbf{WiSig} multipath). Consistent with Proposition~\ref{prop:plugin_fallback} (Claim~3), $10/5$~dB $\Delta\mathrm{F1}$ equals $0.00$~pp on 7 of 8 hosts (and $+0.04$~pp on collapsed \textbf{MCCNN}, where $g_n \sim 10^{-4}$ breaks near-tied low-SNR assignments). Moreover, dual-space soft fusion beats binary \textbf{Hard Switch} ($g_{\mathrm{hard}}(n) = \mathbb{I}[\hat{s}_n \ge \beta^*]$) on 7 of 8 backbones by $+1.74$ to $+5.87$~pp, confirming continuous subband complementarity in the transition band $\hat{s}_n \approx \beta^*$.

\subsection{RQ3: Four-Pillar Module Ablation and Channel-Augmentation Fairness Audit}
\label{sec:module_ablations}

\begin{figure*}[!t]
\centering
\includegraphics[width=0.80\textwidth]{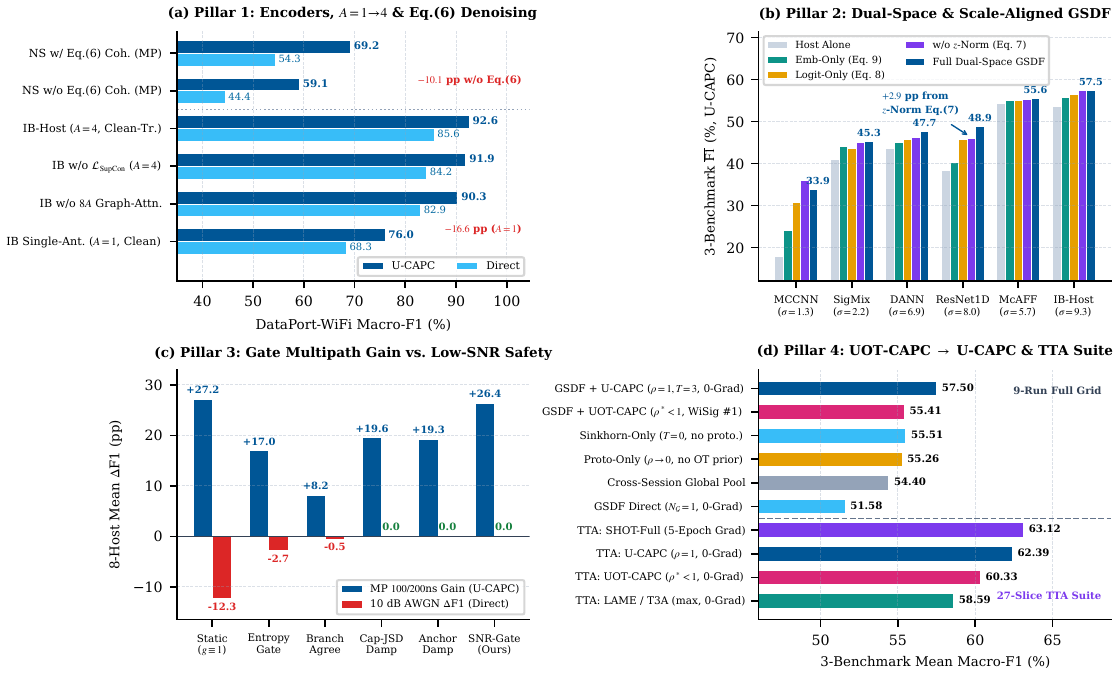}
\caption{Four-pillar module ablation summary of pluggable \textbf{GSDF-CAPC}: (a) \textbf{Pillar~1} internal encoder, single-antenna ($A=1$) vs.\ four-antenna ($A=4$) spatial diversity, and symbol-pair coherent null denoising~\eqref{eq:null_coh} on \textbf{DataPort-WiFi}; (b) \textbf{Pillar~2} dual-space fusion~\eqref{eq:logit_znorm}--\eqref{eq:gsdf_emb} across backbones with disparate clean logit scales ($\sigma \in [1.3, 9.3]$); (c) \textbf{Pillar~3} label-free gate formulations across eight hosts comparing $100/200$~ns multipath U-CAPC gain against $10$~dB AWGN Direct negative transfer; (d) \textbf{Pillar~4} general \textbf{UOT-CAPC} $\to$ balanced \textbf{U-CAPC} step decomposition (Algorithm~\ref{alg:ucapc}) on the 9-run grid and comparison against faithful source-free TTA baselines on the 27-slice suite.}
\label{fig:module_ablations}
\end{figure*}

\begin{table*}[!t]
\centering
\caption{Core Four-Pillar Module Ablations of Pluggable \textbf{GSDF-CAPC} ($3\times 3$ 9-Run Grid; Separate Headers Prevent Cross-Scope Denominator Mixing)}
\label{tab:ablation_core}
\scriptsize
\renewcommand{\arraystretch}{0.87}
\setlength{\tabcolsep}{2.4pt}
\begin{tabular}{llccccl}
\toprule
\multicolumn{7}{l}{\textbf{Pillar 1a: Clean-Trained IB-Host Encoder \& Spatial-Diversity ($A{=}1$ vs.\ $A{=}4$) Ablations on \textbf{DataPort-WiFi} ($E_2,E_3\!\to\!E_1$)}$^{\mathrm{a}}$} \\
\midrule
Encoder / Spatial Control & Params & Clean & Multipath & Noise ($10/5$~dB) & Clean (Direct / U-CAPC) & Statistical / Physical Role \\
\midrule
IB-Host Full ($A{=}4$, Clean-Tr.) & 46.3K & \textbf{92.57} & 42.16 & \textbf{55.68} & \textbf{85.64} / \textbf{92.57} & $8A{=}32$ graph (aug.: $86.35/93.07$) \\
IB-Host Single-Antenna ($A{=}1$) & 46.3K & 75.97 & --- & 48.54 & 68.32 / 75.97 & $-16.60$~pp ($p{=}8.4{\times}10^{-6}$, $0/9$) \\
IB-Host w/o $8A$ Graph-Attn ($A{=}4$) & 46.5K & \secbest{90.35} & --- & --- & \secbest{82.94} / \secbest{90.35} & $-2.22$~pp ($p{=}0.028$, relational) \\
\midrule
\multicolumn{7}{l}{\textbf{Pillar 1b: Paired Symbol-Pair Coherent Null Denoising~\eqref{eq:null_coh} Audit on \textbf{DataPort-WiFi} (3-Room LOEO, 27-Run Grid)}} \\
\midrule
Null-Branch Configuration & Params & 3-Rm Clean & 3-Rm MP & 3-Rm Noise & Clean (Direct / U-CAPC) & Multipath \& Low-SNR Fallback \\
\midrule
\rowcolor{oursbg} \textbf{GSDF w/ Eq.~\eqref{eq:null_coh} Coh.\ Null (Ref.)} & \textbf{92.6K} & \textbf{94.40} & \textbf{69.16} & \textbf{52.89} & \textbf{92.43} / \textbf{94.40} & 3-LOEO ref.\ ($E_1$: $98.69/75.96$) \\
GSDF w/o Eq.~\eqref{eq:null_coh} Coh.\ Null & 92.6K & 92.15 & 59.11 & \textbf{52.89} & 88.86 / 92.15 & $\mathbf{-10.05}$~pp MP; $\mathbf{0.00}$~pp Noise \\
\midrule
\multicolumn{7}{l}{\textbf{Pillar 1c: Three-Benchmark Subband Orthogonality ($\theta_{\mathrm{host}}\perp\theta_{\mathrm{null}}$) vs.\ Multi-Seed Ensembles (9-Run 3-Dataset Grid)}} \\
\midrule
Subband / Ensemble Arm & Params & 3-DS Clean & 3-DS MP & 3-DS Noise & 3-DS (Direct / U-CAPC) & Multi-Benchmark Audit \\
\midrule
NS-Branch Alone (Null Subband) & 46.3K & 55.53 & 44.20 & 9.82 & 30.70 / 36.52 & $+0.30$~pp vs.\ $4.2{\times}$ Xie-SR ($36.2\%$) \\
IB-Host Alone (1-Seed In-Band) & 46.3K & 68.77 & 48.32 & \secbest{43.88} & 47.78 / 53.66 & $50.32\%$ UOT-CAPC baseline \\
2-Seed Homogeneous IB-Host & 92.6K & \secbest{69.79} & \secbest{49.48} & \textbf{45.06} & \secbest{48.62} / \secbest{54.78} & $-6.17$~pp MP vs.\ full GSDF \\
\rowcolor{oursbg} \textbf{IB-Host + GSDF (Full, Ours)} & \textbf{92.6K} & \textbf{72.98} & \textbf{55.65} & \secbest{43.88} & \textbf{51.58} / \textbf{57.50} & \textbf{55.41\% UOT; $>$ 3-Seed ($55.18\%$)} \\
\midrule
\multicolumn{7}{l}{\textbf{Pillar 2: Dual-Space Fusion Formulation~\eqref{eq:logit_znorm}--\eqref{eq:gsdf_emb} on \textbf{IB-Host + GSDF} Across Three Benchmarks}} \\
\midrule
Fusion Variant & Params & \textbf{DataPort} & \textbf{WiSig} & \textbf{MobRFFI} & 3-DS (Direct / U-CAPC) & Cross-Host / Safety Audit \\
\midrule
Binary Hard Switch ($g_{\mathrm{hard}}$) & 92.6K & 68.64 & 46.77 & 40.87 & 48.10 / 52.09 & $-5.41$~pp U-CAPC (transition) \\
Embedding-Only (w/o Eq.~\eqref{eq:gsdf_logit}) & 92.6K & 69.37 & \secbest{51.13} & 46.90 & 47.78 / 55.80 & $0.00$~pp Direct; $-8.57$ ResNet \\
Logit-Only (w/o Eq.~\eqref{eq:gsdf_emb}) & 92.6K & 72.31 & 50.13 & 46.90 & \textbf{51.58} / 56.45 & $-1.05$~pp U-CAPC; $-3.01$ ResNet \\
w/o Logit $z$-Norm (w/o Eq.~\eqref{eq:logit_znorm}) & 92.6K & \secbest{73.71} & \textbf{51.22} & 47.15 & \secbest{51.50} / 57.36 & $-2.88$~pp ResNet, $-1.47$ DANN \\
Static Emb.\ Concat (unwarped) & 92.6K & \textbf{73.80} & 51.02 & \textbf{50.29} & \textbf{51.58} / \textbf{58.37} & Violates Prop.~2 ($-0.41$ noise) \\
\rowcolor{oursbg} \textbf{Full Dual-Space GSDF (Ours)} & \textbf{92.6K} & \secbest{73.71} & 50.64 & \secbest{48.16} & \textbf{51.58} / \secbest{57.50} & \textbf{72/72 wins, $0.00$~pp loss} \\
\midrule
\multicolumn{7}{l}{\textbf{Pillar 3: Blind Gate Rules Averaged Over All Eight Host Backbones (Table~\ref{tab:track_p} Suite)}} \\
\midrule
Gate Rule & Control Signal & \begin{tabular}[c]{@{}c@{}}\textbf{DataPort}\\Clean\end{tabular} & \begin{tabular}[c]{@{}c@{}}\textbf{DataPort}\\Multipath\end{tabular} & \begin{tabular}[c]{@{}c@{}}$10$~dB Direct\\($\Delta$~pp)\end{tabular} & \begin{tabular}[c]{@{}c@{}}\textbf{MobRFFI}\\\texttt{19-19}\end{tabular} & 8-Host Safety Audit \\
\midrule
Host-Only & $g_n \equiv 0$ & 76.53 & 46.43 & \textbf{22.47} ($0.00$) & 50.92 & Reference ($0.00$~pp noise loss) \\
Static-Sum & $g_n \equiv 1$ & \textbf{94.28} & \textbf{73.63} & 10.14 ($-12.33$) & 43.77 & Fails on $8/8$ hosts ($-30.2$ IB) \\
Entropy-Gate & $\tilde{H}_{\mathrm{h}}/(\tilde{H}_{\mathrm{h}}{+}\tilde{H}_{\mathrm{n}})$ & 89.20 & 63.47 & 19.74 ($-2.73$) & \textbf{53.68} & Overconfident OOD ($-11.1$ IB) \\
Branch-Agree & $1{-}\mathrm{JSD}/\log 2$ & 82.06 & 54.59 & \secbest{21.94} ($-0.53$) & 49.23 & Forfeits $70\%$ of MP rescue \\
\rowcolor{oursbg} \textbf{Thermal-SNR (Ours)} & $\sigma(\alpha^*(\hat{s}_n{-}\beta^*))$ & \secbest{94.16} & \secbest{72.86} & \textbf{22.47 ($\mathbf{0.00}$)} & \secbest{51.85} & \textbf{100\% retention, $0.00$~pp loss} \\
\midrule
\multicolumn{7}{l}{\textbf{Pillar 4: Forward-Only \textbf{UOT-CAPC} / \textbf{U-CAPC} vs.\ Source-Free TTA on \textbf{IB-Host + GSDF} (27-Slice TTA Protocol)}} \\
\midrule
Calibration / TTA Method & Target Grad.\ Cost & \textbf{DataPort} & \textbf{WiSig} & \textbf{MobRFFI} & 27-Slice Mean & Calibration / Latency Audit \\
\midrule
Single-Frame \textbf{Direct} & 0-Grad ($N_{\mathcal{G}}{=}1$) & 77.00 & 51.12 & 41.42 & 56.51 & Unadapted baseline \\
T3A~\cite{iwasawa2021t3a} & 0-Grad ($M^*{=}100$) & 79.33 & 49.89 & 42.98 & 57.40 & Degrades on \textbf{WiSig} \\
LAME~\cite{boudiaf2022lame} & 0-Grad ($k\text{NN}^*$) & 78.17 & 53.53 & 44.08 & 58.59 & Laplacian affinity, no OT \\
SHOT-IM~\cite{liang2020shot} & 5-Epoch ($\text{lr}^*{=}10^{-4}$) & \textbf{87.29} & 53.13 & \secbest{47.29} & \secbest{62.57} & Backprop (uncalib.: $\le 28\%$) \\
SHOT-Full~\cite{liang2020shot} & 5-Epoch ($\text{lr}^*{=}10^{-4}$) & \secbest{85.33} & \textbf{58.69} & 45.33 & \textbf{63.12} & $-6.67$~pp vs.\ U-CAPC (\textbf{MobRFFI}) \\
\rowcolor{oursbg} \textbf{General UOT-CAPC (Ours)} & \textbf{0-Grad} ($\gamma_{\mathrm{KL}}^*{=}0.10$) & 79.74 & \secbest{55.49} & 45.75 & 60.33 & \textbf{\#1 0-Grad WiSig} ($96.5\%$, $N_{\mathcal{G}}{=}10$) \\
\rowcolor{oursbg} \textbf{Balanced U-CAPC (Ours)} & \textbf{0-Grad} ($\rho{=}1, T{=}3$) & 82.81 & 52.37 & \textbf{52.00} & 62.39 & \textbf{\#1 MobRFFI} ($<0.08$~ms/fr) \\
\bottomrule
\end{tabular}\par
\vspace{2pt}{\scriptsize\noindent\raggedright \emph{Note:} Single-value regime/dataset columns report calibrated \textbf{U-CAPC} Macro-F1 (\%). $^{\mathrm{a}}$Internal clean-trained component-ablation protocol on $E_2,E_3\!\to\!E_1$ isolating encoder/spatial structure without multi-condition channel augmentation (vs.\ $86.35\% / 93.07\%$ under matched channel augmentation in Table~\ref{tab:main_results}; `---' denotes clean-only structural runs). Pillar~1b evaluates paired 27-run 3-room LOEO \textbf{GSDF} with/without Eq.~\eqref{eq:null_coh}, Pillar~1c/2 evaluate the 3-dataset 5-condition 9-run grid, Pillar~3 averages over all 8 backbones, and Pillar~4 evaluates the 27-slice source-free TTA suite (full 50-row matrices---late pooling/Coord/SupCon encoder ablations, $E_1$ single-direction Eq.~\eqref{eq:null_coh} pair, 3-seed \textbf{IB-Host}/\textbf{McAFF}, \textbf{Xie-SR} gating, capture-damping negative results, Algorithm~\ref{alg:ucapc} step decomposition, and \textbf{McAFF} TTA---are in \supptabsIItoIII).\par}
\end{table*}

To verify that every component in Sections~\ref{sec:encoders}--\ref{sec:ucapc} is necessary and non-redundant, Fig.~\ref{fig:module_ablations} and Table~\ref{tab:ablation_core} (with full disaggregated matrices in \supptabsIItoIII) dissect \textbf{GSDF-CAPC} across its four design pillars, followed by our symmetric channel-augmentation fairness audit:

\subsubsection{Pillar~1a--1c: Subband Encoders, \texorpdfstring{$A=1$ vs.\ $A=4$}{A=1 vs. A=4} Spatial Diversity, \texorpdfstring{Eq.~\eqref{eq:null_coh}}{Eq. (6)} Coherent Denoising, and Subband Orthogonality}
Table~\ref{tab:ablation_core} (Pillar~1a/1b) and \supptabII{} (Panel~A1) first quantify each component inside our native $46.3\text{K}$ branches on \textbf{DataPort-WiFi} (Fig.~\ref{fig:module_ablations}(a)). Under the clean-trained internal ablation suite on $E_2,E_3\!\to\!E_1$ ($85.64\% / 92.57\%$ vs.\ $86.35\% / 93.07\%$ with matched channel augmentation in Table~\ref{tab:main_results}), restricting four-antenna recordings to a \textbf{single antenna ($A=1$)} drops \textbf{IB-Host} clean Macro-F1 by $\mathbf{-17.32 / -16.60}$~pp ($68.32\% / 75.97\%$, $0/9$ wins, $p = 8.4\times 10^{-6}$; $-14.43 / -15.38$~pp mean across seven encoder stems, $0/63$ wins), directly confirming that multi-antenna AP spatial diversity accounts for a major share of \textbf{DataPort-WiFi}'s absolute margin over single-antenna \textbf{WiSig} and \textbf{MobRFFI}. At $A=4$, removing the $8A$-token cross-antenna graph-attention block ($-2.70 / -2.22$~pp, $p = 0.028$) or replacing it with late per-antenna mean pooling ($-1.12 / -2.49$~pp, $p = 0.038$) degrades cross-room transfer, whereas uniform full-graph attention and coordinate jitter yield indistinguishable accuracy ($p > 0.50$), showing that the gain stems from early relational cross-antenna token mixing rather than physical array geometry priors. Removing depthwise $\mathrm{Conv1d}(k=3)$ ($-1.62 / -1.44$~pp) or $\mathcal{L}_{\mathrm{SupCon}}$ ($-1.48 / -0.66$~pp) consistently weakens spherical cluster margins. On the null branch (Pillar~1b), paired 3-room LOEO comparison shows that ablating symbol-pair coherent noise subtraction~\eqref{eq:null_coh} in favor of raw $|Y_a[m]|$ costs $-3.57 / -2.25$~pp clean and $\mathbf{-9.87 / -10.05}$~pp under $100/200$~ns multipath ($44.44\% / 59.11\%$ vs.\ $54.31\% / 69.16\%$; $-12.16 / -12.01$~pp on $E_2,E_3\!\to\!E_1$) while holding $10/5$~dB noise identically at $43.26\% / 52.89\%$ ($\mathbf{0.00}$~pp diff, confirming Proposition~\ref{prop:plugin_fallback}'s exact low-SNR fallback). Across all three benchmarks (Pillar~1c), parameter-matched \textbf{2-Seed Homogeneous IB-Host} ($92.6\text{K}$, $48.62\% / 54.78\%$) trails heterogeneous \textbf{IB-Host + GSDF} ($51.58\% / 57.50\%$) by $-2.96 / -2.72$~pp overall and $-6.17$~pp on $\mathrm{F1}_{\mathrm{mp}}$ ($49.48\%$ vs.\ $55.65\%$), while compact \textbf{NS-Branch} ($46.3\text{K}$) outperforms $4.2\times$ larger \textbf{Xie-SR} ($195.4\text{K}$) both standalone ($36.52\%$ vs.\ $36.22\%$) and when gated with \textbf{IB-Host} ($57.50\%$ vs.\ $56.88\%$, \supptabII).

\subsubsection{Pillar~2: Dual-Space Scale-Aligned Fusion\texorpdfstring{~\eqref{eq:logit_znorm}--\eqref{eq:gsdf_emb}}{ (7)-(9)} and Low-SNR Safety Boundary}
Table~\ref{tab:ablation_core} (Pillar~2), \supptabII{} (Panel~B), and Fig.~\ref{fig:module_ablations}(b) ablate each algebraic element of Section~\ref{sec:gsdf_fusion}. Replacing continuous dual-space fusion with a \textbf{Binary Hard Switch} loses $-5.41$~pp U-CAPC ($52.09\%$ vs.\ $57.50\%$), and even applying a hard step gate $\mathbb{I}[\hat{s}_n \ge \beta^*]$ inside dual-space fusion loses $-1.14$~pp U-CAPC and $-1.73$~pp UOT-CAPC ($53.68\%$ vs.\ $55.41\%$), confirming subband synergy across the transition SNR band. Fusing only embeddings (\textbf{w/o Eq.~\eqref{eq:gsdf_logit}}) forfeits all single-frame Direct gains ($47.78\%$) and loses $-8.57$~pp on \textbf{ResNet1D} ($40.29\%$ vs.\ $48.86\%$), while fusing only logits (\textbf{w/o Eq.~\eqref{eq:gsdf_emb}}) loses $-1.05$~pp U-CAPC ($-1.47$~pp UOT-CAPC; $-3.01$~pp on \textbf{ResNet1D} and $-3.26$~pp on \textbf{MCCNN}). Crucially, omitting per-frame logit $z$-standardization~\eqref{eq:logit_znorm} hurts plug-in generality across hosts whose clean logit standard deviations span $\sigma \in [1.30, 9.30]$ (Fig.~\ref{fig:module_ablations}(b)), costing $\mathbf{-2.88}$~pp on \textbf{ResNet1D} ($45.98\%$ vs.\ $48.86\%$), $-1.47$~pp on \textbf{MTL-DANN}, and $-1.17$~pp on \textbf{MTL-CORAL}. Notably, unwarped \textbf{Static Embedding Concatenation} ($\sqrt{\lambda_{\mathrm{null}}}\hat{\mathbf{z}}_{\mathrm{null}}$ without $\sqrt{g_n}$ in~\eqref{eq:gsdf_emb}) raises clean/multipath U-CAPC on moderate-SNR \textbf{MobRFFI} ($50.29\%$), \emph{but violates Proposition~\ref{prop:plugin_fallback} (Claim~3)}, leaking isotropic null-bin noise into spherical prototypes at $10/5$~dB AWGN across all three datasets ($43.47\%$ vs.\ $43.88\%$, up to $-0.61$~pp on \textbf{MobRFFI}); gate-warping~\eqref{eq:gsdf_emb} is therefore essential for exact zero-negative-transfer guarantees.

\subsubsection{Pillar~3: Eight-Backbone Gate Ablation}
Table~\ref{tab:ablation_core} (Pillar~3), \supptabIII{} (Panel~A), and Fig.~\ref{fig:module_ablations}(c) evaluate eight label-free gating rules across all eight backbones. Unweighted \textbf{Static-Sum} ($g_n \equiv 1$) achieves $+27.20$~pp multipath gain on \textbf{DataPort-WiFi} at the cost of a $-12.33$~pp 8-host mean collapse at 10~dB AWGN ($-30.23$~pp on \textbf{IB-Host}, $-20.58$~pp on \textbf{ResNet1D}, $-13.28$~pp on \textbf{SigMix}) and $-7.15$~pp on \textbf{MobRFFI} \texttt{19-19}. Prediction-driven \textbf{Entropy-Gate} still suffers $-2.73$~pp mean ($-11.09$~pp on \textbf{IB-Host}) from overconfident noise posteriors, while \textbf{Branch-Agree} sacrifices $70\%$ of multipath gains ($+8.16$~pp vs.\ $+26.43$~pp). Capture-level damping variants (\supptabIII) over-suppress clean captures ($44.3\%\text{--}65.6\%$ worst-slice retention) and are reported as negative results. Only \textbf{Lag-64 Thermal-SNR} achieves $\mathbf{0.00}$~pp low-SNR loss on all $8/8$ hosts while retaining $97\%$ of \textbf{Static-Sum}'s multipath rescue ($+26.43$~pp mean; $+35.21$~pp on \textbf{IB-Host}).

\subsubsection{Pillar~4: \texorpdfstring{UOT-CAPC $\to$ U-CAPC}{UOT-CAPC to U-CAPC} Decomposition and Edge Source-Free TTA Comparison}
Figure~\ref{fig:module_ablations}(d) and \supptabIII{} (Panel~B1) trace Algorithm~\ref{alg:ucapc} step by step on the 9-run grid. Pooling frames across sessions (\textbf{Cross-Session Global Pool}) loses $-3.10$~pp versus per-capture calibration ($-6.39$~pp on \textbf{MobRFFI}), confirming that channel/front-end shifts are capture-specific. Within each capture $\mathcal{G}$, \textbf{Proto-Only} ($\rho \to 0$) excels on skewed \textbf{WiSig} ($52.72\%$) but underperforms on balanced \textbf{DataPort-WiFi}/\textbf{MobRFFI} without marginal anti-collapse regularization, whereas \textbf{Sinkhorn-Only} ($T=0$, $\rho=1$) lacks iterative spherical prototype alignment ($55.51\%$). General \textbf{UOT-CAPC} ($\rho^* < 1, T=3$) leads all methods on skewed \textbf{WiSig} ($\mathbf{52.80\%}$) and small windows ($N_{\mathcal{G}}=10$: $96.47\%$ vs.\ $62.83\%$ U-CAPC), while its $\rho=1$ specialization (\textbf{Balanced U-CAPC}) converges by $T=3$ ($56.99\% \to 57.37\% \to \mathbf{57.50\%}$). Dropping Step~6's initial-logit anchor $\mathbf{L}^{(0)}$ boosts clean shift adaptation on \textbf{MobRFFI} ($51.56\%$) but degrades $10/5$~dB noise stability by $-1.79$~pp on \textbf{DataPort-WiFi} and $-2.31$~pp on \textbf{MobRFFI}. Disaggregating the source-calibrated 27-slice TTA suite across datasets and backbones (Table~\ref{tab:ablation_core}, Pillar~4; \supptabIII, Panel~B2; Fig.~\ref{fig:module_ablations}(d)) highlights a decisive advantage on cross-AP Wi-Fi IoT testbeds: on \textbf{ORBIT MobRFFI}, closed-form \textbf{Balanced U-CAPC} reaches $\mathbf{52.00\%}$ ($\#1$ overall, $<0.08$~ms/frame, $0$ backward passes), outperforming 5-epoch gradient-updated \textbf{SHOT-IM} ($47.29\%$, $+4.71$~pp), \textbf{SHOT-Full} ($45.33\%$, $\mathbf{+6.67}$~pp), \textbf{LAME} ($44.08\%$, $+7.92$~pp), and \textbf{T3A} ($42.98\%$, $+9.02$~pp), while also lifting $4.72\text{M}$ \textbf{McAFF} on \textbf{MobRFFI} by $\mathbf{+18.69}$~pp ($26.59\% \to 45.28\%$, beating 5-epoch \textbf{McAFF + SHOT-IM} at $43.77\%$). On naturally skewed \textbf{WiSig}, forward-only \textbf{UOT-CAPC} reaches $\mathbf{55.49\%}$ ($\#1$ gradient-free, beating \textbf{LAME} at $53.53\%$, \textbf{SHOT-IM} at $53.13\%$, and \textbf{T3A} at $49.89\%$), whereas uncalibrated default SHOT ($\text{lr}=10^{-2}$) collapses to $9.6\%\text{--}28.1\%$.

\subsubsection{Symmetric Channel-Augmentation Fairness Audit and Competitor Boundaries}
\label{sec:fairness_controls}
To audit augmentation sensitivity without retraining redundant variants within the same backbone family, Fig.~\ref{fig:snr_mp}(i) and \supptabIV{} test one representative model per paradigm---including our own \textbf{IB-Host}---under unaugmented published training versus matched multi-condition channel augmentations. Omitting channel augmentations severely degrades single-source cross-domain transfer across all families (\textbf{IB-Host} [TC-regularized checkpoint] drops $-41.06$~pp on \textbf{WiSig} Day~8 and $-56.59$~pp on \textbf{MobRFFI} \texttt{RX 20-1}; \textbf{McAFF} drops $-45.50$ and $-54.45$~pp; \textbf{MTL-CORAL} drops $-47.28$~pp), confirming why every baseline in Table~\ref{tab:main_results} is trained with identical matched augmentations. Under this symmetric protocol, two standalone competitor strengths define clear boundaries: (i) on same-day \textbf{DataPort-WiFi} $E_2,E_3\!\to\!E_1$ clean transfers, CFO-assisted \textbf{Xie-Hybrid + U-CAPC}~\cite{xie2024robustrffingerprintidentification} reaches $99.51 \pm 0.03\%$ vs.\ $98.43\%$ for CFO-free \textbf{IB-Host + GSDF} ($+1.08$~pp), as \textbf{GSDF} omits static crystal CFO priors that drift across weeks~\cite{cfo_experimental_review}; and (ii) at $51\text{--}57\times$ larger size, standalone \textbf{McAFF} attains higher three-benchmark $\mathrm{F1}_{\mathrm{mp}}$ ($58.27\%$ vs.\ $55.65\%$) and \textbf{WiSig} balanced U-CAPC ($51.74\%$ vs.\ $50.64\%$) than edge \textbf{IB-Host + GSDF}, at the cost of $-23.41 / -10.41$~pp lower $\mathrm{F1}_{\mathrm{noise}}$ ($14.16\% / 33.47\%$ vs.\ $37.57\% / 43.88\%$)---while attaching \textbf{+GSDF} directly onto \textbf{McAFF} lifts it further to $55.63\%$ three-benchmark $\overline{\mathrm{F1}}$ ($53.71\%$ on \textbf{WiSig}, Table~\ref{tab:track_p}).

\subsection{RQ4: Real-World IoT Deployment Scaling---Multi-Week Drift, Population (\texorpdfstring{$K{=}10\text{--}47$}{K=10-47}), Timing Synchronization, and Session Bursts}
\label{sec:wisig_mob_deepdive}

\begin{figure*}[!t]
\centering
\includegraphics[width=0.81\textwidth]{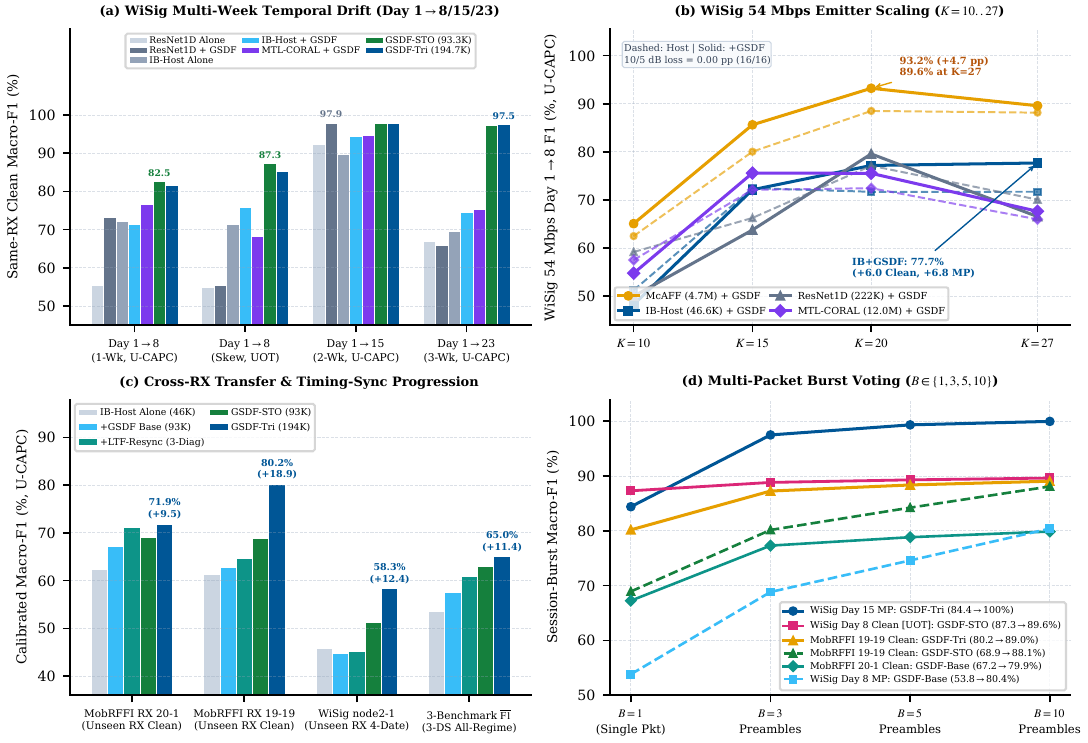}
\caption{Extended \textbf{WiSig} and \textbf{ORBIT MobRFFI} multi-regime IoT scaling analysis ($n=9$ runs across $3\times 3$ source-split $\times$ init-seed grid; uniform calibration metric within every group): (a) same-RX \textbf{WiSig} multi-week temporal drift (\texttt{Day 1$\to$8 [U-CAPC]}, skewed \texttt{Day 1$\to$8 [UOT]}, \texttt{Day 1$\to$15 [U-CAPC]}, \texttt{Day 1$\to$23 [U-CAPC]}) from standalone hosts (\textbf{ResNet1D}, \textbf{IB-Host}) through zero-retrain \textbf{+GSDF}, timing-synchronized \textbf{GSDF-STO} ($93.3\text{K}$), and tri-branch \textbf{GSDF-Tri} ($194.7\text{K}$); (b) \textbf{WiSig} 54~Mbps emitter-population scaling across $K \in \{10, 15, 20, 27\}$ comparing standalone hosts (dashed) and zero-retrain \textbf{+GSDF} (solid, $0.00$~pp $10/5$~dB AWGN loss across all $16$ cells); (c) five-step cross-receiver transfer and timing-synchronization progression under uniform \textbf{U-CAPC} across unseen \textbf{MobRFFI} \texttt{RX 20-1}, \texttt{RX 19-19}, 4-date unseen \textbf{WiSig} \texttt{node2-1} clean mean, and three-benchmark $\overline{\mathrm{F1}}$ (\textbf{IB-Host Alone} $\to$ \textbf{+GSDF Base} $\to$ \textbf{+LTF-Resync} [3-diag] $\to$ \textbf{GSDF-STO} $\to$ \textbf{GSDF-Tri}); (d) zero-label multi-packet session burst log-posterior aggregation across $B \in \{1, 3, 5, 10\}$ consecutive preambles per burst.}
\label{fig:wisig_mob_ext}
\end{figure*}

\begin{table*}[!t]
\centering
\caption{Extended \textbf{WiSig} and \textbf{ORBIT MobRFFI} Multi-Regime IoT Scaling Summary ($3\times 3$ 9-Run Grid)}
\label{tab:wisig_mob_summary}
\scriptsize
\renewcommand{\arraystretch}{0.91}
\setlength{\tabcolsep}{4.5pt}
\begin{tabular}{llcccccc}
\toprule
\multicolumn{8}{l}{\textbf{Panel A: Zero-Retrain \textbf{+GSDF} Across \textbf{WiSig} Multi-Week Drift (\texttt{node1-1}, $K{=}15$) and 54~Mbps Scaling ($K \in \{20, 27\}$)}} \\
\midrule
Model / Plug-In & Params & Day 8 Clean & Day 15 Clean & Day 15 MP & Day 23 Clean & 54M $K{=}20$ & 54M $K{=}27$ \\
\midrule
IB-Host Alone & 46.6K & 72.17 & 89.73 & 60.40 & 69.50 & 71.70 & 71.71 \\
ResNet1D~\cite{he2016deep} Alone & 221.6K & 55.34 & 92.38 & 48.39 & 67.01 & 77.12 & 70.07 \\
MTL-CORAL~\cite{sun2016deep} Alone & 12.03M & 72.76 & 84.55 & 58.04 & 71.43 & 72.45 & 65.98 \\
McAFF~\cite{multichannel_attentive_fusion} Alone & 4.72M & \secbest{77.87} & --- & --- & --- & \secbest{88.50} & \secbest{88.15} \\
\midrule
\rowcolor{plugbg} ResNet1D + \textbf{GSDF} & 268.2K & 73.26 & \textbf{97.94} & 54.88 & 65.78 & 79.57 & 66.55 \\
\rowcolor{plugbg} MTL-CORAL + \textbf{GSDF} & 12.08M & 76.73 & \secbest{94.72} & \textbf{73.62} & \textbf{75.23} & 75.53 & 67.64 \\
\rowcolor{oursbg} \textbf{McAFF + GSDF (Plug-In)} & 4.77M & \textbf{80.04} & --- & --- & --- & \textbf{93.21} & \textbf{89.58} \\
\rowcolor{oursbg} \textbf{IB-Host + GSDF (Ours)} & \textbf{93.3K} & 71.38$^{\mathrm{a}}$ & 94.54 & \secbest{67.29} & \secbest{74.49} & 77.17 & 77.69 \\
\midrule
\multicolumn{8}{l}{\textbf{Panel B: Zero-Retrain 3-Way Host Fusion and Packet-Trigger Timing Synchronization Across Three Benchmarks}} \\
\midrule
Model / Upgrade Arm & Params ($A{=}4/1$) & \textbf{DataPort} & \textbf{WiSig} & \textbf{MobRFFI} & \texttt{19-19} Clean & 3-DS (Direct / U-CAPC) & 3-DS \textbf{UOT-CAPC} \\
\midrule
GSDF-Base (Matched Aug.) & 92.6K / 93.3K & 73.71 & 50.64 & 48.16 & 62.70 & 51.58 / 57.50 & 55.41 \\
\rowcolor{plugbg} IB-Host + CORAL + NS (3-Way) & 12.12M & 73.77 & 54.30 & 49.36 & 66.58 & 52.01 / 59.14 & 55.66 \\
\rowcolor{plugbg} IB-Host + McAFF + NS (3-Way) & 5.39M / 4.82M & 75.83 & 55.07 & 45.29 & 48.94 & 53.44 / 58.73 & 56.46 \\
\rowcolor{plugbg} GSDF + LTF-Resync$^{\mathrm{b}}$ & 92.6K / 93.3K & 74.51 & 56.30 & 51.61 & 64.77 & 53.93 / 60.81 & \secbest{58.40} \\
\rowcolor{oursbg} \textbf{GSDF-STO (STO Jitter)} & \textbf{92.6K / 93.3K} & \secbest{76.40} & \secbest{58.83} & \secbest{53.72} & \secbest{68.92} & \secbest{56.64} / \secbest{62.98} & \textbf{60.96} \\
\rowcolor{oursbg} \textbf{GSDF-Tri (3-Branch Sync)} & 193.7K / 194.7K & \textbf{76.93} & \textbf{60.76} & \textbf{57.47} & \textbf{80.17} & \textbf{57.02} / \textbf{65.05} & \textbf{60.96} \\
\midrule
\multicolumn{8}{l}{\textbf{Panel C: Zero-Label Multi-Packet Session Burst Aggregation at $B{=}10$ Consecutive Preambles/Burst ($0$ Retrain, $0$ Labels)}} \\
\midrule
Session Burst Arm ($B{=}10$) & Params ($A{=}1/4$) & Day 8 [\textbf{UOT}] & Day 8 MP & Day 15 Clean & Day 15 MP & Day 23 MP & \texttt{19-19} Clean \\
\midrule
GSDF-Base ($B{=}10$ Burst) & 93.3K / 92.6K & 81.35 & 80.39 & 99.72 & 97.36 & 74.80 & 70.27 \\
\rowcolor{oursbg} \textbf{GSDF-STO ($B{=}10$ Burst)} & \textbf{93.3K / 92.6K} & \textbf{89.62} & \textbf{87.92} & \secbest{99.99} & \secbest{99.43} & \secbest{95.67} & \secbest{88.08} \\
\rowcolor{oursbg} \textbf{GSDF-Tri ($B{=}10$ Burst)} & 194.7K / 193.7K & \secbest{89.54} & \secbest{86.38} & \textbf{100.00} & \textbf{99.96} & \textbf{97.30} & \textbf{89.04} \\
\bottomrule
\end{tabular}\par
\vspace{2pt}{\scriptsize\noindent\raggedright \emph{Note:} Single-value regime/dataset columns report calibrated \textbf{U-CAPC} Macro-F1 (\%, except \textbf{[UOT]} on skewed Day~8 in Panel~C); full 8-model multi-week Direct/U-CAPC pairs, $K{=}10\text{--}47$ scaling curves, 2-RX source pooling, symmetric timing-tuned baselines, and $B \in \{1, 3, 5, 10\}$ trajectories are in \supptabsVtoVII. $^{\mathrm{a}}$Under general \textbf{UOT-CAPC} on skewed Day~8, \textbf{IB-Host + GSDF} reaches $\mathbf{75.75\%}$ (and $\mathbf{87.29\%}$ for \textbf{GSDF-STO}). $^{\mathrm{b}}$3-diagonal screening grid; all other rows are 9-run means (\textbf{GSDF-STO} and \textbf{GSDF-Tri} add $\Delta\tau \sim \mathcal{U}[-3,+3]$ symbol-timing jitter).\par}
\end{table*}

While Table~\ref{tab:main_results} enforces a conservative single-frame ($B=1$), 5-condition-averaged (including synthetic $20/10/5$~dB AWGN and $100/200$~ns multipath), single-source protocol, real-world IoT WLAN deployments also span unattended multi-week intervals, larger single-rate device populations, multi-AP source pooling, timing-synchronized APs, and multi-packet Wi-Fi bursts. Figure~\ref{fig:wisig_mob_ext} and Table~\ref{tab:wisig_mob_summary} (with complete disaggregated matrices in \supptabsVtoVII) examine these operational dimensions across \textbf{WiSig} and \textbf{ORBIT MobRFFI}:

\subsubsection{Multi-Week Unattended Temporal Transfer (\texorpdfstring{\texttt{Day 1$\to$8 / 15 / 23}}{Day 1 to 8 / 15 / 23})}
Across the eight-model multi-date suite on \textbf{WiSig} \texttt{node1-1} (Table~\ref{tab:wisig_mob_summary}, Panel~A; \supptabV; Fig.~\ref{fig:wisig_mob_ext}(a)), zero-retrain \textbf{+GSDF} consistently elevates both clean and multipath transfer over multi-week IoT operation without re-enrollment. On one-week transfer (\texttt{Day 1$\to$8}), balanced \textbf{U-CAPC} lifts \textbf{ResNet1D + GSDF} from $55.34\%$ to $\mathbf{73.26\%}$ ($+17.92$~pp) and \textbf{MTL-CORAL + GSDF} from $72.76\%$ to $\mathbf{76.73\%}$, while skewness-tolerant \textbf{UOT-CAPC} lifts \textbf{IB-Host + GSDF} from $71.33\%$ to $\mathbf{75.75\%}$ ($\mathbf{87.29\%}$ for \textbf{GSDF-STO}). On two-week transfer (\texttt{Day 1$\to$15}), \textbf{ResNet1D + GSDF} reaches $\mathbf{97.94\%}$ clean ($+5.56$~pp over \textbf{ResNet1D}), \textbf{MTL-CORAL + GSDF} jumps from $84.55\%$ to $\mathbf{94.72\%}$ clean ($\mathbf{+10.17}$~pp) and from $58.04\%$ to $\mathbf{73.62\%}$ multipath ($\mathbf{+15.58}$~pp), \textbf{MTL-SEI + GSDF} gains $\mathbf{+10.33}$~pp clean ($93.21\%$), and $93.3\text{K}$ \textbf{IB-Host + GSDF} rises from $89.73\%$ to $\mathbf{94.54 \pm 1.51\%}$ clean ($+4.81$~pp) and $60.40\% \to 67.29\%$ multipath ($+6.89$~pp). On three-week transfer (\texttt{Day 1$\to$23}), \textbf{MTL-CORAL + GSDF} ($75.23\%$ clean, $+10.36$~pp multipath), \textbf{MTL-DANN + GSDF} ($75.19\%$), and \textbf{IB-Host + GSDF} ($74.49\%$, $+4.99$~pp over \textbf{IB-Host}) all sustain $>74\%$ clean Macro-F1 from Day~1 enrollment alone.

\subsubsection{54~Mbps IoT Emitter-Population Scaling (\texorpdfstring{$K=10\text{--}47$}{K=10-47}), 2-AP Source Pooling, and Zero-Retrain 3-Way Fusion}
Expanding the single-rate 54~Mbps \textbf{WiSig} cohort from $K=10$ to $27$ emitters on Day~$1\!\to\!8$ (Table~\ref{tab:wisig_mob_summary}, Panel~A; \supptabV; Fig.~\ref{fig:wisig_mob_ext}(b)) shows that eliminating mixed modulation/rate heterogeneity substantially raises absolute accuracy across larger IoT device fleets: zero-retrain \textbf{McAFF + GSDF} climbs from $65.08\%$ ($K=10$) to $\mathbf{85.62\%}$ ($K=15$, $+5.58$~pp clean, $+6.89$~pp MP), peaks at $\mathbf{93.21\%}$ ($K=20$, $+4.71$~pp), and holds $\mathbf{89.58\%}$ at $K=27$, while edge $93.3\text{K}$ \textbf{IB-Host + GSDF} scales to $\mathbf{77.17\%}$ at $K=20$ ($+5.47$~pp) and $\mathbf{77.69\%}$ at $K=27$ ($\mathbf{+5.98}$~pp clean, $\mathbf{+6.84}$~pp MP, $\mathbf{56.28\%}$ $20/10/5$~dB noise F1 vs.\ $37.00\%$ for \textbf{McAFF + GSDF}, and $0.00$~pp $10/5$~dB AWGN loss across all $16$ cells). Auditing all 36 raw \texttt{node1-1} archives further confirms $0/36$ cross-date timestamp contamination across 187,123 natural 54~Mbps frames and $K_{\max}=47$ four-date emitters ($\ge 300$ frames/emitter/date): across five nested random emitter subsets over $K \in \{10, 15, 20, 27, 47\}$ averaged across Mar~8/15/23 (\supptabVII, Panel~B), \textbf{GSDF-Tri} sustains $\mathbf{73.05\%}$ U-CAPC ($\mathbf{72.17\%}$ UOT-CAPC, $\mathbf{66.51\%}$ Direct) clean and $\mathbf{64.91\%}$ U-CAPC ($\mathbf{65.07\%}$ UOT-CAPC) under $100$~ns multipath at $K=47$ (\textbf{GSDF-Base}: $66.83\%$ clean / $53.57\%$ MP [$55.68\%$ UOT]), outperforming $4.72\text{M}$ \textbf{McAFF*} ($63.94\%$ clean / $48.01\%$ MP U-CAPC; $46.89\% / 40.30\%$ Direct) and $185.6\text{K}$ \textbf{McAFF-Lite*} ($63.71\%$ / $47.97\%$ U-CAPC). Pooling two Day-1 source APs (\texttt{node1-1 + node2-1} on Mar~1, evaluated across 54 future cells on Mar~8/15/23; \supptabVII, Panel~C) lifts cross-day retention on those two APs to $\mathbf{93.72\%}$ U-CAPC ($\mathbf{93.65\%}$ 6-Seed Holdout, $\mathrm{SD}_{\mathrm{split}}=0.35\%$; $\mathbf{88.39\%}$ UOT-CAPC, $\mathbf{85.19\%}$ Direct; $\mathbf{94.91\%}$ on \texttt{node2-1}) clean and $\mathbf{85.55\%}$ under $100$~ns multipath for \textbf{GSDF-Tri (2-RX)} ($\mathbf{88.35\%}$ / $\mathbf{77.09\%}$ for $93.3\text{K}$ \textbf{GSDF-Base (2-RX)}), ahead of \textbf{MTL-CORAL* (2-RX)} ($90.33\%$ / $70.05\%$), \textbf{McAFF-Lite* (2-RX)} ($84.53\%$ / $72.38\%$), and \textbf{McAFF* (2-RX)} ($83.45\%$ / $73.47\%$), though this seen-AP protocol does not test zero-shot transfer to a third unseen hardware unit. Meanwhile, zero-retrain 3-way host fusion (Table~\ref{tab:wisig_mob_summary}, Panel~B) pushes three-benchmark $\overline{\mathrm{F1}}$ to $\mathbf{59.14\%}$ (\textbf{59.31\%} 6-Seed Holdout; $\mathbf{66.58\%}$ \texttt{RX 19-19} clean) for \textbf{IB-Host + MTL-CORAL + NS-Branch} and $\mathbf{58.73\%}$ ($\mathbf{56.46\%}$ UOT-CAPC) for \textbf{IB-Host + McAFF + NS-Branch}.

\subsubsection{Physical AP Packet-Trigger Timing Diagnosis and Synchronized Upgrade (\textbf{GSDF-STO} \& \textbf{GSDF-Tri})}
Why do single-frame cross-week \textbf{WiSig} (\texttt{Day 8}) and cross-AP \textbf{WiSig}/\textbf{MobRFFI} transfers plateau below same-day \textbf{DataPort-WiFi}? A zero-retrain symbol-timing sweep on \textbf{WiSig} \texttt{Day 8} exposes the root physical bottleneck: simply shifting test preambles by $\Delta n = -2$ samples \emph{without retraining} jumps \textbf{GSDF-Base} clean U-CAPC from $71.84\%$ ($\Delta n = 0$) to $\mathbf{83.16\%}$ ($\mathbf{+11.32}$~pp; \textbf{IB-Host}: $71.80\% \to 79.72\%$), proving that $1\text{--}2$~sample packet-detector trigger drift across days and replacement APs misaligns strided in-band convolutions. Addressing this bottleneck systematically yields a clear five-step progression across unseen APs and all-regime benchmarks (Table~\ref{tab:wisig_mob_summary}, Panel~B; \supptabVI; Fig.~\ref{fig:wisig_mob_ext}(a,c)---under uniform \textbf{U-CAPC}, strictly monotonic on \textbf{MobRFFI} \texttt{RX 19-19} [$61.23\% \to 62.70\% \to 64.77\% \to 68.92\% \to \mathbf{80.17\%}$] and 3-benchmark $\overline{\mathrm{F1}}$, and climbing monotonically from $44.71\% \to 45.20\% \to 51.35\% \to \mathbf{58.32\%}$ [$+12.43$~pp vs.\ $45.89\%$ \textbf{IB-Host Alone}; $49.43\% \to 52.71\% \to \mathbf{56.12\%}$ vs.\ $49.92\%$ under 9-run \textbf{UOT-CAPC}] on 4-date unseen \textbf{WiSig} \texttt{node2-1} once trigger drift is aligned): (i) test-time blind L-LTF cross-correlation resynchronization (\textbf{GSDF + LTF-Resync}, $0$ extra parameters, 3-diagonal screen) already lifts 3-benchmark $\overline{\mathrm{F1}}$ from $57.50\%$ to $\mathbf{60.81\%}$ ($58.11\%$ UOT \textbf{WiSig}, $51.61\%$ \textbf{MobRFFI}, $90.61\%$ STO $\pm 1,\pm 2$); (ii) training our exact same $92.6\text{K}\text{--}93.3\text{K}$ edge architecture ($1.82$~ms ONNX latency, phase-free null magnitudes) with sub-sample fractional symbol-timing-offset jitter $\Delta\tau \sim \mathcal{U}[-3,+3]$ (\textbf{GSDF-STO}) accounts for $72.6\%$ of the total upgrade gain, surging 9-run $\overline{\mathrm{F1}}$ to $\mathbf{62.98\%}$ (\textbf{63.07\%} 6-Seed Holdout, $+5.48$~pp) and lifting \textbf{WiSig} to $\mathbf{58.83\%}$ U-CAPC / $\mathbf{60.89\%}$ UOT ($\mathbf{87.29\%}$ Day~8 UOT, $\mathbf{97.86\%}$ Day~15, $\mathbf{97.35\%}$ Day~23; $\mathbf{92.59\%}$ 3-date clean mean) and \textbf{MobRFFI} to $\mathbf{53.72\%}$ ($68.99\%$ clean mean); and (iii) adding a compact lag-16/64 complex-autocorrelation and spectral auxiliary expert ($+101\text{K}$ active params; $273.8\text{K}\text{--}274.8\text{K}$ raw module params) with source temperature-calibrated CFO-derotated log-probability gating (eliminating hand-set constants $(1.15, 2.5)$ while trading off phase-free residual-CFO gate closure at large $\Delta f$) yields tri-branch \textbf{GSDF-Tri} ($193.7\text{K}\text{--}194.7\text{K}$ active params, $2.91$~ms 1-thread ORT CPU), matching \textbf{GSDF-STO} under prior-relaxed \textbf{UOT-CAPC} ($60.96\%$) while lifting balanced \textbf{U-CAPC} to $\mathbf{65.05\%}$ (\textbf{65.10\%} 6-Seed Holdout, $+7.55$~pp, 95\% CI $[+7.49, +7.63]$~pp), $\mathbf{80.17 \pm 2.20\%}$ on unseen \textbf{MobRFFI} \texttt{RX 19-19} clean ($71.89\%$ on \texttt{RX 20-1}, $\mathbf{57.47\%}$ 10-condition mean), and $\mathbf{58.32\%}$ U-CAPC ($\mathbf{56.12\%}$ UOT-CAPC) 4-date unseen \textbf{WiSig} \texttt{node2-1} clean mean ($\mathbf{71.59\%}$ on Mar~23). To preserve strict augmentation symmetry in Table~\ref{tab:main_results}, \textbf{GSDF-STO} and \textbf{GSDF-Tri} are paired directly against symmetrically timing-tuned external baselines and capacity-matched \textbf{McAFF-Lite} in Section~\ref{sec:gates} and \supptabVII.

\subsubsection{Zero-Label Multi-Packet IoT Session Burst Aggregation (\texorpdfstring{$B \in \{1, 3, 5, 10\}$}{B in 1, 3, 5, 10})}
In practical IoT Wi-Fi APs, a telemetry or control exchange (RTS/CTS, sensor data burst, Block-ACK) naturally delivers $B=3\text{--}10$ consecutive preambles from the same IoT client within milliseconds. Averaging per-frame calibrated log-posteriors across $B$ consecutive same-emitter preambles within each capture ($0$ retraining, $0$ target labels; Table~\ref{tab:wisig_mob_summary}, Panel~C; \supptabVI; Fig.~\ref{fig:wisig_mob_ext}(d)) reveals a sharp physical dichotomy between ergodic per-packet perturbations and static hardware bias. Under $100/200$~ns Rayleigh multipath and thermal noise, independent per-packet fading realizations average out rapidly: even on baseline $93.3\text{K}$ \textbf{GSDF-Base}, $B=1 \to 10$ boosts \textbf{WiSig} Day~8 multipath by $\mathbf{+26.58}$~pp ($53.81\% \to 80.39\%$), Day~15 multipath by $\mathbf{+30.07}$~pp ($67.29\% \to 97.36\%$, Clean $\mathbf{99.72\%}$), and \textbf{MobRFFI} \texttt{RX 20-1} clean from $67.24\%$ to $\mathbf{79.86\%}$ ($+12.62$~pp). Combining timing synchronization with burst voting yields near-ceiling cross-day and cross-AP reliability: at $B=5\text{--}10$, $93.3\text{K}$ \textbf{GSDF-STO} reaches $\mathbf{89.62\%}$ UOT ($\mathbf{90.60\%}$ U-CAPC) on \textbf{WiSig} Day~8, $\mathbf{99.99\%}$ clean / $\mathbf{99.43\%}$ MP on Day~15, $\mathbf{99.20\%}$ clean / $\mathbf{95.67\%}$ MP on Day~23, and $\mathbf{88.08\%}$ clean / $\mathbf{76.36\%}$ MP on unseen \textbf{MobRFFI} \texttt{RX 19-19}; \textbf{GSDF-Tri} reaches $\mathbf{100.00\%}$ clean / $\mathbf{99.96\%}$ MP on Day~15, $\mathbf{99.98\%}$ clean / $\mathbf{97.30\%}$ MP on Day~23, $\mathbf{73.33\%}$ on unseen \texttt{node2-1} (Mar~23), and $\mathbf{89.04\%}$ clean / $\mathbf{80.31\%}$ MP on \textbf{MobRFFI} \texttt{RX 19-19}. Conversely, on clean unseen \texttt{node2-1}, $B=1 \to 10$ adds $+0.2\text{--}1.8$~pp because static cross-RX RF front-end distortion is deterministic across frames, underscoring why architectural/timing alignment (\textbf{GSDF-STO} / \textbf{GSDF-Tri}, $+8.9\text{--}25.2$~pp) and multi-packet aggregation are complementary.

\subsection{RQ5: Natural OTA SNR Quintiles, Capture Windows, Hardware Impairments, and Edge Gateway CPU Latency}
\label{sec:gates}

\begin{figure*}[!t]
\centering
\includegraphics[width=0.76\textwidth]{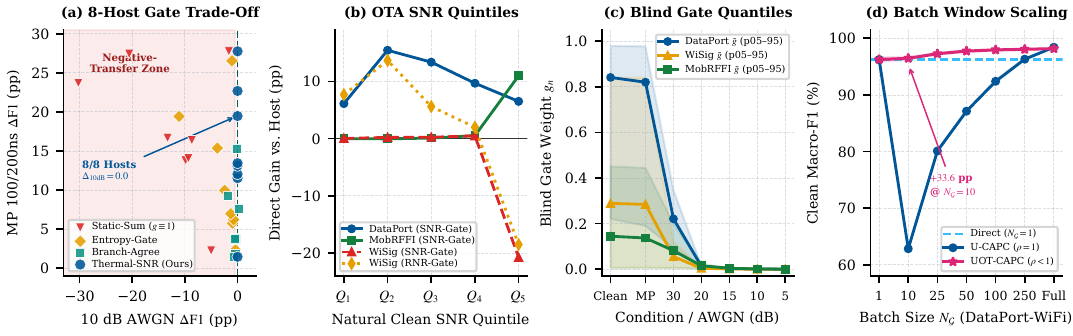}
\caption{Multi-backbone gate ablation, real OTA SNR audit, and batch-window scaling: (a) across all eight RFFI backbones, $100/200$~ns multipath U-CAPC gain vs.\ $10$~dB AWGN Direct negative transfer; (b) single-frame Direct gain over \textbf{IB-Host} across unmodified clean frames binned into natural OTA SNR quintiles ($Q_1$--$Q_5$, zero synthetic AWGN); (c) 5th--95th percentile range of blind gate $g_n$ across three benchmarks; (d) \textbf{DataPort-WiFi} clean Macro-F1 across capture batch sizes $N_{\mathcal{G}} \in \{1, 10, 25, 50, 100, 250, \mathrm{Full}\}$ comparing \textbf{GSDF} \textbf{Direct}, \textbf{U-CAPC} ($\rho=1$), and \textbf{UOT-CAPC} ($\rho < 1$).}
\label{fig:gates}
\end{figure*}

\begin{figure*}[!t]
\centering
\includegraphics[width=0.84\textwidth]{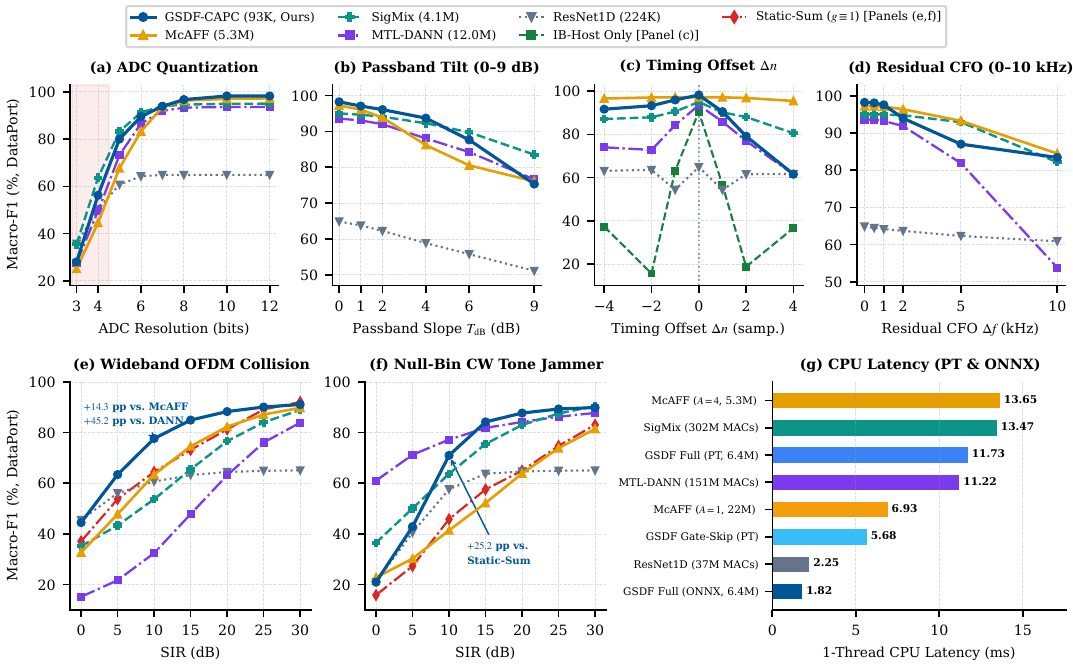}
\caption{Complete multi-baseline evaluation under receiver hardware impairments, co-channel/narrowband interference, and single-core edge CPU latency ($n=9$ runs on \textbf{DataPort-WiFi}, displaying all five core paradigms across panels (a)--(f)): (a) ADC resolution $B_{\mathrm{ADC}} \in [3,12]$~bits; (b) linear passband filter tilt $T_{\mathrm{dB}} \in [0,9]$~dB; (c) integer timing offset $\Delta n \in [-4,+4]$ samples (including strided \textbf{IB-Host}); (d) residual CFO $\Delta f \in [0,10]$~kHz; (e) wideband co-channel OFDM packet collision and (f) narrowband null-bin CW jammer across $\mathrm{SIR} \in [0,30]$~dB (including unweighted \textbf{Static-Sum}); (g) 1-thread CPU batch-1 latency comparing PyTorch eager (\texttt{PT}), conditional gate-skipping, and ONNX Runtime.}
\label{fig:impairments}
\end{figure*}

\subsubsection{Per-Backbone Gate Boundary, Natural OTA SNR Quintiles, and Capture-Window Scaling}
Complementing Table~\ref{tab:ablation_core} (Pillar~3) and \supptabIII, Fig.~\ref{fig:gates}(a,c) resolves the per-backbone trade-off between $100/200$~ns multipath U-CAPC gain and $10$~dB AWGN Direct negative transfer across all eight hosts: whereas \textbf{Static-Sum} loses $-1.64$ to $-30.23$~pp at 10~dB ($-9.88$~pp on \textbf{MTL-DANN}, $-13.28$~pp on \textbf{SigMix}, $-20.58$~pp on \textbf{ResNet1D}, $-30.23$~pp on \textbf{IB-Host}), \textbf{Entropy-Gate} loses up to $-11.09$~pp, and \textbf{Branch-Agree} forfeits $52\%\text{--}94\%$ of per-host multipath gains, \textbf{Thermal-SNR}~\eqref{eq:snr_sigmoid} pins every host to $\Delta_{10\mathrm{dB}} = 0.00$~pp while adapting its 5th--95th percentile gate distribution cleanly across regimes (Fig.~\ref{fig:gates}(c)). Confirming Proposition~\ref{prop:plugin_fallback} (Claim~2) on real IoT captures with naturally mixed per-frame SNRs (zero synthetic AWGN), binning unmodified OTA clean frames into natural SNR quintiles ($Q_1$--$Q_5$, Fig.~\ref{fig:gates}(b)) shows that \textbf{GSDF} Direct beats \textbf{IB-Host} across all quintiles on \textbf{DataPort-WiFi} ($+6.11$ to $+15.43$~pp) and gains $+10.97$~pp in $Q_5$ on \textbf{MobRFFI} \texttt{RX 20-1} while isolating low-SNR frames in $Q_1$--$Q_3$ at $0.00$~pp loss. On \textbf{WiSig} Day~8, $Q_1$--$Q_4$ gain $+0.04$ to $+0.49$~pp while $Q_5$ drops $-20.75$~pp ($61.65\%$ vs.\ $82.40\%$) under close-range cross-week receiver compression; passive-bin regrowth-to-noise gating (\textbf{RNR-Gate}, \supptabVI) lifts $Q_1$--$Q_4$ by $+1.99$ to $+13.66$~pp over \textbf{IB-Host} and narrows $Q_5$ by $+2.26$~pp ($63.91\%$). Across capture windows $N_{\mathcal{G}}$ on \textbf{DataPort-WiFi} (Fig.~\ref{fig:gates}(d)), multinomial class fluctuations at $N_{\mathcal{G}}=10\text{--}50$ violate strict equipartition in balanced \textbf{U-CAPC} ($\rho=1$: $62.83\%\text{--}87.13\%$), whereas general KL-relaxed \textbf{UOT-CAPC}~\eqref{eq:uot_variational} ($\rho^* < 1$) improves monotonically over \textbf{Direct} ($96.26\% \to 96.47\%$ at $N_{\mathcal{G}}=10$ [$+33.64$~pp over \textbf{U-CAPC}], $97.74\%$ at $N_{\mathcal{G}}=50$) before both converge on full sessions ($98.42\%$ vs.\ $98.18\%$).

\subsubsection{AP Receiver Hardware Impairments and Co-Channel / CW Interference}
\label{sec:impairments}
Figure~\ref{fig:impairments}(a)--(f) evaluates \textbf{GSDF-CAPC} against the full baseline suite (\textbf{McAFF}, \textbf{SigMix}, \textbf{MTL-DANN}, and \textbf{ResNet1D}) across four low-cost AP hardware impairments and two congested ISM-band interference regimes on \textbf{DataPort-WiFi}:
\begin{enumerate}[label=\alph*),leftmargin=*]
\item \emph{ADC quantization and passband filter tilt (Fig.~\ref{fig:impairments}(a,b)):} Across $B_{\mathrm{ADC}} \in [8, 12]$~bits, \textbf{GSDF-CAPC} ($96.72\%\text{--}98.24\%$) leads \textbf{McAFF} ($96.27\%\text{--}97.23\%$), \textbf{SigMix} ($94.58\%\text{--}94.99\%$), \textbf{MTL-DANN} ($93.29\%\text{--}93.63\%$), and \textbf{ResNet1D} ($64.81\%\text{--}64.87\%$); at coarse $3\text{--}6$~bit quantization where quantization harmonics raise the null-bin floor, \textbf{SigMix} edges ahead ($63.63\%$ vs.\ $56.20\%$ at 4-bit), setting $B_{\mathrm{ADC}} \ge 6$~bits ($89.52\%$) as the practical IoT AP ADC floor. Under linear passband tilt (Fig.~\ref{fig:impairments}(b)), \textbf{GSDF-CAPC} leads all methods over $0\text{--}4$~dB ($98.24\% \to 93.69\%$, $+7.52$~pp over \textbf{McAFF} at 4~dB), whereas under extreme $6\text{--}9$~dB front-end slope \textbf{SigMix}'s cross-receiver mixing retains $89.74\%\text{--}83.51\%$ vs.\ \textbf{GSDF-CAPC}'s $87.62\%\text{--}75.30\%$.
\item \emph{Timing offset and residual CFO (Fig.~\ref{fig:impairments}(c,d)):} Directly demonstrating Section~\ref{sec:encoders}'s DFT shift-invariance ($|e^{-j2\pi m\Delta n/64}|=1$ within the cyclic prefix), at $\Delta n = -2$ \textbf{GSDF-CAPC} rescues strided \textbf{IB-Host} from $15.59\%$ to $\mathbf{93.28\%}$ ($\mathbf{+77.69}$~pp), surpassing \textbf{SigMix} ($87.93\%$), \textbf{MTL-DANN} ($72.80\%$), and \textbf{ResNet1D} ($63.55\%$); at late overrun $\Delta n = +4$ (where payload samples enter L-LTF), stride-1 \textbf{McAFF} retains $95.49\%$ vs.\ $61.57\%$, motivating standard early-biased timing ($\Delta n \le 0$). Under $\Delta f = 10$~kHz residual CFO (Fig.~\ref{fig:impairments}(d)), inter-carrier interference drops spectral \textbf{MTL-DANN} to $53.83\%$, whereas lag-64 phase rotation $e^{j2\pi\Delta f(64/f_s)}$ conservatively closes $\bar{g}$ ($0.854 \to 0.014$) so \textbf{GSDF-CAPC} retains $\mathbf{83.44\%}$ ($+29.61$~pp over \textbf{MTL-DANN}, matching \textbf{McAFF}'s $84.50\%$ and \textbf{SigMix}'s $82.17\%$).
\item \emph{Wideband OFDM collision vs.\ narrowband CW jamming (Fig.~\ref{fig:impairments}(e,f)):} Comparing the two interference panels reveals complementary spectral vulnerabilities in shared 2.4/5~GHz IoT bands. Under $10$~dB SIR asynchronous wideband OFDM collisions (Fig.~\ref{fig:impairments}(e)), non-periodic packet overlap inflates $\hat{\sigma}_{w,a}^2$~\eqref{eq:median_noise} and throttles $g_n$, enabling \textbf{GSDF-CAPC} ($\mathbf{77.66\%}$) to beat unweighted \textbf{Static-Sum} ($64.47\%$, $+13.19$~pp), \textbf{McAFF} ($63.41\%$, $\mathbf{+14.25}$~pp), \textbf{ResNet1D} ($60.70\%$, $+16.96$~pp), \textbf{SigMix} ($53.71\%$, $+23.95$~pp), and \textbf{MTL-DANN} ($32.42\%$, $\mathbf{+45.24}$~pp). Conversely, under a single-tone null-bin CW jammer (Fig.~\ref{fig:impairments}(f)), jammer energy falls into a single DFT bin: full-band log-spectral \textbf{MTL-DANN} therefore retains $77.23\%$ at 10~dB SIR despite its $32.42\%$ wideband collision failure, while our median gate throttles $\bar{g}$ to $0.038$ so \textbf{GSDF-CAPC} holds $\mathbf{70.99\%}$ at 10~dB SIR ($\mathbf{+25.17}$~pp over \textbf{Static-Sum}'s $45.82\%$, $\mathbf{+29.58}$~pp over \textbf{McAFF}'s $41.41\%$, and $+7.26$~pp over \textbf{SigMix}'s $63.73\%$) and leads all baselines across $\mathrm{SIR} \ge 15$~dB ($84.24\%\text{--}89.89\%$). Protocol-wise, the 128-bin partition targets single-stream IEEE 802.11a/g preambles, and uniform \textbf{U-CAPC} targets closed-set $N_{\mathcal{G}} \ge 50$ balanced captures.
\end{enumerate}

\subsubsection{Edge Gateway Complexity: PyTorch Dispatch vs.\ 1.82~ms ONNX Runtime}
Implementing the four 64-point DFTs as fixed real matrix multiplications ($0.13\text{M}$ MACs / $0.26$~MFLOPs on top of $8.71 + 3.89 = 12.60$~MFLOPs of neural encoder operations) requires only \textbf{6.43M MACs} ($12.86$~MFLOPs at $A=4$; $7.12\text{M}$ MACs at $A=1$)---$6.4\times$ fewer MACs than $A=4$ \textbf{McAFF} ($3.1\times$ at $A=1$), $24\times$ fewer than \textbf{MTL-DANN/CORAL}, and $47\times$ fewer than \textbf{SigMix} (Table~\ref{tab:baseline_specs}, Fig.~\ref{fig:impairments}(g)). In eager PyTorch ($\text{batch}=1$, 1 CPU thread), sequential Python kernel launches across two Transformer/graph-attention branches incur $11.73$~ms ($A=4$), which conditional gate-skipping ($g_n < 0.05$) halves to $5.68$~ms. Exporting the unified graph to static ONNX Runtime eliminates Python dispatch overhead and executes full \textbf{GSDF} (and \textbf{GSDF-STO}) on a single CPU thread in \textbf{1.82~ms per frame} ($\approx 550$~packets/s without GPU acceleration)---and 3-branch \textbf{GSDF-Tri} in \textbf{2.91~ms}---achieving $7.5\times$ lower latency than $A=4$ \textbf{McAFF} and $7.4\times$ lower than \textbf{SigMix} (with $<0.08$~ms/frame amortized \textbf{U-CAPC}/\textbf{UOT-CAPC} overhead) and \textbf{100.00\% argmax parity} across all 224,640 test frames (Macro-F1 diff $\le 4.1\times 10^{-5}$).

\subsubsection{Symmetric Timing-Tuned Baselines (\textbf{McAFF*}, \textbf{MTL-CORAL*}) and Capacity-Matched \textbf{McAFF-Lite}}
To rule out augmentation or capacity asymmetry behind \textbf{GSDF-STO} and \textbf{GSDF-Tri}, \supptabVII{} (Panel~A) screens \textbf{McAFF} ($4.72\text{M}\text{--}5.30\text{M}$) and \textbf{MTL-CORAL} ($11.90\text{M}$ active forward params; $12.03\text{M}$ total module params in Table~\ref{tab:baseline_specs}) unconditionally across the same four causal recipes---matched \texttt{M2} (retraining \textbf{MTL-CORAL} on \textbf{DataPort-WiFi} under the unified 50-epoch cosine channel-perturbed validation schedule, raising its \texttt{M2} baseline from $51.84\%/54.08\%$ in Table~\ref{tab:main_results} to $54.79\%/57.68\%$ and $38.30\%/44.75\%$ 3-benchmark mean while reusing the exact \textbf{WiSig}/\textbf{MobRFFI} checkpoints), STO jitter (\texttt{+STO}, $\Delta\tau \sim \mathcal{U}[-3,+3]$), label-free L-LTF resynchronization (\texttt{+LTF-Resync}), and extended noise/multipath (\texttt{+NoiseMP})---under identical 3-diagonal selection and low-SNR / \texttt{19-19} / $\mathrm{STO}_{\pm 1,\pm 2}$ guardrails, alongside a capacity-matched \textbf{McAFF-Lite} ($185.6\text{K}$ active params at $A=1, K=15$ and $217.2\text{K}$ at $A=4, K=10$; $d_{\mathrm{host}}=128$; $1.12 / 2.45$~ms 1-thread CPU ORT; note that on 4-thread CPU or batched GPU, single-graph \textbf{McAFF-Lite}/\textbf{McAFF} execute faster than serial 3-branch \textbf{GSDF-Tri}). Across all 9 runs, guardrail-locked \textbf{McAFF*} (\texttt{M2}) and \textbf{McAFF + STO} attain $54.24\%$ and $54.35\%$ $\overline{\mathrm{F1}}$ U-CAPC ($54.21\%$ and $54.68\%$ 6-Seed Holdout; $45.58\%$ and $45.57\%$ UOT-CAPC; $43.37\%$ and $42.93\%$ Direct; on the 3-diagonal screen, unpromoted \texttt{McAFF + LTF-Resync} [$55.53\%$] and \texttt{McAFF + NoiseMP} [$57.65\%$] missed the strict $\mathrm{STO}_{\pm 1,\pm 2} \ge 88.14\%$ guardrail at $88.03\%$ and $84.07\%$), L-LTF-resynchronized \textbf{MTL-CORAL*} rises from $44.75\%$ (\texttt{M2}) and $46.01\%$ (\texttt{+STO}) to $47.29\%$ ($47.51\%$ 6-Seed Holdout; $41.13\%$ UOT-CAPC), and \textbf{McAFF-Lite*} (\texttt{M2}) / \textbf{McAFF-Lite + STO} reach $50.26\%$ / $52.02\%$ ($50.09\%$ / $51.61\%$ 6-Seed Holdout; $42.01\%$ / $43.49\%$ UOT-CAPC). Target-slice-clustered bootstrap 95\% CIs confirm that \textbf{GSDF-Tri} ($65.05\%$ U-CAPC, $60.96\%$ UOT-CAPC, $57.02\%$ Direct) leads \textbf{McAFF*} by $\mathbf{+10.81}$~pp U-CAPC ($[+10.21, +11.37]$~pp; $\mathbf{+15.38}$~pp UOT-CAPC, $\mathbf{+13.65}$~pp Direct), \textbf{McAFF-Lite*} by $\mathbf{+14.79}$~pp ($[+14.58, +15.02]$~pp; $+13.03$~pp over \textbf{McAFF-Lite + STO}), and \textbf{MTL-CORAL*} by $\mathbf{+17.76}$~pp ($[+17.10, +18.40]$~pp), while $2.3\times$ smaller \textbf{GSDF-STO} ($92.6\text{K}\text{--}93.3\text{K}$, $62.98\%$) beats \textbf{McAFF-Lite + STO} by $\mathbf{+10.96}$~pp U-CAPC ($\mathbf{+17.47}$~pp UOT-CAPC). Co-reporting slice-level competitor highs, $11.90\text{M}$ \textbf{MTL-CORAL*} attains higher balanced \textbf{U-CAPC} on clean \textbf{WiSig} Day~8 ($86.43\%$ vs.\ $81.71\%$ \textbf{GSDF-Tri}) and clean unseen \texttt{node2-1} Mar~1/Mar~8 ($62.35\% / 52.90\%$ vs.\ $53.75\% / 40.79\%$, where $3.53:1$ skew pulls balanced \textbf{U-CAPC} below \textbf{Direct} [$47.57\%$]), whereas \textbf{GSDF-Tri} leads on skewed \textbf{UOT-CAPC} ($85.24\%$, $64.61\%$, $49.46\%$ vs.\ $75.76\%$, $53.16\%$, $46.46\%$) and across multipath/low-SNR regimes.

\section{Conclusion}
\label{sec:conclusion}

We presented \textbf{GSDF-CAPC}, to the best of our knowledge the first IoT RFFI framework to dynamically modulate populated active in-band subcarriers and unpopulated null/guard-subcarrier spectral regrowth via zero-retrain, blind SNR-gated dual-space fusion and forward-only capture calibration on IEEE 802.11a/g preambles. Grounded in the quadratic $\mathcal{O}((\eta_{\mathrm{null}}\gamma)^2)$ low-SNR Fisher decay of unpopulated bins (Proposition~\ref{prop:snr_collapse}), \textbf{GSDF} pairs any frozen pretrained AP host with an independently trained 128-bin \textbf{NS-Branch} via scale-standardized logits and gate-warped spherical embeddings driven by a source-calibrated lag-64 thermal-SNR gate, guaranteeing projection-free compatibility, mixed-SNR noise isolation, and exact low-SNR fallback (Proposition~\ref{prop:plugin_fallback}). Over unlabeled edge AP captures, KL-regularized \textbf{UOT-CAPC} ($\rho = \gamma_{\mathrm{KL}}/(\gamma_{\mathrm{KL}}+1) \in (0,1)$) aligns shifted centroids under arrival skew and small windows, specializing at $\gamma_{\mathrm{KL}} \to \infty$ ($\rho=1$) to parameter-free balanced \textbf{U-CAPC}. Across \textbf{DataPort-WiFi}, \textbf{WiSig}, and \textbf{ORBIT MobRFFI}, zero-retrain \textbf{GSDF} won all $72/72$ paired runs across eight backbones ($p_{\mathrm{Holm}} < 10^{-4}$, scaling to $93.21\%$ on 54~Mbps \textbf{WiSig} at $K=20$ and $59.14\%$ via 3-way fusion), while backprop-free \textbf{U-CAPC}/\textbf{UOT-CAPC} surpassed 5-epoch gradient-based \textbf{SHOT} on cross-AP \textbf{MobRFFI} ($52.00\%$ vs.\ $45.33\%\text{--}47.29\%$). Resolving AP trigger timing drift (\textbf{GSDF-STO}, $62.98\%$ at $92.6\text{K}\text{--}93.3\text{K}$ params and $1.82$~ms ORT CPU latency; \textbf{GSDF-Tri}, $65.05\%$ 3-DS $\overline{\mathrm{F1}}$, $+10.81$ to $+17.76$~pp over symmetrically timing-tuned $185.6\text{K}\text{--}11.90\text{M}$ baselines, $73.05\%$ at $K_{\max}=47$, $93.72\%$ under 2-RX source pooling, and $80.17\%$ on \textbf{MobRFFI} \texttt{RX 19-19}) and zero-label burst voting ($B=3\text{--}10$, up to $99.9\%\text{--}100.0\%$ \textbf{WiSig} and $88.1\%\text{--}89.0\%$ \textbf{MobRFFI}) confirm real-time edge gateway operational reliability. Future work explores sparse bursts, open-set rejection, and MIMO preambles.

\bibliographystyle{IEEEtran}
\bibliography{references}

% Generated by IEEEtran.bst, version: 1.14 (2015/08/26)
\begin{thebibliography}{10}
\providecommand{\url}[1]{#1}
\csname url@samestyle\endcsname
\providecommand{\newblock}{\relax}
\providecommand{\bibinfo}[2]{#2}
\providecommand{\BIBentrySTDinterwordspacing}{\spaceskip=0pt\relax}
\providecommand{\BIBentryALTinterwordstretchfactor}{4}
\providecommand{\BIBentryALTinterwordspacing}{\spaceskip=\fontdimen2\font plus
\BIBentryALTinterwordstretchfactor\fontdimen3\font minus \fontdimen4\font\relax}
\providecommand{\BIBforeignlanguage}[2]{{%
\expandafter\ifx\csname l@#1\endcsname\relax
\typeout{** WARNING: IEEEtran.bst: No hyphenation pattern has been}%
\typeout{** loaded for the language `#1'. Using the pattern for}%
\typeout{** the default language instead.}%
\else
\language=\csname l@#1\endcsname
\fi
#2}}
\providecommand{\BIBdecl}{\relax}
\BIBdecl

\bibitem{brik2008paradis}
V.~Brik, S.~Banerjee, M.~Gruteser, and S.~Oh, ``Wireless device identification with radiometric signatures,'' in \emph{Proc. ACM Int. Conf. Mobile Comput. Netw. (MobiCom)}, 2008, pp. 116--127.

\bibitem{soltanieh2020review}
N.~Soltanieh, Y.~Norouzi, Y.~Yang, and N.~C. Karmakar, ``A review of radio frequency fingerprinting techniques,'' \emph{IEEE J. Radio Freq. Identif.}, vol.~4, no.~3, pp. 222--233, 2020.

\bibitem{jagannath2022survey}
A.~Jagannath, J.~Jagannath, and P.~S. P.~V. Kumar, ``A comprehensive survey on radio frequency ({RF}) fingerprinting: Traditional approaches, deep learning, and open challenges,'' \emph{Comput. Netw.}, vol. 219, p. 109455, 2022.

\bibitem{rffi_iot_auth}
J.~Zhang, G.~Shen, W.~Saad, and K.~Chowdhury, ``Radio frequency fingerprint identification for device authentication in the internet of things,'' \emph{IEEE Commun. Mag.}, vol.~61, no.~10, pp. 110--115, 2023.

\bibitem{zhu2022sigpro}
M.~Zhu, Z.~Feng, L.~Stankovi{\'c}, L.~Ding, J.~Fan, and X.~Zhou, ``A probe-feature for specific emitter identification using axiom-based {Grad-CAM},'' \emph{Signal Process.}, vol. 201, p. 108685, 2022.

\bibitem{better_data_not_bigger}
E.~Bothereau, R.~Gerzaguet, M.~Gautier, A.~Chillet, and O.~Berder, ``Why {RF} fingerprinting needs better data, not bigger models,'' \emph{IEEE Access}, vol.~13, pp. 171\,348--171\,355, 2025.

\bibitem{tifs_channel_receiver_independent}
J.~Ma, J.~Zhang, G.~Shen, L.~Peng, and A.~Marshall, ``Toward channel-robust and receiver-independent radio frequency fingerprint identification,'' \emph{IEEE Trans. Inf. Forensics Security}, vol.~20, pp. 12\,112--12\,125, 2025.

\bibitem{hanna2022wisig}
S.~Hanna, S.~Karunaratne, and D.~Cabric, ``{WiSig}: A large-scale {WiFi} signal dataset for receiver and channel agnostic {RF} fingerprinting,'' \emph{IEEE Access}, vol.~10, pp. 22\,808--22\,818, 2022.

\bibitem{shen2024mobrffi}
G.~Shen, J.~Zhang, A.~Marshall, R.~Woods, J.~R. Cavallaro, and L.~Chen, ``Towards receiver-agnostic and collaborative radio frequency fingerprint identification,'' \emph{IEEE Trans. Mobile Comput.}, vol.~23, no.~7, pp. 7618--7634, 2024.

\bibitem{xie2024robustrffingerprintidentification}
L.~Xie, L.~Peng, and J.~Zhang, ``Towards robust {RF} fingerprint identification using spectral regrowth and carrier frequency offset,'' in \emph{Proc. IEEE Conf. Comput. Commun. (INFOCOM)}, 2025, pp. 1--10.

\bibitem{ganin2016domain}
Y.~Ganin, E.~Ustinova, H.~Ajakan, P.~Germain, H.~Larochelle, F.~Laviolette, M.~Marchand, and V.~Lempitsky, ``Domain-adversarial training of neural networks,'' \emph{J. Mach. Learn. Res.}, vol.~17, no.~59, pp. 1--35, 2016.

\bibitem{sun2016deep}
B.~Sun and K.~Saenko, ``Deep {CORAL}: Correlation alignment for deep domain adaptation,'' in \emph{Proc. Eur. Conf. Comput. Vis. Workshops (ECCVW)}, 2016, pp. 443--450.

\bibitem{mtl_adversarial_sei}
F.~Wang, Z.~Cai, H.~Wan, T.~Tang, Q.~Wang, and G.~Gui, ``Toward robust receiver-invariant specific emitter identification via multi-task adversarial learning,'' \emph{IEEE Trans. Inf. Forensics Security}, vol.~21, pp. 4264--4274, 2026.

\bibitem{wan2025sigmix}
H.~Wan, F.~Shi, Y.~Wang, Q.~Xuan, Y.~Lin, and G.~Gui, ``{SigMix}: Robust specific emitter identification method enhanced by cross-time and cross-receiver mixing augmentation,'' \emph{IEEE Internet Things J.}, vol.~12, no.~12, pp. 21\,339--21\,349, 2025.

\bibitem{channel_temporal_attention_rff}
H.~Gu, L.~Su, Y.~Wang, W.~Zhang, and C.~Ran, ``Efficient channel-temporal attention for boosting {RF} fingerprinting,'' \emph{IEEE Open J. Signal Process.}, vol.~5, pp. 478--492, 2024.

\bibitem{cross_attention_transformer}
H.~Hui, C.~Wu, and J.~Yao, ``Cross-attention transformer for channel-robust radio frequency fingerprint identification,'' \emph{IEEE Sensors J.}, vol.~25, no.~19, pp. 36\,823--36\,831, 2025.

\bibitem{graph_conv_rff}
H.~Shen, X.~Gu, H.~Chen, C.~Chang, S.~Wang, T.~Tang, Y.~Lin, and G.~Gui, ``Efficient attention-enhanced graph convolutional network for radio frequency fingerprint identification,'' \emph{IEEE Internet Things J.}, vol.~13, no.~8, pp. 16\,496--16\,508, 2026.

\bibitem{equalization_da_rff}
R.~Pan, H.~Chen, H.~Chen, and W.-Q. Wang, ``Equalization-assisted domain adaptation for radio frequency fingerprint identification,'' \emph{IEEE Wireless Commun. Lett.}, vol.~13, no.~7, pp. 1868--1872, 2024.

\bibitem{multichannel_attentive_fusion}
Y.~Zeng, Y.~Gong, J.~Liu, S.~Lin, Z.~Han, R.~Cao, K.~Huang, and K.~B. Letaief, ``Multi-channel attentive feature fusion for radio frequency fingerprinting,'' \emph{IEEE Trans. Wireless Commun.}, vol.~23, no.~5, pp. 4243--4254, 2024.

\bibitem{cfo_experimental_review}
X.~Huan, Y.~Hao, K.~Miao, H.~He, and H.~Hu, ``Carrier frequency offset in internet of things radio frequency fingerprint identification: An experimental review,'' \emph{IEEE Internet Things J.}, vol.~11, no.~5, pp. 7359--7373, 2024.

\bibitem{liang2020shot}
J.~Liang, D.~Hu, and J.~Feng, ``Do we really need to access the source data? source hypothesis transfer for unsupervised domain adaptation,'' in \emph{Proc. Int. Conf. Mach. Learn. (ICML)}, 2020, pp. 6028--6039.

\bibitem{wang2021tent}
D.~Wang, E.~Shelhamer, S.~Liu, B.~Olshausen, and T.~Darrell, ``Tent: Fully test-time adaptation by entropy minimization,'' in \emph{Proc. Int. Conf. Learn. Represent. (ICLR)}, 2021.

\bibitem{prototype_uda_sei}
J.~Chen, L.~Yu, Y.~Chen, X.~Zheng, P.~Chen, and K.~Cheng, ``Prototype-driven unsupervised domain adaptation for specific emitter identification,'' \emph{IEEE Internet Things J.}, vol.~12, no.~1, pp. 637--649, 2025.

\bibitem{psfan}
Z.~Xiao, W.~Xiong, G.~Sun, and H.~Shao, ``{PSFAN}: Prototype-based source-free alignment network for cross-receiver specific emitter identification,'' \emph{IEEE Commun. Lett.}, vol.~30, pp. 907--911, 2026.

\bibitem{learnable_signal_representation}
Y.~Shao, J.~Liu, Y.~Zeng, and Y.~Gong, ``A radio frequency fingerprinting scheme using learnable signal representation,'' \emph{IEEE Commun. Lett.}, vol.~28, no.~1, pp. 73--77, 2024.

\bibitem{mrfe_multidimensional_rff}
Q.~Lu, Z.~Yang, H.~Zhang, F.~Chen, and H.~Xian, ``{MRFE}: A deep-learning-based multidimensional radio frequency fingerprinting enhancement approach for {IoT} device identification,'' \emph{IEEE Internet Things J.}, vol.~11, no.~18, pp. 30\,442--30\,454, 2024.

\bibitem{explainable_cnn_rff}
P.~Korycki, M.~Stojke, and K.~Kepa, ``Explainable deep learning for {RF} fingerprinting using convolutional neural networks,'' in \emph{Proc. Int. Conf. Dependability Comput. Syst. (DepCoS-RELCOMEX)}, ser. Lect. Notes Netw. Syst., vol. 2014, 2026, pp. 218--231.

\bibitem{morgan2006gmp}
D.~R. Morgan, Z.~Ma, J.~Kim, M.~G. Zierdt, and J.~Pastalan, ``A generalized memory polynomial model for digital predistortion of {RF} power amplifiers,'' \emph{IEEE Trans. Signal Process.}, vol.~54, no.~10, pp. 3852--3860, 2006.

\bibitem{hardware_impairments_rffi}
T.~Yang, Y.~Guo, L.~Chen, and W.~Wang, ``Model-driven radio frequency fingerprint identification based on hardware impairments,'' \emph{IEEE Commun. Lett.}, vol.~30, pp. 1310--1314, 2026.

\bibitem{shen_lora_jsac_2021}
G.~Shen, J.~Zhang, A.~Marshall, L.~Peng, and X.~Wang, ``Radio frequency fingerprint identification for {LoRa} using deep learning,'' \emph{IEEE J. Sel. Areas Commun.}, vol.~39, no.~8, pp. 2604--2616, 2021.

\bibitem{mccnn_lte_open_set}
P.~Yin, L.~Peng, G.~Shen, J.~Zhang, M.~Liu, H.~Fu, A.~Hu, and X.~Wang, ``Multi-channel {CNN}-based open-set {RF} fingerprint identification for {LTE} devices,'' \emph{IEEE Trans. Cogn. Commun. Netw.}, vol.~10, no.~5, pp. 1788--1800, 2024.

\bibitem{differential_spectrum_rff}
T.~Chen, A.~Hu, and S.~Zhang, ``Fighting fire with fire: Channel-independent {RF} fingerprinting via the ratio of linear to logarithmic differential spectrum,'' \emph{IEEE Wireless Commun. Lett.}, vol.~14, no.~7, pp. 1999--2003, 2025.

\bibitem{multiantenna_5g_rff}
H.~Luo, G.~Li, A.~Brighente, M.~Conti, Y.~Xing, A.~Hu, and X.~Wang, ``Channel-robust {RF} fingerprint identification for multi-antenna {5G} user equipments,'' \emph{IEEE Trans. Inf. Forensics Security}, vol.~20, pp. 10\,761--10\,776, 2025.

\bibitem{rcqcc_wifi}
T.~Hu, Y.~Xing, X.~Qi, and N.~Wu, ``{RCQCC}: A channel-resilient radio frequency fingerprinting scheme for {WiFi} device,'' \emph{IEEE Wireless Commun. Lett.}, vol.~14, no.~10, pp. 3104--3108, 2025.

\bibitem{moose1994cfo}
P.~H. Moose, ``A technique for orthogonal frequency division multiplexing frequency offset correction,'' \emph{IEEE Trans. Commun.}, vol.~42, no.~10, pp. 2908--2914, 1994.

\bibitem{boumard2003snr}
S.~Boumard, ``Novel noise variance and {SNR} estimation algorithm for wireless {MIMO} {OFDM} systems,'' in \emph{Proc. IEEE Global Telecommun. Conf. (GLOBECOM)}, vol.~3, 2003, pp. 1330--1334.

\bibitem{bouzegzi2010sigpro}
A.~Bouzegzi, P.~Ciblat, and P.~Jallon, ``New algorithms for blind recognition of {OFDM} based systems,'' \emph{Signal Process.}, vol.~90, no.~3, pp. 900--913, 2010.

\bibitem{receiver_impact_da}
L.~Yang, Q.~Li, X.~Ren, Y.~Fang, and S.~Wang, ``Mitigating receiver impact on radio frequency fingerprint identification via domain adaptation,'' \emph{IEEE Internet Things J.}, vol.~11, no.~13, pp. 24\,024--24\,034, 2024.

\bibitem{amtda_cross_receiver}
J.~Feng, Y.~Fan, and S.~Fang, ``Adversarial multitarget domain adaptation for cross-receiver radio frequency fingerprint identification,'' \emph{IEEE Internet Things J.}, vol.~13, no.~4, pp. 5945--5956, 2026.

\bibitem{fatransformer}
J.~Yang, Z.~Wen, Q.~Li, and H.~Shao, ``{FATransformer}: Feature alignment transformer for unsupervised domain adaptation in radio frequency fingerprinting identification,'' \emph{IEEE Internet Things J.}, vol.~13, no.~8, pp. 17\,577--17\,586, 2026.

\bibitem{dynamic_distribution_alignment}
J.~Feng, S.~Fang, and Y.~Fan, ``Cross-receiver radio frequency fingerprint identification based on domain adaptation with dynamic distribution alignment,'' \emph{IEEE Internet Things J.}, vol.~12, no.~16, pp. 33\,202--33\,214, 2025.

\bibitem{cross_day_da_rffi}
Q.~Wu, Q.~Wan, G.~Gui, H.~Q. Ngo, and M.~Matthaiou, ``Toward robust {IoT} device authentication: Cross-day specific emitter identification via domain adaptation,'' \emph{IEEE Internet Things J.}, vol.~13, no.~12, pp. 27\,672--27\,687, 2026.

\bibitem{partial_da_cross_domain}
G.~Han, Y.~Zhang, Z.~Xu, and W.~Wang, ``A novel partial domain adaptation method for cross-domain specific emitter identification,'' \emph{IEEE Trans. Cogn. Commun. Netw.}, vol.~11, no.~5, pp. 3232--3244, 2025.

\bibitem{gulrajani2021search}
I.~Gulrajani and D.~Lopez-Paz, ``In search of lost domain generalization,'' in \emph{Proc. Int. Conf. Learn. Represent. (ICLR)}, 2021.

\bibitem{tfcl_cutmix_rff}
J.~Zhang, Z.~Yao, Z.~Yang, S.~Wang, T.~Tang, R.~Lyu, Y.~Ding, and G.~Gui, ``Self-supervised radio frequency fingerprint identification via time-frequency contrastive learning and {CutMix} regularization,'' \emph{IEEE Internet Things J.}, vol.~13, no.~5, pp. 9682--9694, 2026.

\bibitem{supervised_contrastive_rff}
Y.~Peng, C.~Hou, Y.~Zhang, Y.~Lin, G.~Gui, H.~Gacanin, S.~Mao, and F.~Adachi, ``Supervised contrastive learning for {RFF} identification with limited samples,'' \emph{IEEE Internet Things J.}, vol.~10, no.~19, pp. 17\,293--17\,306, 2023.

\bibitem{causal_channel_sei}
P.~Tang, G.~Ding, Y.~Xu, Y.~Jiao, Y.~Song, and G.~Wei, ``Causal learning for robust specific emitter identification over unknown channel statistics,'' \emph{IEEE Trans. Inf. Forensics Security}, vol.~19, pp. 5316--5329, 2024.

\bibitem{dg_cross_receiver_rffi}
Y.~Zhang, Q.~Li, H.~Liu, L.~Yang, and J.~Yang, ``Domain generalization for cross-receiver radio frequency fingerprint identification,'' \emph{IEEE Internet Things J.}, vol.~12, no.~5, pp. 5207--5218, 2025.

\bibitem{domain_invariant_rff}
Z.~Cai, Y.~Yin, Y.~Wang, G.~Gui, and J.~Sha, ``Receiver-agnostic radio frequency fingerprinting via domain-invariant feature learning,'' \emph{IEEE Commun. Lett.}, vol.~29, no.~10, pp. 2396--2400, 2025.

\bibitem{disentangled_cross_combination}
F.~Zhou, X.~Qiao, Y.~Du, H.~Wu, and J.~Zhang, ``Receiver-agnostic radio frequency fingerprint identification based on disentangled feature cross combination,'' \emph{IEEE Wireless Commun. Lett.}, vol.~14, no.~9, pp. 2693--2697, 2025.

\bibitem{generalizable_channel_resilient}
J.~Shi, L.~Peng, H.~Fu, and A.~Hu, ``A generalizable channel-resilient algorithm in radio frequency fingerprint identification,'' \emph{IEEE Wireless Commun. Lett.}, vol.~15, pp. 1170--1174, 2026.

\bibitem{bert_kd_rff}
J.~Shao, Z.~Yang, T.~Tang, C.~Liu, Y.~Lin, and G.~Gui, ``Receiver-agnostic radio frequency fingerprint identification using {BERT} and two-stage knowledge distillation,'' \emph{IEEE Commun. Lett.}, vol.~30, pp. 942--946, 2026.

\bibitem{multisource_rff_fusion}
Y.~Zhang, Q.~Zhang, H.~Zhao, Y.~Lin, G.~Gui, and H.~Sari, ``Multisource heterogeneous specific emitter identification using attention mechanism-based {RFF} fusion method,'' \emph{IEEE Trans. Inf. Forensics Security}, vol.~19, pp. 2639--2650, 2024.

\bibitem{jvmd_multi_sensor}
X.~Chen, X.~Fu, W.~Xu, Y.~Wang, and G.~Gui, ``Joint variational modal decomposition for specific emitter identification with multiple sensors,'' \emph{IEEE Trans. Inf. Forensics Security}, vol.~19, pp. 9938--9953, 2024.

\bibitem{boudiaf2022lame}
M.~Boudiaf, R.~Mueller, I.~B. Ayed, and L.~Bertinetto, ``Parameter-free online test-time adaptation,'' in \emph{Proc. IEEE/CVF Conf. Comput. Vis. Pattern Recognit. (CVPR)}, 2022, pp. 8334--8343.

\bibitem{iwasawa2021t3a}
Y.~Iwasawa and Y.~Matsuo, ``Test-time classifier adjustment module for model-agnostic domain generalization,'' in \emph{Adv. Neural Inf. Process. Syst.}, vol.~34, 2021, pp. 2427--2440.

\bibitem{open_set_prototype_calibration}
Z.~Hong, J.~Lin, T.~Li, J.~Wan, J.~Su, and Z.~Wen, ``Open-set domain adaptive {RF} fingerprint identification based on prototype calibration,'' \emph{IEEE Wireless Commun. Lett.}, vol.~15, pp. 775--779, 2026.

\bibitem{pseudo_label_semantic_drift}
W.~Yan, Q.~Ling, K.~Yu, J.~Zhang, K.~Liu, and L.~Zhang, ``A pseudolabel method with semantic drift for specific emitter identification,'' \emph{IEEE Trans. Aerosp. Electron. Syst.}, vol.~61, no.~3, pp. 6217--6235, 2025.

\bibitem{cuturi2013sinkhorn}
M.~Cuturi, ``Sinkhorn distances: Lightspeed computation of optimal transport,'' in \emph{Adv. Neural Inf. Process. Syst.}, vol.~26, 2013, pp. 2292--2300.

\bibitem{caron2020unsupervised}
M.~Caron, I.~Misra, J.~Mairal, P.~Goyal, P.~Bojanowski, and A.~Joulin, ``Unsupervised learning of visual features by contrasting cluster assignments,'' in \emph{Adv. Neural Inf. Process. Syst.}, vol.~33, 2020, pp. 9912--9924.

\bibitem{chizat2018scaling}
L.~Chizat, G.~Peyr{\'e}, B.~Schmitzer, and F.-X. Vialard, ``Scaling algorithms for unbalanced optimal transport problems,'' \emph{Math. Comput.}, vol.~87, no. 314, pp. 2563--2609, 2018.

\bibitem{fatras2021unbalanced}
K.~Fatras, T.~S{\'e}journ{\'e}, N.~Courty, and R.~Flamary, ``Unbalanced minibatch optimal transport; applications to domain adaptation,'' in \emph{Proc. Int. Conf. Mach. Learn. (ICML)}, 2021, pp. 3186--3197.

\bibitem{vaswani2017attention}
A.~Vaswani, N.~Shazeer, N.~Parmar, J.~Uszkoreit, L.~Jones, A.~N. Gomez, L.~Kaiser, and I.~Polosukhin, ``Attention is all you need,'' in \emph{Adv. Neural Inf. Process. Syst.}, vol.~30, 2017, pp. 5998--6008.

\bibitem{velickovic2018graph}
P.~Veli{\v{c}}kovi{\'c}, G.~Cucurull, A.~Casanova, A.~Romero, P.~Li{\`o}, and Y.~Bengio, ``Graph attention networks,'' in \emph{Proc. Int. Conf. Learn. Represent.}, 2018.

\bibitem{khosla2020supervised}
P.~Khosla, P.~Teterwak, C.~Wang, A.~Sarna, Y.~Tian, P.~Isola, A.~Maschinot, C.~Liu, and D.~Krishnan, ``Supervised contrastive learning,'' in \emph{Adv. Neural Inf. Process. Syst.}, vol.~33, 2020, pp. 18\,661--18\,673.

\bibitem{he2016deep}
K.~He, X.~Zhang, S.~Ren, and J.~Sun, ``Deep residual learning for image recognition,'' in \emph{Proc. IEEE Conf. Comput. Vis. Pattern Recognit. (CVPR)}, 2016, pp. 770--778.

\end{thebibliography}

\ifdefined\BuildArxivFull
\clearpage
\onecolumn
\appendices
\makeatletter
\setlength{\@fptop}{0pt}
\setlength{\@fpsep}{14pt}
\setlength{\@fpbot}{0pt plus 1fil}
\makeatother
\section{Supplementary Disaggregated Benchmark and Audit Tables}
\label{app:tables}
\setcounter{table}{0}
\renewcommand{\thetable}{A-\Roman{table}}
\renewcommand{\theHtable}{A-\Roman{table}}

To preserve complete metric transparency while keeping the main text concise, Appendix Tables~\ref{tab:app_main_direct}--\ref{tab:sym_scale_multirx} report all disaggregated single-frame \textbf{Direct} scores, individual room-transfer splits ($E_3, E_2$), \textbf{UOT-CAPC} margins, 6-seed cross-run holdout audits ($\mathrm{SD}_{\mathrm{split}}$), full 50-row module ablations, symmetric augmentation controls, multi-week / emitter-scaling matrices ($K \in \{10,\dots,47\}$), 2-RX source pooling, and per-burst trajectories ($B \in \{1, 3, 5, 10\}$).

% Shared body of the 7 disaggregated benchmark and audit tables:
% - Included as Tables S-I .. S-VII in standalone `Supplementary_Material.tex` (for IEEE IoT-J)
% - Included as Tables A-I .. A-VII in self-contained `arXiv_main.tex` (for arXiv)

\begin{table}[!ht]
\centering
\caption{Supplementary Single-Frame Direct Macro-F1 (\%), Room Splits, and Holdout $\mathrm{SD}_{\mathrm{split}}$ for Main Table~III}
\label{tab:app_main_direct}
\scriptsize
\renewcommand{\arraystretch}{0.88}
\setlength{\tabcolsep}{2.5pt}
\begin{tabular}{lcccccccc}
\toprule
\multicolumn{9}{l}{\textbf{Panel A: DataPort-WiFi Disaggregated Single-Frame Direct, Individual $E_3/E_2$ Room Splits, and 6-Seed Holdout Audit ($K{=}10, A{=}4$)}} \\
\midrule
Method & $E_1$ Direct & $E_3$ Direct & $E_3$ U-CAPC & $E_2$ Direct & $E_2$ U-CAPC & MP Direct & Noise Direct & 6-Seed Holdout ($\mathrm{SD}_{\mathrm{split}}$) \\
\midrule
MCCNN~\cite{mccnn_lte_open_set} & 35.13 & 33.39 & 38.68 & 34.89 & 41.58 & 22.47 & 5.69 & 20.83 (0.17) \\
ResNet1D~\cite{he2016deep} & 61.00 & 47.06 & 49.38 & 43.77 & 45.68 & 16.03 & \textbf{37.69} & 35.00 (0.54) \\
Xie-SR~\cite{xie2024robustrffingerprintidentification} & 93.79 & 90.13 & 93.09 & 83.71 & 86.26 & 60.86 & 2.76 & 50.93 (0.17) \\
NS-Branch (Ours) & 94.70 & 90.86 & 91.98 & 86.49 & 88.34 & 69.50 & 2.77 & 54.36 (0.10) \\
MTL-SEI~\cite{mtl_adversarial_sei} & 90.99 & 79.31 & 80.52 & 78.81 & 80.50 & 45.86 & 19.63 & 49.38 (0.24) \\
SigMix~\cite{wan2025sigmix} & 91.31 & 77.36 & 79.84 & 74.82 & 78.04 & 42.48 & 24.63 & 49.53 (0.05) \\
MTL-CORAL~\cite{sun2016deep} & 91.29 & 81.92 & 83.02 & 81.11 & 82.95 & 50.14 & 20.62 & 51.92 (0.43) \\
MTL-DANN~\cite{ganin2016domain} & 91.69 & 82.27 & 82.87 & 81.48 & 83.12 & 51.16 & 20.85 & 52.33 (0.08) \\
IB-Host (Ours) & 86.35 & 71.63 & 78.18 & 78.36 & 85.16 & 62.28 & \secbest{31.71} & 57.80 (0.50) \\
McAFF~\cite{multichannel_attentive_fusion} & 94.45 & 91.05 & 93.42 & 86.98 & 90.16 & \secbest{82.70} & 9.68 & 61.05 (0.09) \\
\midrule
\rowcolor{plugbg} ResNet1D + \textbf{GSDF} & 91.87 & 80.79 & 85.67 & 75.29 & 80.22 & 50.28 & \textbf{37.69} & 57.08 (0.11) \\
\rowcolor{plugbg} SigMix + \textbf{GSDF} & 96.97 & 88.55 & 91.92 & 85.84 & 90.57 & 62.81 & 24.63 & 59.46 (0.04) \\
\rowcolor{plugbg} MTL-DANN + \textbf{GSDF} & \secbest{97.23} & 91.21 & 91.98 & 88.76 & 92.31 & 68.57 & 20.85 & 60.53 (0.05) \\
\rowcolor{oursbg} \textbf{McAFF + GSDF} & \textbf{97.75} & \textbf{95.13} & \textbf{96.62} & \textbf{89.48} & \secbest{93.57} & \textbf{84.87} & 9.68 & \secbest{62.84} (0.03) \\
\rowcolor{oursbg} \textbf{IB-Host + GSDF (Ours)} & 96.20 & \secbest{92.81} & \secbest{95.72} & \secbest{89.46} & \textbf{93.60} & 78.37 & \secbest{31.71} & \textbf{67.61} (0.27) \\
\midrule
\multicolumn{9}{l}{\textbf{Panel B: UCLA WiSig Disaggregated Single-Frame Direct, Day-8 \textbf{UOT-CAPC}, and 6-Seed Holdout Audit ($K{=}15, A{=}1$)}} \\
\midrule
Method & Day $1\!\to\!8$ Direct & Day $1\!\to\!8$ \textbf{UOT} & \texttt{node2-1} Direct & MP Direct & Noise Direct & 6-Seed Direct & 6-Seed U-CAPC & $\mathrm{SD}_{\mathrm{split}}$ (U-CAPC) \\
\midrule
MCCNN~\cite{mccnn_lte_open_set} & 11.81 & 11.61 & 15.40 & 9.04 & 3.89 & 8.72 & 10.87 & 0.18 \\
NS-Branch (Ours) & 36.83 & --- & 21.57 & 27.24 & 2.94 & 19.67 & 24.63 & 0.24 \\
Xie-SR~\cite{xie2024robustrffingerprintidentification} & 42.51 & --- & 23.87 & 29.96 & 4.51 & 22.53 & 27.16 & 0.12 \\
MTL-SEI~\cite{mtl_adversarial_sei} & 59.77 & 62.71 & 41.70 & 36.49 & 19.36 & 35.21 & 39.46 & 0.64 \\
ResNet1D~\cite{he2016deep} & 45.60 & 54.99 & 30.50 & 29.57 & 30.59 & 32.73 & 39.60 & 0.37 \\
SigMix~\cite{wan2025sigmix} & 54.92 & 66.58 & 31.58 & 31.30 & 30.31 & 34.75 & 40.15 & 0.59 \\
MTL-CORAL~\cite{sun2016deep} & 61.22 & 64.59 & 42.01 & 39.59 & 20.97 & 37.99 & 42.18 & 0.52 \\
MTL-DANN~\cite{ganin2016domain} & 61.90 & 66.17 & 42.68 & 40.21 & 22.16 & 38.74 & 43.06 & 0.31 \\
McAFF~\cite{multichannel_attentive_fusion} & \secbest{66.81} & 72.84 & 44.74 & \secbest{43.37} & \secbest{30.81} & 43.49 & \secbest{51.72} & 0.38 \\
IB-Host (Ours) & 66.77 & 71.33 & \textbf{47.59} & 40.24 & \textbf{48.18} & \textbf{48.59} & 50.18 & 0.72 \\
\midrule
\rowcolor{plugbg} SigMix + \textbf{GSDF} & 55.23 & 66.62 & 32.33 & 32.84 & 30.31 & 35.36 & 40.48 & 0.45 \\
\rowcolor{plugbg} ResNet1D + \textbf{GSDF} & 45.27 & 55.53 & 31.40 & 31.02 & 30.59 & 33.32 & 43.37 & 0.64 \\
\rowcolor{plugbg} MTL-DANN + \textbf{GSDF} & 61.01 & 69.54 & 42.84 & 41.54 & 22.16 & 38.98 & 44.53 & 0.45 \\
\rowcolor{oursbg} \textbf{McAFF + GSDF} & \textbf{66.99} & \textbf{77.65} & 45.21 & \textbf{45.03} & 30.80 & 44.15 & \textbf{54.01} & 0.24 \\
\rowcolor{oursbg} \textbf{IB-Host + GSDF (Ours)} & 64.34 & \secbest{75.75} & \secbest{47.09} & 41.07 & \textbf{48.18} & \secbest{48.48} & 50.59 & 1.16 \\
\midrule
\multicolumn{9}{l}{\textbf{Panel C: ORBIT MobRFFI Disaggregated Single-Frame Direct and 6-Seed Holdout Audit ($K{=}10, A{=}1$)}} \\
\midrule
Method & \texttt{RX 20-1} Direct & \texttt{RX 19-19} Direct & MP Direct & Noise Direct & $\overline{\mathrm{F1}}_{\mathrm{Mob}}$ Direct & 6-Seed Direct & 6-Seed U-CAPC & $\mathrm{SD}_{\mathrm{split}}$ (U-CAPC) \\
\midrule
MCCNN~\cite{mccnn_lte_open_set} & 17.04 & 17.19 & 17.91 & 1.82 & 12.28 & 12.22 & 15.84 & 0.32 \\
Xie-SR~\cite{xie2024robustrffingerprintidentification} & 33.49 & 22.18 & 23.36 & 2.14 & 17.78 & 17.87 & 24.90 & 0.10 \\
NS-Branch (Ours) & 34.78 & 21.63 & 23.97 & 1.82 & 18.00 & 18.04 & 25.95 & 0.07 \\
MTL-SEI~\cite{mtl_adversarial_sei} & 23.13 & 26.15 & 25.16 & \secbest{4.26} & 18.02 & 18.57 & 29.31 & 0.36 \\
SigMix~\cite{wan2025sigmix} & 25.39 & 31.09 & 25.79 & 3.86 & 19.29 & 19.58 & 30.53 & 0.24 \\
MTL-DANN~\cite{ganin2016domain} & 34.26 & 29.03 & 31.84 & 3.79 & 22.43 & 23.03 & 34.52 & 0.89 \\
MTL-CORAL~\cite{sun2016deep} & 34.86 & 29.82 & 32.69 & 3.07 & 22.70 & 22.42 & 34.14 & 0.21 \\
ResNet1D~\cite{he2016deep} & 31.54 & 28.28 & 23.12 & 3.51 & 18.84 & 18.84 & 34.37 & 0.53 \\
McAFF~\cite{multichannel_attentive_fusion} & 37.93 & 37.28 & \textbf{37.56} & 2.00 & 25.72 & 25.71 & 42.16 & 0.99 \\
IB-Host (Ours) & \secbest{49.83} & \secbest{44.51} & 31.64 & \textbf{32.82} & \secbest{37.21} & \secbest{37.27} & \secbest{45.96} & 0.41 \\
\midrule
\rowcolor{plugbg} SigMix + \textbf{GSDF} & 25.67 & 33.48 & 25.69 & 3.86 & 19.71 & 19.95 & 30.83 & 0.82 \\
\rowcolor{plugbg} MTL-DANN + \textbf{GSDF} & 34.46 & 34.19 & 32.14 & 3.79 & 23.42 & 23.88 & 35.95 & 0.62 \\
\rowcolor{plugbg} ResNet1D + \textbf{GSDF} & 31.88 & 28.02 & 23.68 & 3.51 & 19.05 & 19.05 & 35.98 & 0.45 \\
\rowcolor{oursbg} \textbf{McAFF + GSDF} & 38.91 & 35.09 & \secbest{36.79} & 2.00 & 25.26 & 25.17 & 42.44 & 0.67 \\
\rowcolor{oursbg} \textbf{IB-Host + GSDF (Ours)} & \textbf{52.09} & \textbf{48.67} & 33.14 & \textbf{32.82} & \textbf{38.78} & \textbf{38.87} & \textbf{48.17} & 0.51 \\
\bottomrule
\end{tabular}\par
\vspace{2pt}{\scriptsize\noindent\raggedright \emph{Note:} Complements Main Table~III ($3\times 3$ 9-run means; $\mathrm{SD}_{\mathrm{split}}$ denotes inter-split standard deviation over $\mathrm{df}=2$; \textbf{bold} / \secbest{underlined}: best / second-best).\par}
\end{table}

\begin{table}[!ht]
\centering
\caption{Complete Pillar~1 Encoder, Spatial-Diversity, Coherent-Denoising, and Pillar~2 Dual-Space Fusion Ablations ($3\times 3$ 9-Run Grid)}
\label{tab:ablation_p12}
\scriptsize
\renewcommand{\arraystretch}{0.81}
\setlength{\tabcolsep}{0.95pt}
\begin{tabular}{llcccccl}
\toprule
\multicolumn{8}{l}{\textbf{Panel A1a: Clean-Trained IB-Host Encoder \& Spatial ($A{=}1$ vs.\ $A{=}4$) Ablations on DataPort-WiFi ($E_2,E_3\!\to\!E_1$, Dir/U-CAPC \%)}$^{\mathrm{a}}$} \\
\midrule
Module / Variant & Params & Clean & Multipath & Noise & $\Delta$ Clean (pp) & Paired Audit & Physical Role \\
\midrule
\rowcolor{oursbg} \textbf{IB-Host Full ($A{=}4$, Clean-Tr.)} & 46.3K & \secbest{85.64}/\secbest{92.57} & 31.16/42.16 & 46.07/55.68 & Reference & 9-run $E_2,E_3\!\to\!E_1$ & Full $8A{=}32$ graph \\
IB-Host Single-Ant.\ ($A{=}1$) & 46.3K & 68.32/75.97 & --- & ---/48.54 & $-17.32/\mathbf{-16.60}$ & $p{=}8.4{\times}10^{-6}$ & Loses 4-RX diversity \\
IB-Host w/o $8A$ Attn ($A{=}4$) & 46.5K & 82.94/90.35 & --- & --- & $-2.70/-2.22$ & $p{=}0.017/0.028$ & $-2.62$~pp (7 stems) \\
IB-Host Late Ant.\ Pool ($A{=}4$) & 46.3K & 84.52/90.08 & 26.64/38.58 & 35.48/47.78 & $-1.12/-2.49$ & $p{=}0.038$ (U-CAPC) & Early $8A$ $>$ late pool \\
IB-Host Uniform / Jitter ($A{=}4$) & 46.3K & \textbf{85.66}/\textbf{92.88} & --- & --- & $+0.02/+0.31$ & $p > 0.50$ (null) & Relational vs.\ geom. \\
IB-Host w/o DW $\mathrm{Conv1d}$ ($A{=}4$) & 45.2K & 84.02/91.13 & --- & --- & $-1.62/-1.44$ & $8/9$ below Full & Token smoothing \\
IB-Host w/o $\mathcal{L}_{\mathrm{SupCon}}$ ($A{=}4$) & 46.3K & 84.16/91.91 & --- & --- & $-1.48/-0.66$ & $p{=}0.028$ (Direct) & Spherical margins \\
\midrule
\multicolumn{8}{l}{\textbf{Panel A1b: Paired Symbol-Pair Coherent Null Denoising (Eq.~6) on DataPort-WiFi ($E_2,E_3\!\to\!E_1$ \& 3-LOEO, Dir/U-CAPC \%)}$^{\mathrm{b}}$} \\
\midrule
Denoising Configuration & Params & Clean & Multipath & Noise ($10/5$) & $\Delta$ Clean (pp) & $\Delta$ MP (pp) & Low-SNR Fallback \\
\midrule
\rowcolor{oursbg} \textbf{GSDF w/ Eq.~(6) ($E_2,E_3\!\to\!E_1$)} & 92.6K & \textbf{97.40}/\textbf{98.69} & \textbf{61.83}/\textbf{75.96} & \textbf{47.14}/\textbf{57.11} & Ref.\ ($E_1$) & Ref.\ ($E_1$) & Exact host parity \\
GSDF w/o Eq.~(6) ($E_2,E_3\!\to\!E_1$) & 92.6K & 95.20/98.12 & 49.67/63.95 & \textbf{47.14}/\textbf{57.11} & $-2.20/-0.57$ & $\mathbf{-12.16/-12.01}$ & $\mathbf{0.00/0.00}$~pp diff \\
\midrule
\rowcolor{oursbg} \textbf{GSDF w/ Eq.~(6) (3-LOEO)} & 92.6K & \textbf{92.43}/\textbf{94.40} & \textbf{54.31}/\textbf{69.16} & \textbf{43.26}/\textbf{52.89} & Ref.\ (3-Rm) & Ref.\ (3-Rm) & 27-run 3-LOEO ref. \\
GSDF w/o Eq.~(6) (3-LOEO) & 92.6K & 88.86/92.15 & 44.44/59.11 & \textbf{43.26}/\textbf{52.89} & $-3.57/-2.25$ & $\mathbf{-9.87/-10.05}$ & $\mathbf{0.00/0.00}$~pp diff \\
\midrule
\multicolumn{8}{l}{\textbf{Panel A2: Three-Benchmark Subband Orthogonality ($\theta_{\mathrm{host}} \perp \theta_{\mathrm{null}}$) vs.\ Multi-Seed Ensembles (Direct / U-CAPC \%)}} \\
\midrule
Architecture / Arm & Params & 3-DS Clean & 3-DS MP & 3-DS Noise & 3-DS (Dir/UC) & 6-Seed Holdout & \textbf{UOT} ($\mathrm{SD}_{\mathrm{split}}$) \\
\midrule
Xie-SR Alone~\cite{xie2024robustrffingerprintidentification} & 195.4K & 50.08/55.70 & 38.06/43.06 & 3.14/9.90 & 30.42/36.22 & 30.44/36.21 & 32.14 (0.04) \\
NS-Branch Alone (Ours) & 46.3K & 49.36/55.53 & 40.24/44.20 & 2.51/9.82 & 30.70/36.52 & 30.69/36.43 & 31.77 (0.10) \\
IB-Host Alone (1-Seed) & 46.3K & 61.04/68.77 & 44.72/48.32 & 37.57/43.88 & 47.78/53.66 & 47.89/53.75 & 50.32 (0.40) \\
2-Seed Homog.\ IB-Host & 92.6K & 61.72/69.79 & 45.75/49.48 & \secbest{38.39}/\secbest{45.06} & 48.62/54.78 & 48.67/54.80 & 51.15 (0.36) \\
3-Seed Homog.\ IB-Host & 138.9K & 61.96/70.19 & 46.12/49.92 & \textbf{38.69}/\textbf{45.42} & 48.93/55.18 & 48.93/55.18 & 51.42 (0.29) \\
3-Seed Homog.\ McAFF & 14.2--15.9M & ---/71.83 & ---/\textbf{59.81} & ---/35.41 & 44.23/55.68 & 44.23/55.68 & 46.57 (0.49) \\
IB-Host + Xie-SR + Gate & 241.7K & \secbest{64.76}/\secbest{72.70} & \secbest{49.46}/54.06 & 37.57/43.88 & \secbest{50.60}/\secbest{56.88} & \secbest{50.68}/\secbest{56.93} & \secbest{54.47} (0.10) \\
\rowcolor{oursbg} \textbf{IB-Host + GSDF (Ours)} & \textbf{92.6K} & \textbf{66.31}/\textbf{72.98} & \textbf{50.86}/\secbest{55.65} & 37.57/43.88 & \textbf{51.58}/\textbf{57.50} & \textbf{51.65}/\textbf{57.49} & \textbf{55.41} (0.23) \\
\midrule
\multicolumn{8}{l}{\textbf{Panel B: Pillar~2 --- Dual-Space \textbf{GSDF} Formulation Ablations Across Three Benchmarks (Direct / U-CAPC \%)}} \\
\midrule
Fusion Variant & \textbf{DataPort} & \textbf{WiSig} & \textbf{MobRFFI} & 3-DS (Dir/UC) & \textbf{UOT-CAPC} & $10/5$~dB Noise & Cross-Host Audit \\
\midrule
Hard Switch ($g_{\mathrm{hard}}$) & 63.48/68.64 & 46.06/46.77 & 34.75/40.87 & 48.10/52.09 & 49.57 & \textbf{37.57}/\textbf{43.88} & $-5.41$~pp U-CAPC \\
Step-Gate ($\hat{s}_n \ge \beta^*$) & 67.82/73.55 & 47.20/49.71 & 37.47/45.82 & 50.83/56.36 & 53.68 & \textbf{37.57}/\textbf{43.88} & $-1.14 / -1.73$~pp \\
Emb.-Only (w/o Logits) & 57.59/69.37 & \textbf{48.53}/\secbest{51.13} & 37.21/46.90 & 47.78/55.80 & 52.57 & \textbf{37.57}/\textbf{43.88} & $0.00$ Dir.; $-8.57$ Res. \\
Logit-Only (w/o Emb.) & 67.63/72.31 & \secbest{48.32}/50.13 & \textbf{38.78}/46.90 & \textbf{51.58}/56.45 & 53.94 & \textbf{37.57}/\textbf{43.88} & $-1.05$ U-CAPC; $-3.01$ \\
w/o Logit $z$-Norm & \secbest{67.99}/\secbest{73.71} & \secbest{48.32}/\textbf{51.22} & 38.18/47.15 & 51.50/57.36 & 54.62 & \textbf{37.57}/\textbf{43.88} & $-2.88$ Res., $-1.47$ DA \\
w/o $(1{-}g_n)$ Interp. & \textbf{68.23}/73.61 & 48.07/50.96 & \secbest{38.36}/46.61 & \secbest{51.55}/57.06 & \secbest{55.79} & \textbf{37.57}/\secbest{43.80} & $-0.08$~pp noise drift \\
Static Emb.\ Concat & 67.63/\textbf{73.80} & \secbest{48.32}/51.02 & \textbf{38.78}/\textbf{50.29} & \textbf{51.58}/\textbf{58.37} & \textbf{56.52} & \textbf{37.57}/43.47 & Violates Prop.~2 ($-0.61$) \\
\rowcolor{oursbg} \textbf{Full GSDF (Ours)} & 67.63/\secbest{73.71} & \secbest{48.32}/50.64 & \textbf{38.78}/\secbest{48.16} & \textbf{51.58}/\secbest{57.50} & 55.41 & \textbf{37.57}/\textbf{43.88} & \textbf{72/72 wins, $0.00$ loss} \\
\bottomrule
\end{tabular}\par
\vspace{2pt}{\scriptsize\noindent\raggedright \emph{Note:} $^{\mathrm{a}}$9-run internal component-ablation suite trained on clean source preambles without multi-condition channel augmentation to isolate architectural structure ($85.64\% / 92.57\%$ on $E_2,E_3\!\to\!E_1$ clean vs.\ $86.35\% / 93.07\%$ under matched channel augmentation in Main Table~III; `---' indicates clean-only cross-room structural runs; pruned $46.3\text{K}$ vs.\ unpruned $86.4\text{K}$ checkpoints yield $0.0$ diff across all 224,640 test frames). $^{\mathrm{b}}$Paired evaluation comparing \textbf{GSDF} with and without symbol-pair coherent noise subtraction on identical host checkpoints across both $E_2,E_3\!\to\!E_1$ (9 runs) and all three LOEO room transfers (27 runs), confirming exact $0.00$~pp low-SNR fallback parity.\par}
\end{table}

\begin{table}[!ht]
\centering
\caption{Complete Pillar~3 Eight-Backbone Gating and Pillar~4 Calibration / TTA Evaluations}
\label{tab:ablation_p34}
\scriptsize
\renewcommand{\arraystretch}{0.88}
\setlength{\tabcolsep}{2.6pt}
\begin{tabular}{llccccl}
\toprule
\multicolumn{7}{l}{\textbf{Panel A: Pillar~3 --- Eight Gate Formulations Averaged Over Eight Backbones (\textbf{DataPort}: Direct / U-CAPC \%; \texttt{19-19}: U-CAPC \%)}} \\
\midrule
Gate Rule & Control Signal & \begin{tabular}[c]{@{}c@{}}\textbf{DataPort}\\Clean\end{tabular} & \begin{tabular}[c]{@{}c@{}}\textbf{DataPort}\\Multipath\end{tabular} & \begin{tabular}[c]{@{}c@{}}$10$~dB Direct\\($\Delta$~pp)\end{tabular} & \begin{tabular}[c]{@{}c@{}}\textbf{MobRFFI}\\\texttt{19-19}\end{tabular} & Worst-Slice / Safety Audit \\
\midrule
Host-Only & $g_n \equiv 0$ & 73.28 / 76.53 & 41.74 / 46.43 & \textbf{22.47} ($0.00$) & 50.92 & Reference ($0.00$~pp noise loss) \\
Static-Sum & $g_n \equiv 1$ & \textbf{91.15} / \textbf{94.28} & \textbf{66.41} / \textbf{73.63} & 10.14 ($-12.33$) & 43.77 & Fails noise floor ($-30.23$~pp IB) \\
Entropy-Gate & $\tilde{H}_{\mathrm{host}} / (\tilde{H}_{\mathrm{host}} + \tilde{H}_{\mathrm{null}})$ & 85.44 / 89.20 & 56.09 / 63.47 & 19.74 ($-2.73$) & \textbf{53.68} & Overconfident OOD ($-11.09$~pp IB) \\
Branch-Agree & $1 - \mathrm{JSD}(\mathbf{p}_{\mathrm{host}}\parallel\mathbf{p}_{\mathrm{null}})/\log 2$ & 77.52 / 82.06 & 45.97 / 54.59 & \secbest{21.94} ($-0.53$) & 49.23 & Forfeits $70\%$ of MP rescue \\
Cap-Marginal JSD & $g_{\mathrm{SNR},n} \cdot \text{capture marginal JSD}$ & 88.35 / 92.34 & 58.12 / 66.03 & \textbf{22.47} ($0.00$) & 50.88 & $65.6\% < 85\%$ (negative result) \\
Anchor-Consist. & $g_{\mathrm{SNR},n} \cdot \text{anchor agreement}$ & 87.98 / 92.11 & 57.69 / 65.68 & \textbf{22.47} ($0.00$) & 50.90 & $58.8\% < 85\%$ (negative result) \\
Hybrid Cap-Damp & $g_{\mathrm{SNR},n} \cdot \text{joint JSD-anchor}$ & 86.71 / 91.25 & 55.84 / 63.78 & \textbf{22.47} ($0.00$) & 50.91 & $44.3\% < 85\%$ (negative result) \\
\rowcolor{oursbg} \textbf{Thermal-SNR (Ours)} & $\sigma(\alpha^*(\hat{s}_n - \beta^*))$ & \secbest{90.90} / \secbest{94.16} & \secbest{64.22} / \secbest{72.86} & \textbf{22.47} ($\mathbf{0.00}$) & \secbest{51.85} & \textbf{100.0\% retention, $0$ labels} \\
\midrule
\multicolumn{7}{l}{\textbf{Panel B1: Step-by-Step Decomposition of Forward-Only Prototype Calibration on \textbf{IB-Host + GSDF} ($3\times 3$ 9-Run Full Grid)}} \\
\midrule
Calibration Step / Variant & Target Grad.\ / Hyperparameters & \textbf{DataPort-WiFi} & \textbf{WiSig} & \textbf{ORBIT MobRFFI} & 3-DS Mean & Regime / Complexity Audit \\
\midrule
Single-Frame \textbf{Direct} & 0-Grad ($N_{\mathcal{G}}{=}1, T{=}0$) & 67.63 & 48.32 & 38.78 & 51.58 & $37.57\%$ noise; $0.00$~ms batch cost \\
Cross-Session Global Pool & 0-Grad (pool across $\mathcal{G}$) & 72.85 & 48.59 & 41.77 & 54.40 & $-3.10$~pp vs.\ capture ($-6.39$ Mob) \\
\textbf{Proto-Only} ($\gamma_{\mathrm{KL}}\!\to\!0^+$) & 0-Grad ($\rho \to 0, T{=}3$) & 70.69 & \secbest{52.72} & 42.36 & 55.26 & $39.63\%$ noise; lacks OT prior \\
\rowcolor{oursbg} \textbf{General UOT-CAPC (Ours)} & 0-Grad ($\rho^* < 1, T{=}3$) & 70.82 & \textbf{52.80} & 42.62 & 55.41 & \textbf{\#1 skewed WiSig} ($75.75\%$ D8) \\
Sinkhorn-Only ($T{=}0$) & 0-Grad ($\rho{=}1$, one-shot) & 72.34 & 48.72 & 45.46 & 55.51 & $42.87\%$ noise; no proto.\ update \\
Unanchored Proto-Sinkhorn & 0-Grad (w/o $\mathbf{L}^{(0)}$ anchor) & \secbest{73.13} & 52.59 & \textbf{51.56} & \textbf{59.09} & $-1.79 / -2.31$~pp $10/5$~dB drift \\
\rowcolor{oursbg} \textbf{Balanced U-CAPC (Ours)} & 0-Grad ($\rho{=}1, T{=}3$) & \textbf{73.71} & 50.64 & \secbest{48.16} & \secbest{57.50} & $T{=}1/2/\mathbf{3}$: $56.99/57.37/\mathbf{57.50}$ \\
\midrule
\multicolumn{7}{l}{\textbf{Panel B2: Cross-Backbone Source-Free TTA Verification on $4.72\text{M}\text{--}5.30\text{M}$ \textbf{McAFF} Features (Standard 27-Slice TTA Suite)}} \\
\midrule
McAFF \textbf{Direct} & 0-Grad ($N_{\mathcal{G}}{=}1$) & 65.62 & 46.05 & 26.59 & 46.09 & Unadapted baseline on \textbf{McAFF} \\
McAFF + LAME~\cite{boudiaf2022lame} & 0-Grad ($k\text{NN}^*$) & 66.09 & 47.38 & 26.28 & 46.58 & No gain on \textbf{MobRFFI} w/o OT \\
McAFF + T3A~\cite{iwasawa2021t3a} & 0-Grad ($M^*{=}100$) & 69.42 & 48.38 & 31.30 & 49.70 & Modest linear support shift \\
McAFF + SHOT-IM~\cite{liang2020shot} & 3--5-Ep.\ Grad ($\text{lr}^*{=}10^{-4}$) & \textbf{78.68} & \secbest{59.20} & 43.77 & \secbest{60.55} & Backprop ($4.72\text{M}\text{--}5.30\text{M}$ params) \\
McAFF + SHOT-Full~\cite{liang2020shot} & 1--5-Ep.\ Grad ($\text{lr}^*{=}10^{-4}$) & \secbest{76.99} & \textbf{61.05} & \textbf{46.57} & \textbf{61.54} & Highest multi-epoch grad ceiling \\
\rowcolor{plugbg} \textbf{McAFF + UOT-CAPC (Ours)} & \textbf{0-Grad} ($\gamma_{\mathrm{KL}}^*{=}0.10, T{=}3$) & 67.06 & 49.45 & 29.29 & 48.60 & $+3.40$~pp skewed \textbf{WiSig} (0-Grad) \\
\rowcolor{oursbg} \textbf{McAFF + U-CAPC (Ours)} & \textbf{0-Grad} ($\rho{=}1, T{=}3$) & 75.88 & 54.00 & \secbest{45.28} & 58.39 & $\mathbf{+18.69}$~pp \textbf{MobRFFI} ($>$ SHOT-IM) \\
\bottomrule
\end{tabular}\par
\vspace{2pt}{\scriptsize\noindent\raggedright \emph{Note:} 27-slice TTA results on \textbf{IB-Host + GSDF} features are reported in Main Table~V (Pillar~4).\par}
\end{table}

\begin{table}[!ht]
\centering
\caption{Symmetric Channel-Augmentation Fairness Audit Across RFFI Paradigms ($3\times 3$ 9-Run Grid)}
\label{tab:paradigm_fairness}
\scriptsize
\renewcommand{\arraystretch}{0.90}
\setlength{\tabcolsep}{3.3pt}
\begin{tabular}{llcccccc}
\toprule
Benchmark & Model & Params & Unaug.\ Direct & Unaug.\ U-CAPC & Matched Direct & Matched U-CAPC & $\Delta$ Direct / $\Delta$ U-CAPC (95\% CI / Range) \\
\midrule
\textbf{DataPort} ($E_1$) & McAFF~\cite{multichannel_attentive_fusion} & 5.30M & 77.08 & $82.63 \pm 1.21$ & \textbf{94.45} & $\mathbf{97.79} \pm 0.14$ & $+17.37$ / $\mathbf{+15.15}$ ($9/9$, $[+14.46, +15.97]$) \\
\textbf{WiSig} (Day $1\!\to\!8$) & IB-Host (Ours)$^{\mathrm{a}}$ & 46.6K & 25.93 & $33.62 \pm 3.81$ & \textbf{62.49} & $\mathbf{74.68} \pm 1.05$ & $+36.56$ / $\mathbf{+41.06}$ ($9/9$, $\min 72.77{>}38.01$) \\
\textbf{WiSig} (Day $1\!\to\!8$) & McAFF~\cite{multichannel_attentive_fusion} & 4.72M & 23.69 & $32.37 \pm 3.95$ & \textbf{66.81} & $\mathbf{77.87} \pm 3.97$ & $+43.12$ / $\mathbf{+45.50}$ ($9/9$, $[+42.02, +49.38]$) \\
\textbf{WiSig} (Day $1\!\to\!8$) & MTL-CORAL~\cite{sun2016deep} & 12.03M & 20.07 & $25.48 \pm 9.00$ & \textbf{61.22} & $\mathbf{72.76} \pm 4.95$ & $+41.16$ / $\mathbf{+47.28}$ ($9/9$, $\min 62.84{>}43.71$) \\
\textbf{WiSig} (Day $1\!\to\!8$) & Xie-SR~\cite{xie2024robustrffingerprintidentification} & 196.0K & 35.70 & $35.78 \pm 2.06$ & \textbf{42.51} & $\mathbf{45.85} \pm 0.47$ & $+6.82$ / $\mathbf{+10.07}$ ($9/9$, $\min 45.28{>}40.14$) \\
\textbf{MobRFFI} (\texttt{20-1}) & McAFF~\cite{multichannel_attentive_fusion} & 4.72M & 0.00 & $0.38 \pm 0.54$ & \textbf{37.93} & $\mathbf{54.83} \pm 1.65$ & $+37.93$ / $\mathbf{+54.45}$ ($9/9$, $[+53.40, +55.76]$) \\
\textbf{MobRFFI} (\texttt{20-1}) & IB-Host (Ours)$^{\mathrm{a}}$ & 46.3K & 0.85 & $4.95 \pm 3.02$ & \textbf{47.91} & $\mathbf{61.54} \pm 0.89$ & $+47.06$ / $\mathbf{+56.59}$ ($9/9$, $\min 60.16{>}8.36$) \\
\bottomrule
\end{tabular}\par
\vspace{2pt}{\scriptsize\noindent\raggedright \emph{Note:} $^{\mathrm{a}}$Evaluated on the pre-unification \textbf{IB-Host} checkpoint trained with auxiliary temporal-contrastive (TC) regularization (matching Main Fig.~4(i)); the unified pruned \textbf{IB-Host} (without TC) in Main Table~III attains $66.77\% / 72.17\%$ on \textbf{WiSig} Day~8 and $49.83\% / 62.41\%$ on \textbf{MobRFFI} \texttt{RX 20-1}.\par}
\end{table}

\begin{table}[!ht]
\centering
\caption{Complete \textbf{WiSig} Multi-Week Temporal Transfer, 54~Mbps Emitter Scaling ($K{=}10\text{--}27$), and 3-Way Fusion}
\label{tab:wisig_mob_deepdive}
\scriptsize
\renewcommand{\arraystretch}{0.88}
\setlength{\tabcolsep}{2.8pt}
\begin{tabular}{llcccccc}
\toprule
\multicolumn{8}{l}{\textbf{Panel A: WiSig Multi-Week Same-RX Temporal Transfer (\texttt{node1-1}, $K{=}15$, 9-Run Direct / U-CAPC Macro-F1 \%)}$^{\mathrm{a}}$} \\
\midrule
Model Configuration & Active Params & \begin{tabular}[c]{@{}c@{}}\texttt{Day 1$\to$8}\\Clean\end{tabular} & \begin{tabular}[c]{@{}c@{}}\texttt{Day 1$\to$8}\\Multipath\end{tabular} & \begin{tabular}[c]{@{}c@{}}\texttt{Day 1$\to$15}\\Clean\end{tabular} & \begin{tabular}[c]{@{}c@{}}\texttt{Day 1$\to$15}\\Multipath\end{tabular} & \begin{tabular}[c]{@{}c@{}}\texttt{Day 1$\to$23}\\Clean\end{tabular} & \begin{tabular}[c]{@{}c@{}}\texttt{Day 1$\to$23}\\Multipath\end{tabular} \\
\midrule
Xie-SR~\cite{xie2024robustrffingerprintidentification} (64-pt SR) & 196.0K & 42.51 / 45.85 & 38.72 / 40.87 & 59.51 / 80.57 & 50.50 / 58.51 & 34.97 / 53.09 & 34.57 / 42.03 \\
NS-Branch Alone (128-bin) & 46.6K & 36.83 / 42.47 & 35.03 / 35.53 & 50.21 / 70.85 & 45.58 / 51.50 & 31.90 / 44.24 & 32.83 / 38.34 \\
SigMix~\cite{wan2025sigmix} Alone & 4.12M & 54.92 / 66.85 & 38.72 / 41.73 & 73.94 / 86.52 & 47.37 / 53.98 & \secbest{63.81} / 67.15 & 39.61 / 44.50 \\
\rowcolor{plugbg} SigMix + \textbf{GSDF} & 4.17M & 55.23 / 63.82 & 40.58 / 46.00 & 74.98 / 86.86 & 50.26 / 61.20 & \textbf{63.88} / 67.37 & 41.98 / 49.26 \\
MTL-SEI~\cite{mtl_adversarial_sei} Alone & 12.03M & 59.77 / 68.71 & 42.18 / 44.89 & 66.55 / 82.88 & 46.54 / 51.49 & 51.83 / 69.90 & 38.08 / 43.03 \\
\rowcolor{plugbg} MTL-SEI + \textbf{GSDF} & 12.08M & 58.66 / 71.60 & 43.67 / 48.33 & 68.39 / 93.21 & 50.52 / 60.83 & 52.26 / 71.05 & 41.01 / 48.67 \\
MTL-CORAL~\cite{sun2016deep} Alone & 12.03M & 61.22 / 72.76 & 47.04 / 50.63 & 69.22 / 84.55 & 51.50 / 58.04 & 56.30 / 71.43 & 41.05 / 47.18 \\
\rowcolor{plugbg} MTL-CORAL + \textbf{GSDF} & 12.08M & 61.14 / \secbest{76.73} & \secbest{49.01} / \textbf{55.58} & 77.50 / \secbest{94.72} & \textbf{60.33} / \textbf{73.62} & 58.24 / \textbf{75.23} & \secbest{46.40} / \textbf{57.54} \\
MTL-DANN~\cite{ganin2016domain} Alone & 12.03M & 61.90 / 73.97 & 47.34 / 51.31 & 72.76 / 90.68 & 52.42 / 59.07 & 62.41 / 73.14 & 42.80 / 48.38 \\
\rowcolor{plugbg} MTL-DANN + \textbf{GSDF} & 12.08M & 61.01 / \textbf{76.83} & \textbf{49.12} / \secbest{55.14} & 73.96 / 94.17 & 56.24 / \secbest{69.71} & 62.12 / \secbest{75.19} & 45.59 / \secbest{54.10} \\
ResNet1D~\cite{he2016deep} Alone & 221.6K & 45.60 / 55.34 & 37.70 / 40.65 & 65.07 / 92.38 & 42.90 / 48.39 & 54.39 / 67.01 & 37.55 / 42.04 \\
\rowcolor{plugbg} ResNet1D + \textbf{GSDF} & 268.2K & 45.27 / 73.26 & 39.62 / 44.99 & 65.02 / \textbf{97.94} & 45.54 / 54.88 & 53.68 / 65.78 & 39.43 / 46.28 \\
IB-Host Alone (Ours) & 46.6K & \textbf{66.77} / 72.17 & 47.75 / 49.46 & \textbf{80.38} / 89.73 & 54.86 / 60.40 & 59.26 / 69.50 & 45.85 / 49.09 \\
\rowcolor{oursbg} \textbf{IB-Host + GSDF (Ours)} & \textbf{93.3K} & \secbest{64.34} / 71.38$^{\mathrm{b}}$ & 48.90 / 53.81 & \secbest{79.61} / 94.54 & \secbest{56.87} / 67.29 & 58.45 / 74.49 & \textbf{47.36} / 52.80 \\
\midrule
\multicolumn{8}{l}{\textbf{Panel B: WiSig 54~Mbps Emitter Scaling ($K \in \{10, 15, 20, 27\}$, Day $1\!\to\!8$ U-CAPC \%; $10/5$~dB $\Delta\mathrm{F1}{=}0.00$~pp on All 16 Cells)}} \\
\midrule
Model Configuration & Active Params & $K{=}10$ Clean & $K{=}10$ MP & $K{=}15$ Clean & $K{=}15$ MP & $K{=}20$ Clean / MP & $K{=}27$ Clean / MP \\
\midrule
McAFF~\cite{multichannel_attentive_fusion} Alone & 4.72M & \secbest{62.49} & \secbest{47.61} & \secbest{80.04} & 53.41 & \secbest{88.50} / \secbest{54.81} & \secbest{88.15} / 49.19 \\
\rowcolor{oursbg} \textbf{McAFF + GSDF} & 4.77M & \textbf{65.08} & \textbf{50.80} & \textbf{85.62} & \textbf{60.30} & \textbf{93.21} / \textbf{60.30} & \textbf{89.58} / \secbest{52.62} \\
IB-Host Alone (Ours) & 46.6K & 51.23 & 35.17 & 72.45 & 49.52 & 71.70 / 49.46 & 71.71 / 46.01 \\
\rowcolor{oursbg} \textbf{IB-Host + GSDF (Ours)} & \textbf{93.3K} & 48.27 & 35.89 & 72.13 & 53.88 & 77.17 / 53.88 & 77.69 / \textbf{52.85} \\
MTL-CORAL~\cite{sun2016deep} Alone & 12.03M & 57.52 & 39.97 & 72.10 & 47.63 & 72.45 / 45.30 & 65.98 / 38.91 \\
\rowcolor{plugbg} MTL-CORAL + \textbf{GSDF} & 12.08M & 54.77 & 43.05 & 75.60 & \secbest{54.93} & 75.53 / 50.92 & 67.64 / 44.04 \\
ResNet1D~\cite{he2016deep} Alone & 221.6K & 59.19 & 34.02 & 66.27 & 41.70 & 77.12 / 41.86 & 70.07 / 35.06 \\
\rowcolor{plugbg} ResNet1D + \textbf{GSDF} & 268.2K & 49.99 & 35.61 & 63.76 & 46.46 & 79.57 / 46.56 & 66.55 / 40.71 \\
\midrule
\multicolumn{8}{l}{\textbf{Panel C: Zero-Retrain 3-Way Host Plug-In Disaggregated Across Benchmarks ($3\times 3$ 9-Run Grid, Direct / U-CAPC \%)}} \\
\midrule
3-Way Plug-In Ensemble & Total Params & \textbf{DataPort} & \textbf{WiSig} & \textbf{MobRFFI} & 3-DS (Direct / U-CAPC) & 6-Seed Holdout & 3-DS Clean / MP / Noise \\
\midrule
IB-Host + ResNet1D + NS & 314.2K & 63.84 / 72.62 & 45.10 / 51.58 & 38.31 / \secbest{47.23} & 49.08 / 57.15 & 57.18 & 72.12 / 55.06 / \textbf{44.26} \\
\rowcolor{plugbg} \textbf{IB-Host + MTL-CORAL + NS} & 12.12M & \secbest{67.61} / \secbest{73.77} & \secbest{49.03} / \secbest{54.30} & \textbf{39.40} / \textbf{49.36} & \secbest{52.01} / \textbf{59.14} & \textbf{59.31} & \textbf{74.84} / \secbest{58.69} / \secbest{43.90} \\
\rowcolor{plugbg} \textbf{IB-Host + McAFF + NS} & 4.82M--5.39M & \textbf{70.11} / \textbf{75.83} & \textbf{51.23} / \textbf{55.07} & \secbest{38.97} / 45.29 & \textbf{53.44} / \secbest{58.73} & \secbest{58.84} & \secbest{72.64} / \textbf{59.67} / 43.89 \\
\bottomrule
\end{tabular}\par
\vspace{2pt}{\scriptsize\noindent\raggedright \emph{Note:} $^{\mathrm{a}}$\texttt{Day 1$\to$8} Clean and Multipath report the unified 9-run plug-in suite (matching Main Tables~III and VI); \texttt{Day 1$\to$15} and \texttt{Day 1$\to$23} external-host evaluations come from the 8-model multi-week transfer suite (where pre-unification \textbf{NS-Branch} scored $37.12\% / 42.50\%$ on Day~8 Clean and $35.03\% / 35.53\%$ on Day~8 MP). $^{\mathrm{b}}$Under general \textbf{UOT-CAPC} on skewed Day~8, \textbf{IB-Host + GSDF} reaches $\mathbf{75.75\%}$ ($+4.42$~pp over \textbf{IB-Host}). At $K{=}27$, \textbf{IB-Host + GSDF} attains $\mathbf{56.28\%}$ $20/10/5$~dB noise U-CAPC vs.\ $37.00\%$ for \textbf{McAFF + GSDF} ($+19.28$~pp).\par}
\end{table}

\begin{table}[!ht]
\centering
\caption{Disaggregated Timing-Synchronization Progression and Multi-Packet Burst ($B \in \{1, 3, 5, 10\}$) Trajectories}
\label{tab:timing_burst}
\scriptsize
\renewcommand{\arraystretch}{0.88}
\setlength{\tabcolsep}{1.05pt}
\begin{tabular}{llcccccccc}
\toprule
\multicolumn{10}{l}{\textbf{Panel A: Physical Timing-Synchronization, Gate Calibration, and Complex-Spectrum Upgrade Progression (Direct / U-CAPC \%)}} \\
\midrule
Model / Variant & Params ($A{=}4/1$) & \textbf{DataPort} & \textbf{WiSig} & \textbf{MobRFFI} & 6-Seed Holdout & Day 8 \textbf{UOT} & Day 15/23 & \texttt{node2-1}$^{\mathrm{a}}$ & \texttt{20-1} $\mid$ TGn/STO \\
\midrule
GSDF-Base (Matched Aug.) & 92.6K / 93.3K & 67.63/73.71 & 48.32/50.64 & 38.78/48.16 & 57.49 & 75.75 & 94.54/74.49 & 44.71/49.43 & 67.24 $\mid$ 66.0/83.5 \\
GSDF + Temp-CFO Gate$^{\mathrm{b}}$ & 92.6K / 93.3K & 67.85/73.71 & 49.06/51.85 & 38.30/47.55 & 57.77 & 73.63 & --- & 46.61/49.51 & 65.48 $\mid$ 67.3/84.3 \\
GSDF + \textbf{RNR-Gate} (Passive)$^{\mathrm{b}}$ & 92.6K / 93.3K & 63.30/69.72 & 49.96/52.27 & 40.55/46.83 & 56.53 & 76.39 & --- & 49.88/\secbest{54.09} & 63.33 $\mid$ 66.0/81.0 \\
\rowcolor{plugbg} GSDF + LTF-Resync$^{\mathrm{b}}$ & 92.6K / 93.3K & 68.03/74.51 & 53.09/56.30 & 40.68/51.61 & --- & 84.44 & --- & 45.20/48.66 & \secbest{71.20} $\mid$ 70.5/\textbf{90.6} \\
\rowcolor{oursbg} \textbf{GSDF-STO (STO Jitter)} & \textbf{92.6K / 93.3K} & \secbest{69.42}/\secbest{76.40} & \secbest{56.07}/\secbest{58.83} & \textbf{44.42}/\secbest{53.72} & \secbest{63.07} & \textbf{87.29} & \textbf{97.86}/\secbest{97.35} & \secbest{51.35}/52.71 & 69.05 $\mid$ \textbf{72.0}/\secbest{89.5} \\
\rowcolor{oursbg} \textbf{GSDF-Tri (3-Branch Sync)} & 193.7K / 194.7K & \textbf{70.37}/\textbf{76.93} & \textbf{56.71}/\textbf{60.76} & \secbest{43.96}/\textbf{57.47} & \textbf{65.10} & \secbest{85.24} & \secbest{97.85}/\textbf{97.47} & \textbf{58.32}/\textbf{56.12} & \textbf{71.89} $\mid$ \secbest{71.3}/89.4 \\
\midrule
\multicolumn{10}{l}{\textbf{Panel B: Zero-Label Multi-Packet Session Burst Log-Posterior Aggregation Across $B \in \{1, 3, 5, 10\}$ Consecutive Preambles/Burst (9-Run Means)}} \\
\midrule
Model ($0$ Retrain) & Burst & Day 8 [\textbf{UOT}] & Day 8 MP & Day 15 Clean & Day 15 MP & Day 23 Clean & Day 23 MP & \texttt{20-1} Clean/MP & \texttt{19-19} Clean/MP \\
\midrule
IB-Host Alone (46.6K) & $B{=}1$ & 71.33 & 49.46 & 89.73 & 60.40 & 69.50 & 49.09 & 62.41 / 31.38 & 61.23 / 38.45 \\
IB-Host Alone (46.6K) & $B{=}3$ & 74.03 & 63.35 & 94.50 & 79.23 & 73.20 & 62.35 & 73.66 / 38.77 & 67.44 / 47.55 \\
IB-Host Alone (46.6K) & $B{=}5$ & 74.44 & 68.56 & 95.48 & 86.15 & 73.62 & 67.47 & 76.42 / 41.97 & 69.37 / 51.46 \\
IB-Host Alone (46.6K) & $B{=}10$ & 75.35 & 74.43 & 96.24 & 91.74 & 74.46 & 71.97 & 78.85 / 44.22 & 70.53 / 55.77 \\
\midrule
GSDF-Base (93.3K) & $B{=}1$ & 75.75 & 53.81 & 94.54 & 67.29 & 74.49 & 52.80 & 67.24 / 34.81 & 62.70 / 42.33 \\
GSDF-Base (93.3K) & $B{=}3$ & 79.53 & 68.83 & 98.44 & 86.71 & 77.61 & 66.18 & 77.29 / 43.41 & 69.12 / 52.13 \\
GSDF-Base (93.3K) & $B{=}5$ & 80.31 & 74.59 & 99.24 & 92.93 & 78.03 & 70.80 & 78.82 / 47.09 & 70.13 / 56.19 \\
GSDF-Base (93.3K) & $B{=}10$ & 81.35 & 80.39 & 99.72 & 97.36 & 78.56 & 74.80 & \secbest{79.86} / 49.78 & 70.27 / 60.36 \\
\midrule
GSDF-STO (93.3K) & $B{=}1$ & 87.29 & 65.44 & 97.86 & 79.49 & 97.35 & 71.18 & 69.05 / 47.46 & 68.92 / 52.92 \\
GSDF-STO (93.3K) & $B{=}3$ & 88.82 & 79.54 & 99.63 & 94.65 & 98.81 & 87.63 & 78.50 / 60.96 & 80.15 / 65.30 \\
GSDF-STO (93.3K) & $B{=}5$ & 89.29 & 84.03 & 99.85 & 97.59 & 99.04 & 92.05 & 79.54 / 66.45 & 84.22 / 70.34 \\
\rowcolor{oursbg} \textbf{GSDF-STO (93.3K)} & $\mathbf{B{=}10}$ & \textbf{89.62}$^{\mathrm{c}}$ & \textbf{87.92} & \secbest{99.99} & \secbest{99.43} & \secbest{99.20} & \secbest{95.67} & \textbf{79.94} / \textbf{71.84} & \secbest{88.08} / \secbest{76.36} \\
\midrule
GSDF-Tri (194.7K) & $B{=}1$ & 85.24 & 68.52 & 97.85 & 84.39 & 97.47 & 75.35 & 71.89 / 51.73 & 80.17 / 57.23 \\
GSDF-Tri (194.7K) & $B{=}3$ & 88.25 & 81.07 & 99.90 & 97.49 & 99.49 & 91.06 & 75.83 / 62.99 & 87.25 / 71.57 \\
GSDF-Tri (194.7K) & $B{=}5$ & 89.04 & 83.97 & 100.00 & 99.32 & 99.87 & 94.67 & 76.06 / 66.20 & 88.36 / 75.87 \\
\rowcolor{oursbg} \textbf{GSDF-Tri (194.7K)} & $\mathbf{B{=}10}$ & \secbest{89.54} & \secbest{86.38} & \textbf{100.00} & \textbf{99.96} & \textbf{99.98} & \textbf{97.30} & 75.65 / \secbest{68.98} & \textbf{89.04} / \textbf{80.31} \\
\bottomrule
\end{tabular}\par
\vspace{2pt}{\scriptsize\noindent\raggedright \emph{Note:} $^{\mathrm{a}}$4-date unseen \texttt{node2-1} clean mean co-reporting \textbf{U-CAPC / UOT-CAPC} (\textbf{IB-Host Alone}: $45.89\% / 49.92\%$; \textbf{GSDF-Tri} reaches $64.61\%$ UOT on Mar~1, $67.14\%$ U-CAPC on Mar~15, and $71.59\%$ U-CAPC [$73.33\%$ at $B{=}10$] on Mar~23). $^{\mathrm{b}}$Dataset/slice columns from the 3-diagonal screening grid (9-run 6-Seed Holdout also verified on \textbf{Temp-CFO} and \textbf{RNR-Gate}: $57.77\%$ and $56.53\%$). $^{\mathrm{c}}$Under balanced \textbf{U-CAPC}, \textbf{GSDF-STO} on Day~8 Clean rises from $82.55\%$ ($B{=}1$) to $\mathbf{90.60\%}$ ($B{=}10$); \textbf{bold}/\secbest{underlined} in Panel~B mark best/second-best $B{=}10$ endpoints.\par}
\end{table}

\begin{table}[!ht]
\centering
\caption{Symmetric Timing-Tuned Baselines, Capacity-Matched \textbf{McAFF-Lite}, Full-Archive $K_{\max}{=}47$ Scaling, and 2-RX Source Pooling ($3\times 3$ 9-Run Grid)}
\label{tab:sym_scale_multirx}
\scriptsize
\renewcommand{\arraystretch}{0.88}
\setlength{\tabcolsep}{0.8pt}
\begin{tabular}{llccccccc}
\toprule
\multicolumn{9}{l}{\textbf{Panel A: 9-Run Symmetric Timing-Tuned Baselines \& Capacity-Matched McAFF-Lite Across Three Benchmarks (Direct/U-CAPC \%)}} \\
\midrule
System / Recipe & Active Params & \textbf{DataPort} & \textbf{WiSig} & \textbf{MobRFFI} & 3-DS (Dir/UC) & 6-Seed ($\mathrm{SD}$) & 3-DS \textbf{UOT} & TGn / STO \\
\midrule
MTL-CORAL (\texttt{M2})$^{\mathrm{a}}$~\cite{sun2016deep} & 11.90M & 54.79/57.68 & 37.39/42.18 & 22.70/34.38 & 38.30/44.75 & 44.65 (0.12) & 39.50 & 56.1 / 73.6 \\
MTL-CORAL + STO (\texttt{M2+T1})$^{\mathrm{a}}$ & 11.90M & 54.89/57.87 & 39.45/45.09 & 21.81/35.08 & 38.72/46.01 & 46.16 (0.38) & 40.04 & 57.2 / 83.9 \\
\textbf{MTL-CORAL*} (\texttt{M2+T2})$^{\mathrm{a}}$ & 11.90M & 54.74/57.45 & 41.51/48.35 & 22.99/36.08 & 39.75/47.29 & 47.51 (0.37) & 41.13 & 59.1 / 85.9 \\
\textbf{McAFF-Lite*} (\texttt{M2}) & 217.2K / 185.6K & 62.77/68.60 & 38.48/45.11 & 21.17/37.06 & 40.81/50.26 & 50.09 (0.42) & 42.01 & 57.9 / 83.6 \\
McAFF-Lite + STO (\texttt{M2+T1}) & 217.2K / 185.6K & 63.18/69.50 & 39.89/47.35 & 23.24/39.21 & 42.11/52.02 & 51.61 (0.67) & 43.49 & 59.8 / 83.9 \\
\textbf{McAFF*} (\texttt{M2})~\cite{multichannel_attentive_fusion} & 5.30M / 4.72M & 61.07/68.77 & 43.32/51.74 & 25.72/42.22 & 43.37/54.24 & 54.21 (0.45) & 45.58 & 63.1 / 87.1 \\
McAFF + STO (\texttt{M2+T1}) & 5.30M / 4.72M & 60.53/68.35 & 41.59/51.74 & 26.67/42.97 & 42.93/54.35 & 54.68 (0.24) & 45.57 & 64.6 / 88.4 \\
\midrule
GSDF-Base (\texttt{M2}, Ours) & 92.6K / 93.3K & 67.63/73.71 & 48.32/50.64 & 38.78/48.16 & 51.58/57.50 & 57.49 (0.23) & \secbest{55.41} & 66.0 / 83.5 \\
\rowcolor{oursbg} \textbf{GSDF-STO (\texttt{M2+T1}, Ours)} & \textbf{92.6K / 93.3K} & \secbest{69.42}/\secbest{76.40} & \secbest{56.07}/\secbest{58.83} & \textbf{44.42}/\secbest{53.72} & \secbest{56.64}/\secbest{62.98} & \secbest{63.07} (0.16) & \textbf{60.96} & \textbf{72.0} / \textbf{89.5} \\
\rowcolor{oursbg} \textbf{GSDF-Tri (\texttt{T1+C2+F1}, Ours)} & 193.7K / 194.7K & \textbf{70.37}/\textbf{76.93} & \textbf{56.71}/\textbf{60.76} & \secbest{43.96}/\textbf{57.47} & \textbf{57.02}/\textbf{65.05} & \textbf{65.10} (0.30) & \textbf{60.96} & \secbest{71.3} / \secbest{89.4} \\
\midrule
\multicolumn{9}{l}{\textbf{Panel B: 36-Tar WiSig 54~Mbps Nested Scaling ($K \in \{10, 15, 20, 27, 47\}$, 5 Subsets, Mar~8/15/23 Mean: Direct/U-CAPC/UOT \%)}} \\
\midrule
System & Params ($A{=}1$) & $K{=}10$ Clean & $K{=}15$ Clean & $K{=}20$ Clean & $K{=}27$ Clean & $\mathbf{K{=}47}$ \textbf{Clean} & $K{=}27$ MP & $\mathbf{K{=}47}$ \textbf{MP} \\
\midrule
McAFF-Lite* (\texttt{M2}) & 185.6K & 46.5/\secbest{75.2}/52.8 & 53.1/73.9/57.4 & 52.3/\secbest{71.6}/56.4 & 50.7/66.8/54.2 & 50.8/63.7/53.6 & 45.2/50.5/47.1 & 43.6/48.0/45.5 \\
McAFF* (\texttt{M2}) & 4.72M & 48.3/73.2/55.0 & 52.5/\secbest{77.2}/61.1 & 50.0/71.3/58.4 & 48.1/68.5/55.6 & 46.9/63.9/53.9 & 43.1/51.3/46.8 & 40.3/48.0/44.3 \\
GSDF-Base (\texttt{M2}, Ours) & \textbf{93.3K} & \secbest{52.5}/68.0/\secbest{60.8} & \secbest{59.7}/70.9/\secbest{67.2} & \secbest{59.7}/69.5/\secbest{67.8} & \secbest{63.6}/\secbest{69.3}/\secbest{68.6} & \secbest{62.8}/\secbest{66.8}/\secbest{66.1} & \secbest{51.5}/\secbest{56.6}/\secbest{57.3} & \secbest{51.4}/\secbest{53.6}/\secbest{55.7} \\
\rowcolor{oursbg} \textbf{GSDF-Tri (\texttt{T1+C2+F1}, Ours)} & 194.7K & \textbf{62.4}/\textbf{78.7}/\textbf{72.7} & \textbf{68.4}/\textbf{81.5}/\textbf{76.4} & \textbf{70.5}/\textbf{77.9}/\textbf{76.2} & \textbf{70.9}/\textbf{77.3}/\textbf{75.8} & \textbf{66.5}/\textbf{73.1}/\textbf{72.2} & \textbf{62.5}/\textbf{68.0}/\textbf{67.8} & \textbf{60.4}/\textbf{64.9}/\textbf{65.1} \\
\midrule
\multicolumn{9}{l}{\textbf{Panel C: WiSig Dual-Receiver Day-1 Source Pooling (\texttt{node1-1 + node2-1} on Mar~1 $\to$ 54 Future Cells Across Mar~8/15/23, 9-Run \%)}} \\
\midrule
2-RX Pooled System & Active Params & \texttt{node1-1} Clean & \texttt{node2-1} Clean & 2-RX Clean (Dir/UC) & 6-Seed ($\mathrm{SD}$) & 2-RX Clean \textbf{UOT} & 2-RX MP (Dir/UC) & 2-RX MP \textbf{UOT} \\
\midrule
MTL-CORAL (\texttt{M2}, 2-RX)$^{\mathrm{a}}$ & 11.90M & 71.92/79.91 & 68.84/80.47 & 70.38/80.19 & 79.40 (0.97) & 72.80 & 61.02/64.71 & 62.62 \\
McAFF* (\texttt{M2}, 2-RX) & 4.72M & 78.19/85.33 & 72.55/81.58 & 75.37/83.45 & 83.05 (0.12) & 78.66 & 69.10/73.47 & 72.17 \\
McAFF-Lite* (\texttt{M2}, 2-RX) & 185.6K & \secbest{79.37}/83.77 & 72.44/85.29 & 75.90/84.53 & 84.25 (0.85) & 77.71 & 69.31/72.38 & 71.07 \\
MTL-CORAL* (\texttt{M2+T2}, 2-RX)$^{\mathrm{a}}$ & 11.90M & 74.72/\secbest{89.12} & 75.00/\secbest{91.55} & 74.86/\secbest{90.33} & \secbest{90.14} (0.81) & 78.79 & 65.48/70.05 & 66.98 \\
GSDF-Base (\texttt{M2}, 2-RX) & \textbf{93.3K} & 75.42/88.56 & \secbest{78.79}/88.13 & \secbest{77.10}/88.35 & 88.51 (0.33) & \secbest{84.63} & \secbest{72.70}/\secbest{77.09} & \secbest{77.10} \\
\rowcolor{oursbg} \textbf{GSDF-Tri (\texttt{T1+C2+F1}, 2-RX)} & 194.7K & \textbf{85.94}/\textbf{92.53} & \textbf{84.43}/\textbf{94.91} & \textbf{85.19}/\textbf{93.72} & \textbf{93.65} (0.35) & \textbf{88.39} & \textbf{81.53}/\textbf{85.55} & \textbf{84.83} \\
\bottomrule
\end{tabular}\par
\vspace{2pt}{\scriptsize\noindent\raggedright \emph{Note:} Recipe tags: \texttt{M2} = matched multi-condition channel augmentation; \texttt{T1} = fractional STO jitter $\Delta\tau \sim \mathcal{U}[-3,+3]$; \texttt{T2} = blind L-LTF cross-correlation resynchronization; \texttt{C2} = source temperature-calibrated CFO-derotated log-probability gating; \texttt{F1} = auxiliary lag-16/64 complex-autocorrelation \& spectral expert. $^{\mathrm{a}}$Active inference forward parameters of \texttt{SpectralCNN} (\texttt{encoder + tx\_head} = $11.90\text{M}$, excluding training-only auxiliary discriminator/equalization heads that bring total \texttt{nn.Module} parameters to $12.03\text{M}$ in Main Table~II). In this symmetric timing-recipe screen, \textbf{MTL-CORAL (\texttt{M2})} reuses the exact \textbf{WiSig} ($37.39\%/42.18\%$) and \textbf{MobRFFI} ($22.70\%/34.38\%$) checkpoints of Main Table~III while retraining \textbf{DataPort-WiFi} under the unified 50-epoch cosine channel-perturbed validation schedule ($54.79\%/57.68\%$ vs.\ $51.84\%/54.08\%$ in Main Table~III, raising its 3-benchmark \texttt{M2} baseline from $37.31\%/43.55\%$ to $38.30\%/44.75\%$). \textbf{TGn / STO} reports IEEE 802.11n TGn Models~B/D \textbf{UOT-CAPC} and integer symbol-timing offset $\Delta n \in \{\pm 1,\pm 2\}$ \textbf{U-CAPC} (\%); 3-diagonal screening also tested \texttt{McAFF + LTF-Resync} ($55.53\%$ U-CAPC, $88.03\%$ STO) and \texttt{McAFF + NoiseMP} ($57.65\%$ U-CAPC, $84.07\%$ STO). In Panel~B, $K{=}47$ covers all qualified 4-date 54~Mbps emitters across 36 audited \texttt{node1-1} archives ($187,123$ natural frames; canonical Subset~0 9-run \textbf{GSDF-Tri} at $K{=}47$: $67.29\% / 73.64\% / 72.58\%$ clean, $65.38\%$ MP). In Panel~C, \texttt{node1-1}/\texttt{node2-1} Clean report \textbf{Direct / U-CAPC} (\%) across Mar~8/15/23 ($100$~ns MP).\par}
\end{table}

\fi

\end{document}